\documentclass[10pt, twocolumn, letterpaper]{article}
\usepackage[accsupp]{axessibility}
\usepackage{iccv}
\usepackage{times}
\usepackage{epsfig}
\usepackage{graphicx}
\usepackage{amsmath}
\usepackage{amssymb}
\usepackage{style}
\usepackage{algorithm}
\usepackage{algpseudocode}

\usepackage[utf8]{inputenc} 
\usepackage[T1]{fontenc}    

\usepackage{booktabs}       
\usepackage{amsfonts}       
\usepackage{nicefrac}       
\usepackage{microtype}      
\usepackage{xcolor}         
\usepackage{amsmath,mathtools}
\usepackage{amssymb}
\usepackage{amsthm}
\usepackage{graphicx}
\usepackage{enumitem}
\usepackage{bm}
\usepackage{bbm}
\usepackage{enumitem}
\usepackage{tabularx}
\usepackage{multirow}
\usepackage{adjustbox}
\usepackage{multicol}
\usepackage{authblk}
\DeclareMathOperator{\diag}{diag}

\usepackage{tabularx}
\newcolumntype{L}{>{\raggedright\arraybackslash}X}
\newcolumntype{C}{>{\centering\arraybackslash}X}

\definecolor{georgecolor}{RGB}{255, 87, 51}

\definecolor{mkcolor}{RGB}{255,0, 128}

\definecolor{shimincolor}{RGB}{0,128,128}

\definecolor{kecolor}{RGB}{0,128,0}

\definecolor{leocolor}{RGB}{0,0,255}

\newcommand{\methodname}{{{DiffFacto}}\xspace}
\newcommand{\dif}{\mathrm{d}}

\newcommand{\tPart}{{\tau}}     
\newcommand{\tAll}{{\bm{\tPart}}}      
\newcommand{\TPart}{{T}}     
\newcommand{\TAll}{{\mathbf{\TPart}}}     

\usepackage[pagebackref=true,breaklinks=true,letterpaper=true,colorlinks,bookmarks=false]{hyperref}
\usepackage{xfp}
\newcommand\SupplementaryMaterials{%
  \xdef\presupfigures{\arabic{figure}}
  \xdef\presuptables{\arabic{table}}
  \xdef\presupsections{\arabic{section}}
  \renewcommand\thefigure{S\fpeval{\arabic{figure}-\presupfigures}}
  \renewcommand\thetable{S\fpeval{\arabic{table}-\presuptables}}
  \renewcommand\thesection{S\fpeval{\arabic{section}-\presupsections}}
}
\usepackage{cleveref}
\crefname{section}{\S}{\S\S}
\crefname{subsection}{\S}{\S\S}
\crefname{conj}{Conj.}{Conj.}

\Crefname{assumption}{\textbf{H}\hspace{-3pt}}{\textbf{H}\hspace{-3pt}}
\crefname{assumption}{\textbf{H}}{\textbf{H}}

\crefname{algorithm}{\text{Alg.}}{\text{Alg.}}
\crefname{assumption}{\textbf{H}}{\textbf{H}}
\crefname{equation}{\text{Eq}}{\text{Eq}}
\crefname{definition}{\text{Dfn.}}{\text{Dfn.}}
\crefname{lemma}{\text{Lemma}}{\text{Lemma}}
\crefname{dfn}{\text{Dfn.}}{\text{Dfn.}}
\crefname{thm}{\text{Thm.}}{\text{Thm.}}
\crefname{tab}{\text{Tab.}}{\text{Tab.}}
\crefname{fig}{\text{Fig.}}{\text{Fig.}}
\crefname{table}{\text{Tab.}}{\text{Tab.}}
\crefname{figure}{\text{Fig.}}{\text{Fig.}}
\newcommand{\parahead}[1]{\vspace{1.5mm}\noindent\textbf{#1.}\ }

\iccvfinalcopy 

\def\iccvPaperID{1943} 

\ificcvfinal\fi

\begin{document}
\title{\methodname: Controllable Part-Based 3D Point Cloud Generation \\ with Cross Diffusion \vspace{-0.8cm}}

\author{George Kiyohiro Nakayama$^{1}$~~~Mikaela Angelina Uy$^{1}$~~~Jiahui Huang$^{2}$\\ 
Shi-Min Hu$^{2}$~~~Ke Li$^{3}$~~~Leonidas Guibas$^{1}$\\
\vspace{0.2cm}
$^1$Stanford University~~~~~$^2$Tsinghua University~~~~~$^3$Simon Fraser University
}

\twocolumn[{%
\renewcommand\twocolumn[1][]{#1}%
\begin{center}
    \captionsetup{type=figure}
    \vspace{-1.2cm}
\maketitle
\includegraphics[width=\textwidth]{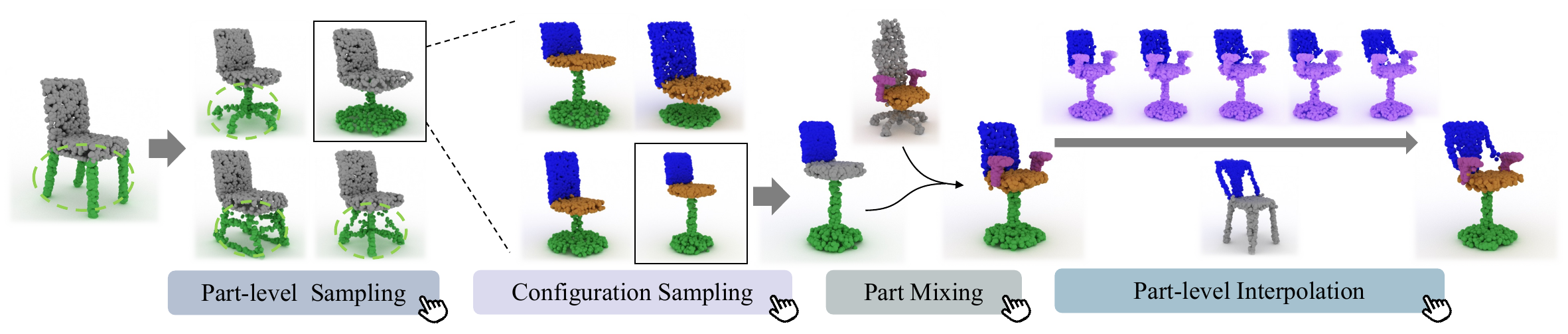}
    \captionof{figure}{\textbf{\methodname}: Our approach tackles the task of controllable part-based point cloud generation, where we are able to generate novel shapes - novel configurations of novel parts. Our probabilistic generative model learns a \emph{factorized} representation of shapes with a \emph{cross diffusion} network allowing for control. We demonstrate that \textbf{\methodname} not only enables controllable generation but also various shape editing tasks.}
\end{center}
}]
\begin{abstract}


   While the community of 3D point cloud generation has witnessed a big growth in recent years, there still lacks an effective way to enable intuitive user control in the generation process, hence limiting the general utility of such methods. Since an intuitive way of decomposing a shape is through its parts, we propose to tackle the task of controllable part-based point cloud generation. We introduce \textbf{\methodname}, a novel probabilistic generative model that learns the distribution of shapes with part-level control. We propose a \emph{factorization} that models independent part style and part configuration distributions, and present a novel \emph{cross diffusion} network that enables us to generate coherent and plausible shapes under our proposed factorization. Experiments show that our method is able to generate novel shapes with multiple axes of control. It achieves state-of-the-art part-level generation quality and generates plausible and coherent shape while enabling various downstream editing applications such as shape interpolation, mixing, and transformation editing. Please visit our project webpage at \url{https://difffacto.github.io/}
\end{abstract}
\vspace{-2em}
\section{Introduction}


3D shape generation~\cite{xu2022survey} is an important and popular task
, where point clouds are one popular representation~\cite{pointflow, ShapeGF, lion, dpm} -- due to their \emph{simple yet powerful} expressivity as well as data availability, \ie just a set of points and can directly be acquired by sensors. %
However, the generation of arbitrary plausible shapes is often of limited utility, as users often have a conceptual design idea and of what they want to generate.


A prerequisite of shape generation is to be able to learn a space of all possible shapes. To represent this space, one parsimonious way is to represent them as a combination of simpler atoms, known as \emph{parts}. In this flavor, we propose the task of \emph{controllable part-based generation}, which aims to generate plausible novel shapes with user control over individual parts. As mentioned before, a shape is a combination of parts, thus a `novel shape' can be defined in three different ways: 
(1) \emph{novel} configurations of existing parts, 
(2) existing configurations of \emph{novel} parts, and 
(3) \emph{novel} configurations of \emph{novel} parts. 

The first is explored in existing graphics literature such as part retrieval~\cite{Sung:2017} and shape assembly~\cite{coalesce}, while the second can be tackled by existing generation methods~\cite{pointflow, ShapeGF, lion} trained on parts. In contrast, we tackle the third, which is the more challenging case subsuming the first and second. 
A challenge arises because a shape is a combination of novel parts, leading to an exponential explosion of plausible shapes while having only  limited training data. 
A further challenge stems from enabling control as this requires an approach that can vary individual parts and configurations while still generating plausible shapes.

To this end, we introduce a new method that tackles this task in a principled way by building a \emph{probabilistic generative model} that learns the distribution of shapes while enabling control on parts and configurations.
Specifically, we propose a \emph{factorization} that decomposes the shape space into 
(i) the individual canonicalized (semantic) parts, and 
(ii) their transformations (position and size). These factors can be sampled or encoded independently, allowing for different modes of control in generation and intuitive editing. 




Our approach learns independent latent spaces for each canonicalized (semantic) part through \emph{part stylizers}. Then conditioned on the canonicalized parts, we also introduce learn a \emph{transformation sampler} that learns a distribution of part configurations. Naive approaches can result in mode collapse since multiple parameter configurations can output a valid shape, if conditioned only on canonicalized parts. We leverage on a sampling-based approach to learn a multimodal distribution of part configurations through conditional Implicit Maximum Likelihood~\cite{li2020multimodal} (cIMLE).

To generate plausible shapes through independently sampled factors, we also introduce our \emph{cross diffusion} network that allows for the learning of a better shape distribution under our proposed factorization. Our cross-attention diffusion network, conditions on the proposed factors, \ie independent part style and transformations, in the reverse diffusion process. Our design allows each generated point in the point cloud to be informed of both the global shape as well as the local part, resulting in more plausible and coherent output shapes while still enabling control. Moreover, we also introduce a generalized forward diffusion kernel that allows the explicit encoding of each part transformations, enabling better shape reconstruction and transformation extrapolation.

We dub our method \textbf{\methodname}, for \textbf{facto}rized represention with cross \textbf{diff}usion. To our best knowledge, we are the first to introduce a factorized representation that allows for control in both part styles and part configurations as we model independent part style distributions and transformation distribution, enabling each to be independently sampled. Experiments show that our approach achieves better intra-part and inter-part level scores compared to baselines. We also show that our approach generates novel and coherent shapes through a segmentation-based plausibility experiment and human study. Furthermore, we demonstrate that our approach also allows for controllable and localized shape editing on various applications such as part-level shape interpolation, shape mixing and transformation editing.

\section{Related Work}
\noindent \textbf{Point Cloud Generative Models.}
The literature mainly targets at an accurate modeling of the underlying data distribution, using probabilistic tools and parameterized deep networks %
developed in variational auto-encoders (VAEs)~\cite{vae} used by PSG~\cite{fan2017point}, generative adversarial networks (GANs)~\cite{goodfellow2020generative} used by PointGAN~\cite{achlioptas2018learning}, auto-regressive models~\cite{box2015time} used by PointGrow~\cite{sun2020pointgrow}, normalizing flows~\cite{rezende2015variational} used by PointFlow~\cite{pointflow} or Softflow~\cite{kim2020softflow}, \etc, with proper conditioning~\cite{sohn2015learning,cgan,li2020multimodal}.
Notably, the most recent success of Diffusion Models~\cite{diffusionho} for image synthesis~\cite{rombach2022high} is further expanded to the domain of 3D point clouds generation~\cite{dpm,Zhou_2021_pvd,ShapeGF,lion,nichol2022point,wavelet}.
Among this line of works, PVD~\cite{Zhou_2021_ICCV} and LION~\cite{lion} treat the full point cloud as a \emph{single} sample from the learned distribution in either primal or latent space.
However, the points themselves from a cloud can be viewed as samples from a geometric distribution, which is explicitly modeled by, \eg, DPM~\cite{dpm} or ShapeGF~\cite{ShapeGF}, and PointFlow~\cite{pointflow} as \emph{distribution of distributions}~\cite{pointflow}.
This enables the desirable capability to generate \emph{arbitrary} number of points from a single shape without the need of re-training.
Our work takes this approach and additionally builds a \emph{factorized} prior to sample from allowing for part-level control, which in contrast to all previous works that only model a \emph{single} latent space and can only generate a full shape. 

\noindent \textbf{Structure-aware Shape Generation.}
Different from \emph{full-shape} generation, structure-aware methods enrich the synthesized geometry with useful structural information and allow for easy user manipulations~\cite{chaudhuri2020learning}.
Related literature has explored different structure representations, such as segmented parts~\cite{nash2017shape,li2020learning,editvae}, relationship graphs~\cite{sdmnet,wang2019planit}, or layers/hierarchies~\cite{li2017grass,zekun2020dualsdf,yang2022dsg}, generated using either auto-regressive models~\cite{vaswani2017attention,nichol2022point}, recursive networks~\cite{li2017grass}, or hierarchical models~\cite{yang2022dsg,vahdat2020nvae}.
Among these works, %
SPAGHETTI~\cite{hertz2022spaghetti} uses cross-attention to mix the part latents that are decoded into an implicit shape representation.
SP-GAN~\cite{li2021sp} employs a spherical proxy geometry for even finer-grained controls. 
Nevertheless, most of the above existing methods model the shape prior as a single latent code, and user controls are \emph{detached} from the generative process.
They hence rely on post-optimization/inversion~\cite{zekun2020dualsdf} steps to maintain the shape plausibility.
In contrast, our \emph{factorized} prior allows for explicit part-based control and sampling \emph{during} the generative process. 

\noindent \textbf{Shape Editing.}
In graphics, \emph{low-level} shape editing usually involves computing and modifying geometric handles such as cages~\cite{sederberg1986free,corda2020real} or skeletons~\cite{yoshizawa2003free,borosan2010hybrid}.
On the other hand, the exploitation of \emph{high-level} semantics allow for more intuitive controls of the shape such as mixing and stitching semantic parts from a shape collection~\cite{coalesce}, tweaking the scales and deformations of the bounding boxes~\cite{neuform,sung2020deformsyncnet}, changing the status of a coarse proxy geometry~\cite{yang2021cpcgan,lyu2021conditional}, or using natural languages~\cite{koo2022partglot}. They typically however start from an \emph{existing source shape} which is edited to a desired target.
In contrast, in addition to editing existing geometry, our method can also create \emph{novel} shapes in a \emph{controllable} manner with our introduced factorization, and enabling a wider range of applications than the prior arts.


\section{Problem Overview}
\label{sec:3}

\begin{figure*}[t]
\centering
\includegraphics[width=0.97\linewidth]{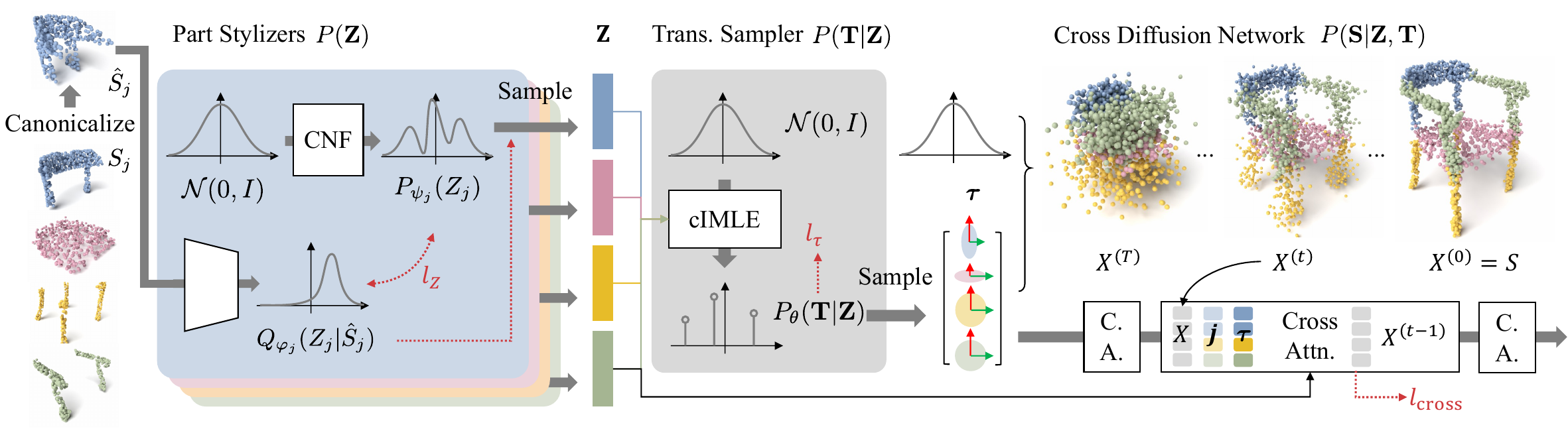}
\vspace{-0.4cm}
\caption{\textbf{Method Overview.} We factorize the 3D shape distribution into three key components, containing $m$ \textbf{part stylizers} for each part that model the shape prior $P(\mathbf{Z})$, a \textbf{transforamtion sampler} that models the conditional distribution of transformations given the part latents $P(\mathbf{T}|\mathbf{Z})$, and a \textbf{cross diffusion network} that samples the point cloud jointly considering the part geometry and their configurations $P(\mathbf{S}|\mathbf{Z},\mathbf{T})$. Red dashed lines indicate losses incorporated in the training stage. For definitions of variables, see Sec. \ref{sec:3}.}
\label{fig:overview}
\vspace{-0.7em}
\end{figure*}

Our goal is to learn a \emph{controllable} generative model on 3D shapes.
Given a set of shapes $\mathbf{S}=\{S^{(i)}\}$, 
we want to learn a distribution $P(\mathbf{S})$ from which we can sample with part-level control. 
To generate realistic and plausible shapes,  a shape \emph{prior} is commonly learned. 
Existing works~\cite{pointflow,ShapeGF,dpm, spaghetti} model the shape prior as one random variable $w \sim P(W)$, \ie a single global latent code, which \emph{does not} allow for any level of structure-aware control on the shape being sampled/generated. 



We introduce a \textbf{factorized} representation of the prior in order to obtain more localized control of the generation process.
An intuitive granularity for users to control shape generation is through a natural decomposition of shapes into semantic parts~\cite{mo2019partnet}, 
which we leverage on in our proposed factorization. 
For the rest of the paper, we assume that shapes in $\mathbf{S}$ from a given category have a predefined set of $m$ semantic parts. 

A shape $S^{(i)}$ is decomposed into its semantic parts $\{S^{(i)}_j\}_{j=1,...,m}$, where $S^{(i)}_j$ is the geometry of each shape part the superscript is omitted in what follows for brevity. 
We further factorize $S_j$ into its \emph{canonicalized geometry}, $\hat{S}_j$, \ie \textbf{part style}, and its corresponding \textbf{instancing transformation}, $\TPart_j\in \mathbb{R}^{D_1}$. 
We then model a independent distributions $P(Z_j)$ for each canonicalized part, where $Z_j \in \mathbb{R}^{D_2}$ $j=1,...,m$ is the canonicalized part latents, which we call as part styliz\emph{er}s. 
For simplicity, we use bold type for random variables and samples on the collection for all parts, e.g. $\mathbf{Z} = \{Z_j\}$, uppercase letters are random variables and lowercase letters are their corresponding samples, e.g. $z_j\sim Z_j, \tPart_j \sim \TPart_j$. Conditioned on all the part style latents $\mathbf{Z}$, we further model a distribution on their transformations, $P(\TAll|\mathbf{Z})$, where $\TAll = \{\TPart_j\}$. The shape prior distribution is then the \emph{joint} distribution of individual part styles and their transformations. Our proposed shape factorization is then given as:


\begin{equation}
\begin{split}
    P(\mathbf{S}) &= \int \int P(\mathbf{S}|\bm{z}, \tAll) P(\tAll|\bm{z})P(\bm{z})\,\dif\bm{z}\,\dif\tAll\,,
\end{split}
\end{equation}


\noindent where our shape prior is now \emph{factorized} as $P(\mathbf{Z}, \TAll) = \prod_{j=1}^m P(Z_j) P(\TAll|\mathbf{Z})$. Our factorization allows for control in our learned generative model, since we are able to independently sample from each part style latent distribution $z_j \sim P(Z_j)$ and a valid set of transformations $\bm{\tau} \sim P(\TAll|\mathbf{Z})$, thus introducing multiple different knobs for variation. We detail our joint prior $P(\mathbf{Z}, \TAll)$ in \cref{sec-joint_prior}. To capture and generate more plausible and holistic shapes, we further introduce our \emph{cross diffusion network} that models $P(\mathbf{S}|\mathbf{Z},\TAll)$, a conditional distribution of points on the shape's surface.
We elaborate our diffusion-based network and also introduce a generalized forward kernel in \cref{sec-cross_diffusion}. We showcase that our proposed factorization not only allows us to sample and generate diverse and plausible shapes, but also enables multiple shape editing and variation applications in \cref{sec-ae}. 
\vspace{-0.35cm}
\section{Method}\label{sec:4}
We learn a shape distribution $P(\mathbf{S})$ given segmented shapes, represented as a point cloud $S= \{x_k\}_{k=1,...,N}$, $x_k\in\mathbb{R}^3$ with their semantic parts $\{S_j\}_{j=1,...,m}$, where $S =\cup_{j=1}^m S_j$. Each part $S_j$ is decomposed into its part style $\hat{S}_j$ and transformation $\tPart_j$. Concretely, the part style $\hat{S}_j$ is the canonicalized part geometry given by the transformation $\tPart_j \in \mathbb{R}^6$ representing the shift $(c_j \in \mathbb{R}^3)$  and axis-aligned scale $(s_j \in \mathbb{R}_{>0}^3)$, giving us:
\begin{equation}
    S_j = \Diag(s_j)\hat{S}_j + c_j. \nonumber
\end{equation}

Our method models the data distribution $P(\mathbf{S})$ with a factorized joint prior $P(\mathbf{Z}, \TAll)$ and has three components: 
(i) independent \textbf{part stylizers} that model $P(Z_j)$ for $j = 1, ..., m$, representing part styles, 
(ii) a \textbf{transformation sampler} that models a distribution $P(\TAll|\mathbf{Z})$ of the part transformations conditioned on their styles, and 
(iii) a \textbf{cross diffusion network} that models the conditional distribution $P(S|\mathbf{Z},\TAll)$ of points on the surface of the shape given the factorized prior. \cref{fig:overview} illustrates our method overview.

\subsection{Training Objective}
We first derive the training objective for our factorized joint prior $P(\mathbf{Z}, \TAll)$. We maximize the likelihood of our learned distribution $P(\mathbf{S})$ through the evidence lower bound. We let $\bm{\psi}, \theta, \phi$ be the model parameters of our part stylizer, transformation sampler and cross-diffusion network, respectively. In addition, we assume an evidence distribution $Q_{\bm{\varphi}}\paren{\mathbf{Z}, \TAll|S}$ that can be formulated as:
\begin{equation}
\small
    \label{equ:evidence distribution}
    \begin{split}
    Q_{\bm{\varphi}}\paren{\mathbf{Z}, \TAll|S} &= Q_{\bm{\varphi}}\paren{\mathbf{Z}|S}Q\paren{\TAll|S}\\
    &= Q\paren{\TAll|S}\prod_{j=1}^mQ_{\varphi_j}\paren{Z_j|S_j}\\
    &=\prod_{j=1}^mQ_{\varphi_j}\paren{Z_j|\hat{S}_j},
    \end{split}
\end{equation}
where $Q\paren{\TAll|S}$ is deterministic since we assumed known segmentation $S = \{S_j\}$, and we assume $Z_{j_1} \perp Z_{j_2}$ $ \forall j_1, j_2 \in \set{1, \dots, m}$, $j_1 \neq j_2$, \ie $\mathbf{Z}$ and $\TAll$ are conditionally independent given $S$.
Thus, the evidence lower bound (ELBO) for the likelihood can be derived as:
\begin{equation}
\scriptsize
    \label{equ:gen-obj}
    \begin{split}
& \E_{S}\bracket{\log P_{\phi, \bm{\psi}, \theta}\paren{S}}\\ 
& = \E_{S}\bracket{\log \iint P_{\phi}\paren{S| \bm{z}, \tAll}P_{\bm{\psi}, \theta}\paren{\bm{z}, \tAll}\,\dif\bm{z}\dif\tAll}\\
& = \E_{S}\Bigg [ \log \iint \frac{P_{\phi}\paren{S| \bm{z}, \tAll}P_{\bm{\psi}}(\bm{z})P_{\theta}(\tAll|\bm{z})}{Q_{\bm\varphi}(\bm{z}, \tAll|S)}  Q_{\bm\varphi}(\bm{z}, \tAll|S)\dif\bm{z}\dif\tAll\Bigg ]\\
&\geq  \E_{S,\bm{z}} \bigg[ \underbrace{\log P_{\phi}\paren{S| \bm{z}, \tAll}}_{\mathcal{L}_{\text{recon}}}  + \underbrace{\sum_{j=1}^m\log \frac{P_{\psi_j}(z_j)}{Q_{\varphi_j}(z_j|\hat{S}_j)}}_{\mathcal{L}_{\bm{Z}}}+ \underbrace{\log P_{\theta}\paren{\tAll| \bm{z}}}_{\mathcal{L}_{\tAll}}\bigg ],
    \end{split}
\end{equation}
where the inequality is through Jensen's (see supplement). 
We now go through each of our components, and how they are trained to maximize this derived training objective with the corresponding loss functions (denoted as $l_{(\cdot)}$). 

\subsection{Factorized Joint Prior $P(\mathbf{Z}, \TAll)$}
\label{sec-joint_prior}
Our factorization allows us to define multiple random variables to control the generative model and output plausible shapes. These random variables can be independently sampled and are the `control knobs' in the generation process, thus allowing for user-controllability. These random variables are $Z_j$ to control per-part style through our part stylizers and $\TAll$ that controls the set of part transformation through our transformation sampler. 
Moreover, this also allows the network to consolidate part information, \eg the canonicalized geometry of a rectangular back of the bench is the same as a dining chair when transformation is factored out. This decomposition also enables resampling certain part geometries $Z_j$ while keeping the rest fixed, creating variations in part configurations by resampling $\TAll$, and local shape editing through encoding and modifying $z_j$ or $\tau_j$. \\

\noindent \textbf{Part Stylizer.} The part stylizer learns to model the independent distributions $P(\hat{S}_j)$ 
of part styles for $j=1,...,m$. 
Each $\hat{S}_j$ is encoded into a part latent code $Z_j$ with encoder $Q_{\varphi_j}(Z_j|\hat{S}_j)$, representing the evidence distribution. We use a continuous normalizing flow~\cite{chen2018neural} (CNF) model to learn part priors $P_{\psi_j}(Z_j)$ with parameters $\psi_j$. Formally, we have the part stylizer loss $\ell_{\mathbf{Z}}$ given by:
\begin{equation}
\small
    \begin{split}
        \ell_{\mathbf{Z}} &= \sum_{j=1}^m \KL{Q_{\varphi_j}\paren{Z_j|\hat{S}_j}}{P_{\psi_i}\paren{Z_j}}\\
        & = -\sum_{j=1}^m \E_{z_j\sim Q}\bracket{\log P_{\psi_j}\paren{z_j}}+H\paren{Q_{\varphi_j}\paren{Z_j|\hat{S}_j}}
    \end{split}
\end{equation}

\noindent where $H$ is the entropy and $P_{\psi_j}(Z_j)$ is a complex distribution transformed from Gaussian. This is equivalent to maximizing $\mathcal{L}_{\mathbf{Z}}$ in \cref{equ:gen-obj}. See supplement for more details on CNF. 
\\

\noindent \textbf{Transformation Sampler.}
Given part styles $\bm{z}$, there are different plausible configurations, \ie multiple sets of transformations, that result in valid shapes. Hence, we cannot simply regress the transformations $\TAll$ for given part styles $\bm{z}$, leading us to model a conditional \emph{distribution} of transformations $P(\TAll|\mathbf{Z})$. 
Learning this distribution is non-trivial because of three main reasons: 
(i) we have to learn a diverse set of shape variations captured only by the transformation parameters, \eg a dining chair and a bench can have the same set of part styles, 
(ii) each training example $S \in \mathbf{S}$ only provides one-to-one pairs $(\bm{z},\tAll)$ of part styles and transformations, and 
(iii) the desired conditional distribution $P(\TAll|\mathbf{Z})$ may be \emph{multimodal}. 


To satisfy these properties, we leverage on conditional Implicit Maximum Likelihood (cIMLE)~\cite{li2020multimodal} that trains an implicit generative model, $P_{\theta}(\TAll|\mathbf{Z})$ in our case, by encouraging \emph{some} generated output to match the observation from $S$, in contrast to unimodal approaches~\cite{cgan,sohn2015learning} that enforces \emph{all} generated outputs to match the observation leading to mode collapse.
Concretely, transformation sampler $T_{\theta}$ outputs samples $\tAll_k=T_\theta\paren{\bm{z}, y_k}$ for part style latents $\bm{z}$ and random latent variable $y_k\sim \mathcal{N}(0, I), y \in \mathbb{R}^{D_{\tPart}}$.
We sample multiple latents $y_1,...,y_K$ and encourage that at least one of them matches the observed data $S$. As shown in IMLE~\cite{li2018implicit}, maximizing the likelihood is then equivalent to minimizing the loss:
\begin{equation}
\label{equ:cimle_obj}
\small
\begin{split}
       & \ell_{\tAll} = \sum_{S\in\mathbf{S}} \min_{k=1, \dots, K}\ell_{\text{fit}}\paren{T_\theta\paren{\bm{z}_S, y_k}, \tAll_S},  
\end{split}
\end{equation}
\noindent where 
$\tAll_S$ is the observed part transformations for shape $S$. We define $\ell_{\text{fit}}$ as the distance between the generated set $\{\tAll_k\}$ and observation  $\tAll_S$  summed across all parts, given as
$$
\small{
\ell_{\text{fit}}\paren{\tAll, \tAll_S} = \sum_{j=1}^m \norm{c_j - c_{S, j}}^2_2 + \norm{\log s_j - \log s_{S, j}}^2_2.
}
$$

\noindent Minimizing $\ell_{\tAll}$ is equivalent to maximizing $\mathcal{L}_{\tAll}$ in \cref{equ:gen-obj}.

\subsection{Cross Diffusion Network}
\label{sec-cross_diffusion}
In order to capture plausible and holistic shapes given our proposed factorization, we model the conditional shape distribution $P(\mathbf{S}| \mathbf{Z}, \TAll)$ given part style latent codes and part transformations with our \emph{cross diffusion network}. Specifically, we represent a shape as a distribution of points on its surface. Given segmented shapes $S$ with $M$ parts 
, we model $P(x| \mathbf{Z}, \TAll, j)$ $\forall x \in \mathbb{R}^3$ as the probability that $x$ lies on the surface of part $S_j$. 
Since we use part semantic labels as an additional condition, this also allows us to output segmented shapes by specifying $j$.
Each point is treated as an independent sample (denoted by the random variable $X$), which leads to the conditional likelihood of a shape $S$:

\begin{equation}
    P\paren{S|\mathbf{Z}, \TAll} = \prod_{j=1}^m \prod_{x\in S_j} P\paren{x|\bm{z}, \tAll, j} \footnotemark{}.
\end{equation}

\footnotetext{By the same assumption of a deterministic mapping between $x$ and $j$.}

Our cross diffusion network leverages on denoising diffusion probabilistic model (DDPM) to learn the conditional likelihood $P_\phi(\mathbf{Z}, \TAll, j)$ 
through an iterative denoising process. DDPM models a probability distribution using a reverse process which is a Markov chain with a fixed prior, and to learn DDPM, we approximately maximize the likelihood with the forward process as the approximate posterior. Instead of directly maximizing each conditional likelihood~\cite{dpm, diffusionho}, we reparameterize learn $\varepsilon_{\phi}$ that predicts the noise at a timestep $t$ given the noisy data point $x^{(t)}$ and minimize the distance between the predicted noise $\varepsilon_{\phi}$ and the ground truth noise $\varepsilon$. Thus maximizing $\mathcal{L}_{\text{recon}}$ in \cref{equ:gen-obj} is equivalent to minimizing
\begin{equation}\label{equ:l-recon}
    \small
\ell_{\text{cross}} = \sum_{j=1}^m\sum_{x\in S_j}\E_{\varepsilon, \bm{z}, t\footnotemark}\bracket{\norm{\varepsilon - \varepsilon_\phi\paren{x^{(t)}, \bm{z}, \tAll_S, j, t}}^2_2}, 
\end{equation}
\footnotetext{$\varepsilon \sim \mathcal{N}\paren{\bm{0}, \bm{I}}, \bm{z}\sim Q_{\bm{\varphi}}\paren{\bm{Z}|\hat{S}}, \text{ and } t \sim \Uniform\set{1, \dots, T}$}%
where the predicted noise $\varepsilon_\phi$ is conditioned on part styles $\bm{z}$ 
, transformations $\tAll$, semantic label $j$ and timestep $t$. Our cross diffusion network uses our introduced \emph{generalized} forward kernel that allows for the preservation of information from part transformations ($\tPart_j = (c_j, s_j)$) in the forward diffusion process while only adding noise to the canonicalized part geometries, \ie part styles. We also show the reverse process of our generalized forward kernel can also be similarly reparameterized arriving at the same loss in \cref{equ:l-recon}. \cref{fig:overview} shows one example for the forward diffusion process. Our experiments show that this modification allows for better reconstruction quality and part transformation extrapolation. 

We use a cross-attention network with $L$ cross attention layers to instantiate $\varepsilon_\phi$. For a timestep $t$, the network predicts $x^{(t-1)}$ conditioned on $(x^{(t)}, j, \tPart_j)$. The input to each cross attention layer attends to $m$ tokens each being the concatenation of $(z_j, \tPart_j, j, t)$, for $j=1, ..., m$. This design allows a point $x^{(t)}$ to be informed of both the global shape through the $m$ tokens and the local part by concatenating the coordinate $x^{(t)}$ with its corresponding part label and transformation, enabling us to capture and generate more plausible and holistic shapes. \\

\begin{figure*}[t]
\centering
\includegraphics[width=\linewidth]{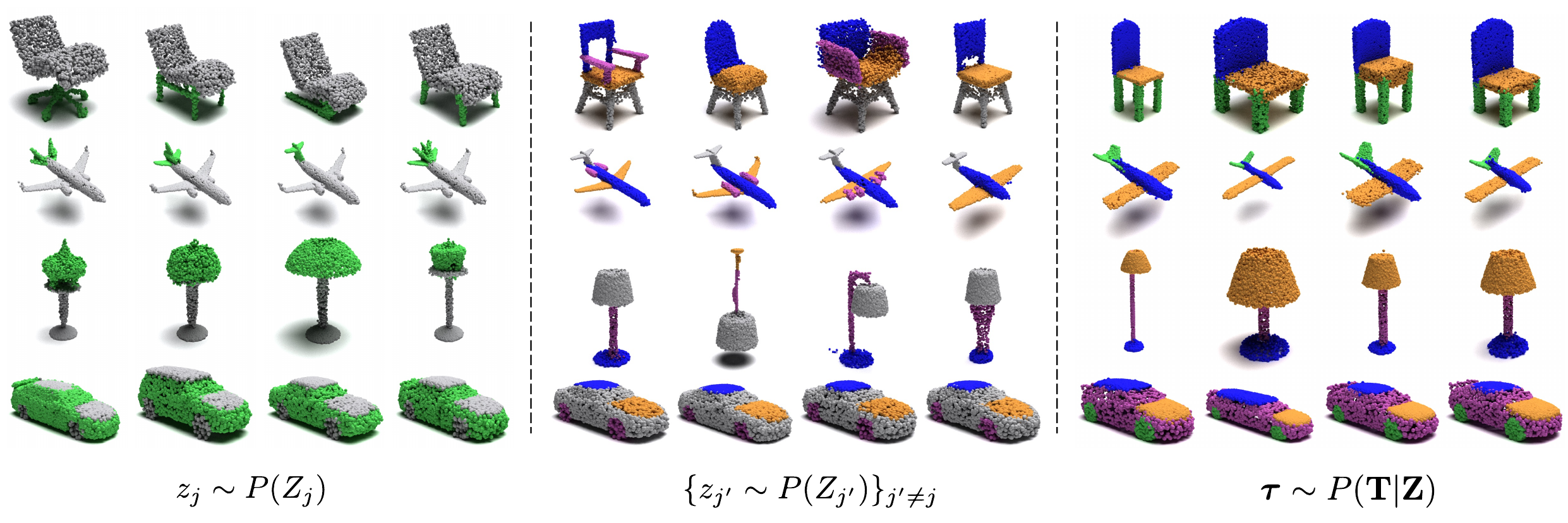}
 \vspace{-0.7cm}
\caption{ \textbf{Generated Shapes with Controlled Variation}. (Left) Re-sampling a selected part style while keeping the rest fixed. (Middle) Fixing a selected part while re-sampling the rest. (Right) Generating multiple part configurations for a given set of part styles. Gray refers to the fixed part while colored parts are being modified.  \vspace{-0.4cm}
}
\label{fig:controlled-variation}
\end{figure*}

\noindent \textbf{Generalized Forward Kernel.} Finally, we introduce a generalization of the forward kernel used to add noise $\varepsilon$ to the data points $x$. Existing works~\cite{dpm, Zhou_2021_pvd} use a forward kernel that diffuses all points on the surface to the standard unit Gaussian. We show that our generalized forward kernel is theoretically equivalent to diffusing all points to a \emph{scaled} and \emph{shifted} Gaussian. Our modification allows to incorporate an additional prior (scale and shift) to the forward process. Specifically, for a $d$ dimensional diffusion process, a shift  $\mu \in \R^d$ and variance $\Sigma \in \R^{d\times d}$ parameter is incorporated into the forward kernel so that it becomes:
\begin{equation}
    \label{equ:fk}
\begin{split}
    &Q\paren{X^{(t)}| x^{(t-1)}, \mu, \Sigma}\\
    &= \mathcal{N}\paren{\sqrt{\alpha_t}x^{(t-1)} + \paren{1 - \sqrt{\alpha_t}}\mu, \paren{1 - \alpha_t}\Sigma},
\end{split}
\end{equation}
for $t = 1, \dots, T.$ Here $\alpha_t$'s are variance schedule hyperparameters. As $T \to \infty,$ we show that the final distribution approaches a parameterized Gaussian with mean $\mu$ and variance $\Sigma$ (see supplement for the full proof): 
\begin{dmath}
    Q\paren{X^{(T)}| x^{(0)}, \mu, \Sigma}
    = \mathcal{N}\paren{\sqrt{\overline{\alpha}_T}x^{(0)} + \paren{1 - \sqrt{\overline{\alpha}_T}}\mu,
    (1 - \overline{\alpha}_T)\Sigma} \cong \mathcal{N}\paren{\mu, \Sigma}.
\end{dmath}
 Here $\overline{\alpha}_T = \prod_{t=1}^T\alpha_t$. Note that the standard forward kernel is a special case of our generalization by setting $\mu = \bm{0}$ and $\Sigma = \diag(\bm{1})$. For our task, we set $\mu = c_{j} \text{ and } \Sigma = \Diag(s_{j}^2)$ for points $x\in S_j$. \\
\noindent Please see supplement for more details and full derivations. The total loss is then $\ell_{\text{total}} = \ell_{\text{recon}} + \lambda_1\ell_{\mathbf{Z}} + \lambda_2\ell_{\tAll} $.

\subsection{Enabling Shape Editing}
\label{sec-ae}

Our method does not only allow for shape generation through \emph{sampling} from individual part style $P(Z_j)$ and transformation $P(\TAll|\mathbf{Z})$ distributions, but it is also able to \emph{encode} specified parts and \emph{modify} them, allowing for local shape editing and controllable variation synthesis. We highlight that enabling local edits/variations is non-trivial as there is a tension between \emph{preservation} and \emph{adaptation} post-edit. In other words, an ideal edit would keep the unmodified parts unchanged as much as possible, but at the same time still adapt the changes required to maintain a plausible shape after incorporating the specified edit.

1) In the simplest setting, our framework can be trained as an \emph{autoencoder} by deterministically regressing each part latent $Z_j$ and directly feeding the part transformations $\tAll_{S}$ for a shape $S$. The training objective is simply equivalent to $\ell_{\text{recon}}$. 2) A step further from the autoencoder set-up is to \emph{synthesize variations} of a shape by training the transformation sampler. Given $S$, we deterministically encode the part styles, and then either a) sample different transformations $\tAll$, or b) manually edit selected part transformations $\tPart_j$ from $\tPart_{S}$, achieving variations on (part) transformation configurations while keeping part styles fixed. This set-up also allows for shape mixing-and-matching with transformation variations. 3) Moreover, we also enable \emph{local} editing where specified part $j$ of a shape can change while keeping the rest fixed that is done by either a) resampling the corresponding part latent code from $P(Z_j)$ or b) interpolating between part styles. The applications are shown in the results section.

\section{Results}

\subsection{Dataset and Evaluation Protocol} 
We use four classes from ShapeNet~\cite{shapenet2015} dataset: chair, airplane, lamps, and cars. We train/test the networks per object class with the splits provided by~\cite{Yi16}. Each category contains 3053, 2349, 1261, 740 training shapes and 704, 341, 286, 158 test shapes, respectively. The semantic labels for all classes come from~\cite{Yi16}. See supplement for implementation details — network architecture, training time, etc.

For our task on controllable part-based generation, we propose to measure intra-part and inter-part level scores. For more details, please refer to the supplement. 

\begin{table}[t]
\centering
\small
\setlength{\tabcolsep}{7pt}
\begin{tabular}{l|ccc}
    \toprule
 Chair  &  MMD-P ($\downarrow$) &  COV-P ($\uparrow$) &  1NNA-P \\ 
   \midrule
   PointFlow~\cite{pointflow}  & 4.68 & 27.3 & 87.77   \\
   DPM~\cite{dpm}  & 4.17 & 28.2 & 85.65 \\
   ShapeGF~\cite{ShapeGF}  & 3.52 & 42.3 & 68.65 \\
   LION~\cite{lion}  & 3.99 & 35.1 & 69.25 \\
   \midrule
   \textbf{\methodname} (Ours)  & \textbf{3.27} & \textbf{42.5} & \textbf{65.23} \\ \bottomrule
\end{tabular}\\
\vspace{0.5em}
\setlength{\tabcolsep}{7pt}
\begin{tabular}{l|ccc}
    \toprule
  Airplane &  MMD-P ($\downarrow$) &  COV-P ($\uparrow$) &  1NNA-P \\ 
   \midrule
   PointFlow~\cite{pointflow}  & 4.61 & 32.0 &86.11  \\
   DPM~\cite{dpm}  & 3.52 & 37.7 &78.74 \\
   ShapeGF~\cite{ShapeGF} & 3.50 & 40.0 & 72.04 \\
   LION~\cite{lion} & 3.68 & 38.8 & 68.73\\
   \midrule
   \textbf{\methodname} (Ours)  & \textbf{3.20} & \textbf{46.2} & \textbf{68.72} \\ \bottomrule
\end{tabular}
\vspace*{-0.5\baselineskip}
\caption{\textbf{Global Shape Code Baselines.} MMD-P score is multiplied by $10^{-2}$. COV-P and 1NNA-P are reported in $\%$.
\vspace{-1cm}}
\label{tab:global-code-comp}
\end{table}

\parahead{Intra-part Score} We evaluate the quality of individual part distributions $P(S_j)\forall j$ 
using the standard generation metrics following~\cite{pointflow}: minimum matching distance (MMD-P), coverage (COV-P) and 1-NN classifier accuracy\footnotemark{}(1NNA-P), measuring the similarity between the distributions of canonicalized parts of the generated shapes compared to the test set from~\cite{Yi16} of segmented shapes, where the score is computed for each part and averaged across all \emph{P}arts (hence the suffix `-P').
\footnotetext{Arrow for 1NNA-P is left out in the table because an optimal score for this metric is 50 \%.}

\parahead{Inter-part Score} We also measure the local part-to-part coherence of the generated shape as having control in individual parts may result in disjoint or incoherent outputs. We use a snapping metric (SNAP) that measures local connectivity between independently generated parts, given as the Chamfer distance (CD) between the $N_{\text{snap}}$-closest points between connected parts. We report the average score across all connections of all generated samples.

\parahead{Plausibility} We also further evaluate the plausibility of our generated shapes by using them to augment training datasets for point cloud segmentation. The idea is if our approach generates \emph{novel} and coherent shapes with part labels then using them for data augmentation would improve the segmentation score of part segmentation networks.

\parahead{Human Study} We also conduct a user study to evaluate controllability, methods being evaluated each generate a triplet of edited shapes, and users are asked to select the triplet containing shapes that are most plausible, where an `abstain' option can be selected. An edit for a given shape is defined as resampling a pre-selected parts to output a novel shape where plausibility is measured.

\subsection{Baselines}

\parahead{Global Shape Code} Existing point cloud generation works do not explicitly model individual parts as they sample from a single \emph{global} shape distribution, which \emph{do not} provide individual part-level control. To evaluate individual part distributions and measure the intra-part score, we use a pretrained part segmenter~\cite{qi2017pointnetplusplus} to decompose generated shapes into parts. We compare against recent works PointFlow~\cite{pointflow}, DPM~\cite{dpm}, ShapeGF~\cite{ShapeGF} and LION~\cite{lion} on intra-part level scores. Inter-part scores are not measured as these works directly generate a full shape.

\begin{table}[t]
\centering
\small
\setlength{\tabcolsep}{6pt}
\begin{tabular}{l|ccc}
\toprule
                    & Ctrl-ShapeGF & Ctrl-LION & \textbf{\methodname} (Ours) \\ \midrule
SNAP ($\downarrow$) & 41.12                                                   & 31.76                                                & \textbf{13.32}                                                        \\ \bottomrule
\end{tabular}
\vspace*{-0.5\baselineskip}
\caption{\textbf{Control-Enabled Baselines.} Snapping metric on three connections for chairs: back to leg or seat; seat to legs; and arms to seat or back. (CD $\times 10^{-2}$).
\vspace{-0.4cm}}
\label{tab:controllable-comp}
\end{table}


\parahead{Control-Enabled} We also introduce new baselines that allow for part-level control. We modify ShapeGF~\cite{ShapeGF} and LION~\cite{lion} to have part-level control (prefixed by ``Ctrl-'') by modeling independent part distributions, \ie $P(\textbf{S})=\prod_j P(S_j|w_j)$
, where each part latent distribution is modeled with a (hierarchical) variational encoder $Q(w_j|S_j)$. These baselines allow for independent sampling at the part-level unlike existing approaches. We measure the inter-part score for these baselines to evaluate the coherence of the generated shape. Intra-part scores are not measured as these baselines are individually trained per part.




\subsection{Controllable Shape Generation}
\label{sec:exp_generation}
\cref{fig:controlled-variation} shows shapes generated by sampling from different components in our factorization enables both control in part style and part configurations: Our approach is able to \emph{sample} (left) or \emph{fix} (middle) a specified part, and we are also able to generate various plausible configurations of the shape (right) with fixed part styles.


\parahead{Comparison with Global Shape Code Baselines} 
\cref{tab:global-code-comp} shows the intra-part scores of our approach compared to the global shape baselines, showcasing that our approach learns better individual part-level distributions.

\parahead{Comparison with Control-Enabled Baselines} 
\cref{tab:controllable-comp} shows the inter-part scores of our approach compared to the control-enabled baselines. We output more coherent shapes than the baselines that naively enable part-level control without the modeling of part relationships

\begin{table}[t]
\centering
\small
\setlength{\tabcolsep}{8pt}
\begin{tabular}{l|c|c|c}
    \toprule
   &  Orig. & $+$ Multi (700) & $+$ Control (60)    \\
   \midrule
PointNet & 0.709 &\textbf{0.788} & \underline{0.780}  \\
PointNet++ & 0.800 & \textbf{0.808} & \underline{0.801}     \\
\bottomrule
\end{tabular} 
\vspace*{-0.5\baselineskip}
\caption{\textbf{Plausibility score.} mIOU on the cars from~\cite{Yi16} trained original and augmented datasets.}
\label{tab:sem-seg}
\end{table}


\parahead{Plausibility} We use the car category containing the least training data originally with 704 shapes (Orig.), and augment it with 700 randomly generated shapes by \methodname ($+$ Multi). Moreover, we test our capability for control by augmenting with only 60 (controlled) race cars, with very few examples in the original training data. \cref{tab:sem-seg} shows that in both cases off-the-shelf part segmentation networks improve by augmenting the training set with our generated shapes. 

\parahead{Human Study} Our human study has 100 participants comparing our approach with the control-enabled baselines, \ie Ctrl-ShapeGF and Ctrl-LION. 
We drew 10 shapes from each of the methods with randomly selected parts to edit, and on average the participants favour \textbf{85\%} (8.5 out of 10) of our generated shapes more than other baselines.

\subsection{Shape Editing}
\label{sec:exp_shape_editing}
We demonstrate that our controllable generation approach also allows for various shape-editing applications.



\begin{figure}[t]
\begin{center}
   \includegraphics[width=\linewidth]{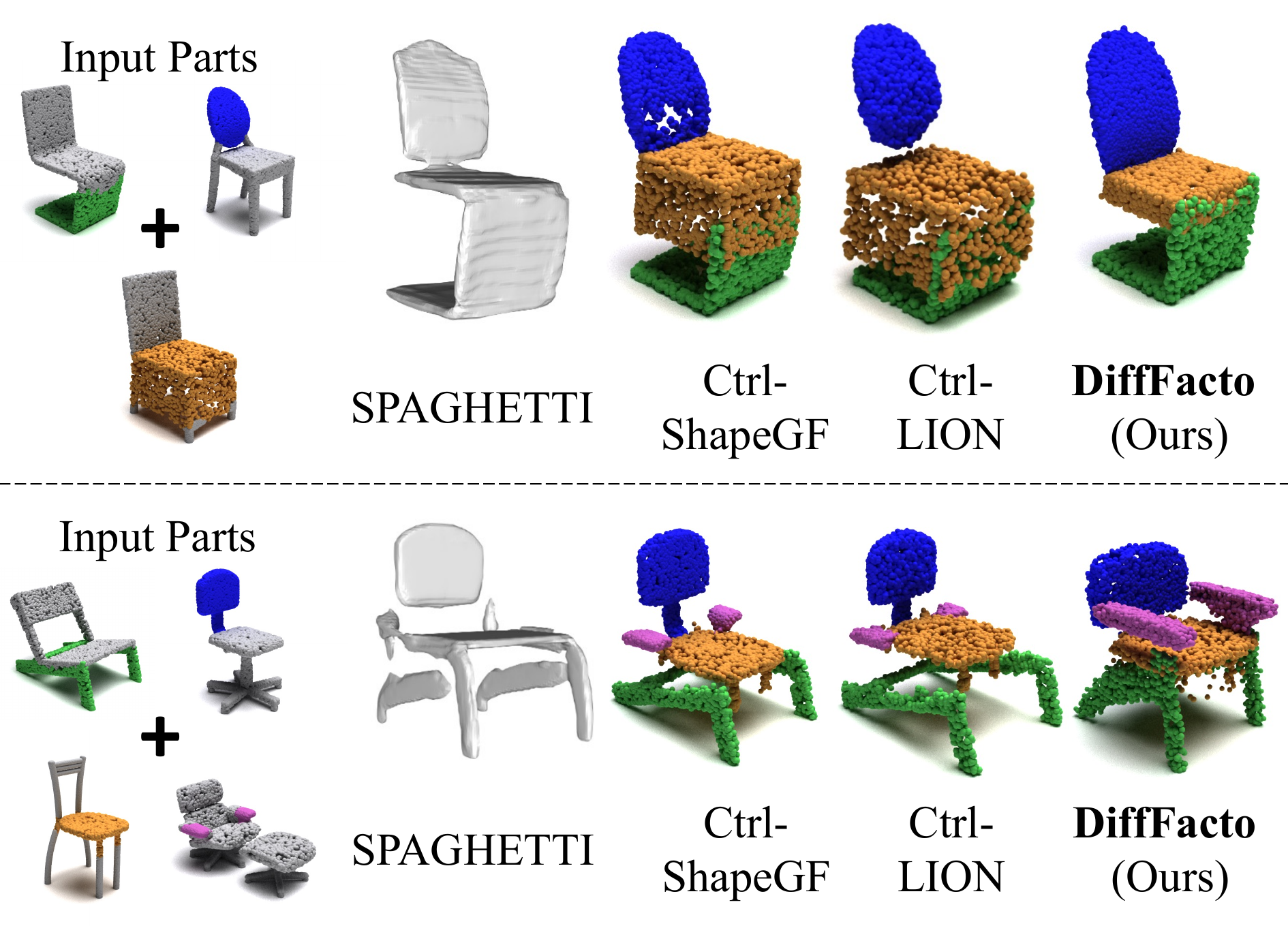}
\end{center}
\vspace{-0.4cm}
   \caption{\textbf{Part Style Mixing.} The colored parts from the left are selected and mixed to provide the shapes on the right.}
   \vspace{-0.4cm}
\label{fig:mixing}
\end{figure}

\begin{figure}[t]
\begin{center}
   \includegraphics[width=\linewidth]{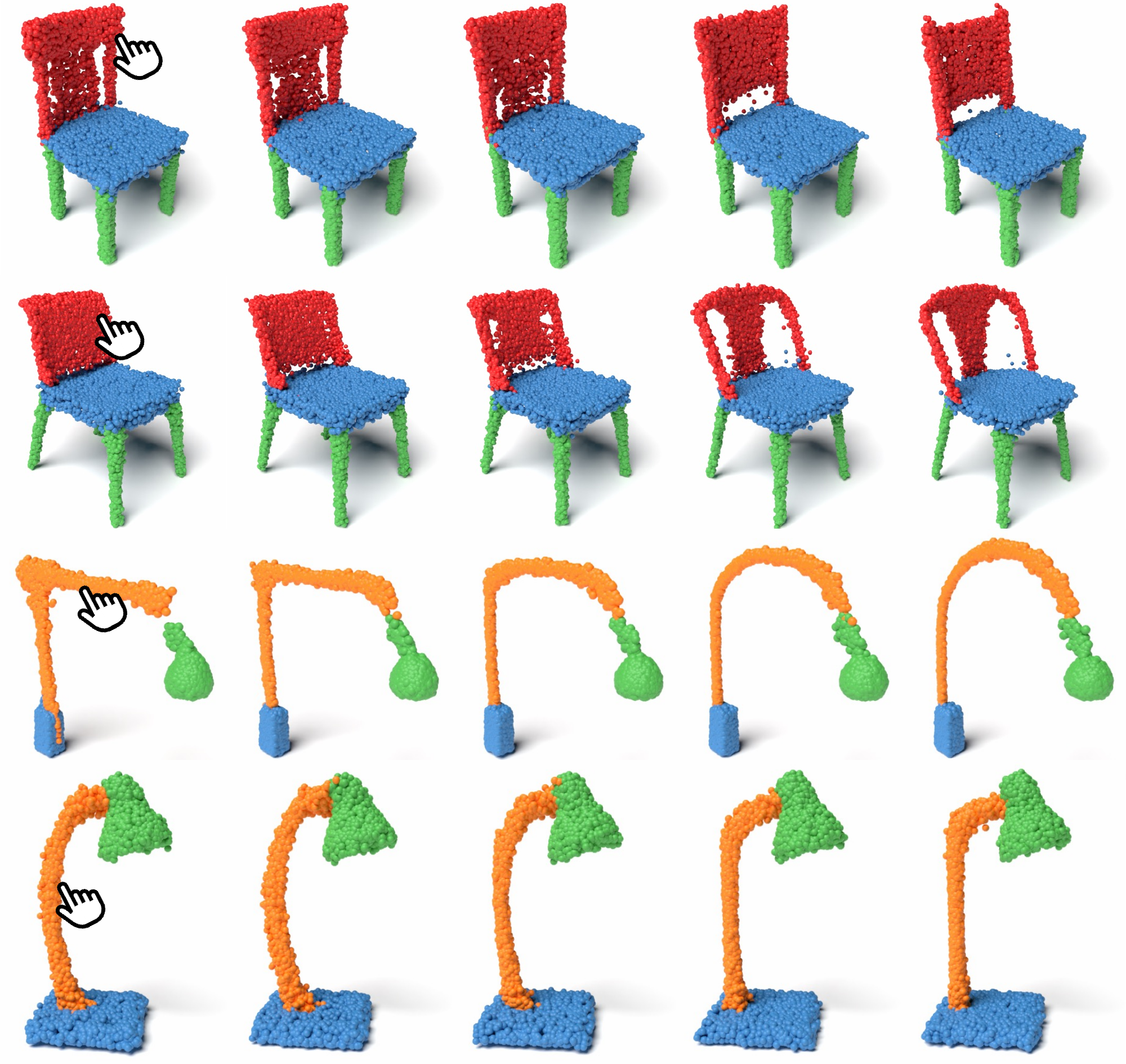}
\end{center}
\vspace*{-\baselineskip}
\caption{\textbf{Part Interpolation}. We interpolate the chair backs (red) in the first 2 rows and the lamp poles (orange) in the last 2 rows (indicated by the hand icons).}
\label{fig:part-interp}
\vspace{-0.5cm}
\end{figure}

\parahead{Part Style Mixing} 
We showcase our ability for part style mixing in \cref{fig:mixing}. We compare with SPAGHETTI~\cite{spaghetti}, an implicit-based shape editing work, as well as our control-enabled baselines Ctrl-ShapeGF and Ctrl-LION\footnote{We note that SPAGHETTI~\cite{spaghetti} requires mesh supervision.}. As shown, when selecting a combination of parts from different shapes, our approach generates a novel and coherent shape from different input parts, compared to other approaches that are unable to adapt the parts to produce a plausible output.


\parahead{Part-level Interpolation} 
\cref{fig:part-interp} qualitatively shows our part-level interpolation performance, where for each example we interpolate only one selected part latent $z_j$.
Thanks to our factorized probabilistic formulation, we are able to interpolate only the selected part while keeping the geometry of the other parts unchanged.
In the meantime, the \emph{transformations} of the other parts are automatically adapted to make the shape globally coherent.

\begin{figure}[t]
\begin{center}
\includegraphics[width=\linewidth]{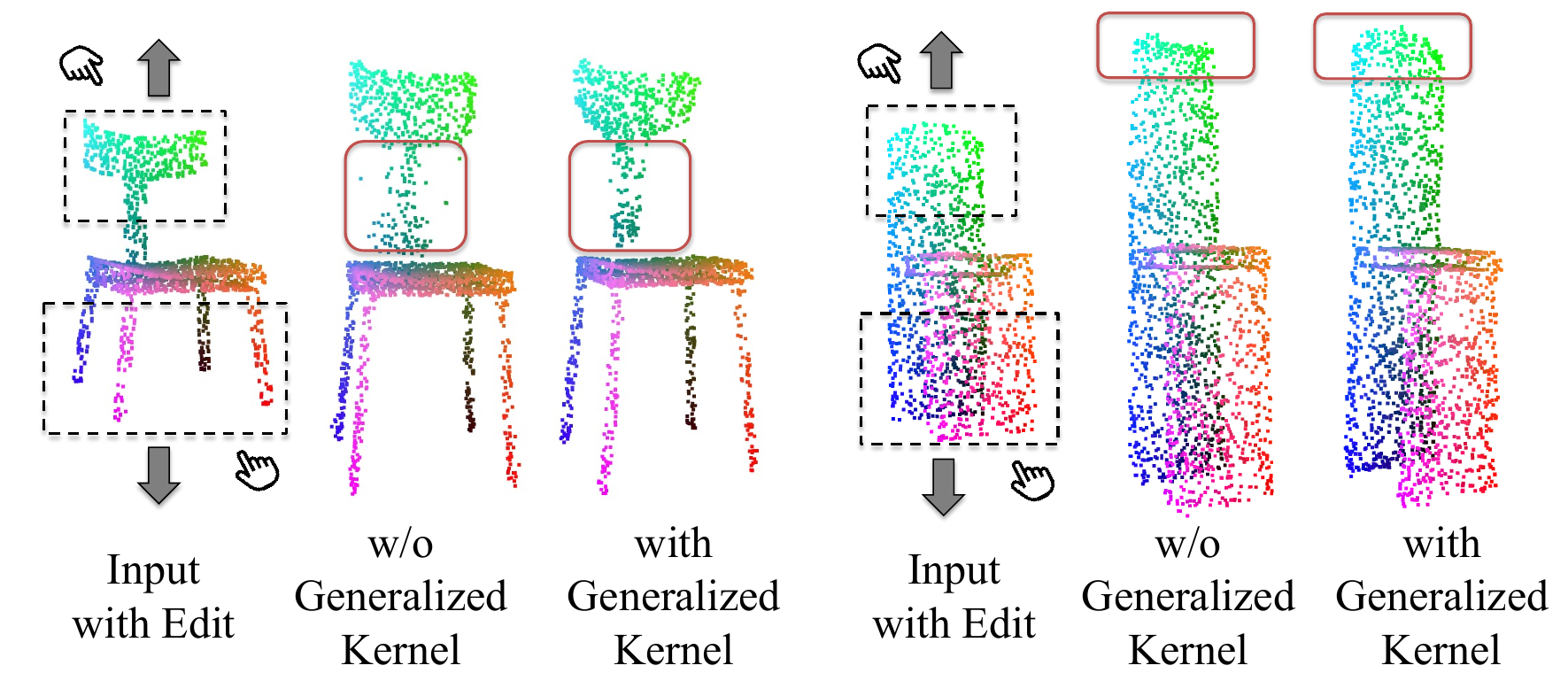}
\end{center}
\vspace{-1em}
   \caption{\textbf{Part Editing.} In both examples the user stretches the lengths of the chair backs and legs by modifying the corresponding transformations.}
\label{fig:editing}
\vspace{-0.7cm}
\end{figure}
\parahead{Transformation Editing}
Our approach also enables direct user editing on the part transformations $\tau_j$ for a selected part $S_j$. We directly optimize $y$ to find $\bm{\tau}$ that satisfies the edit while still traversing along the space of valid part configurations. 
As \cref{fig:editing} (left) shows, elongating the chair back retains its thin geometric structure while still keeping the shape plausibility.


\subsection{Ablations}
\label{sec:exp_ablations}


\begin{table}[t]
\centering
\small
\setlength{\tabcolsep}{6pt}
\begin{tabular}{c|cccc}
\toprule
                    & \textit{Separate} & \begin{tabular}[c]{@{}c@{}}\textit{Post}\\\textit{Transform}\end{tabular} & \begin{tabular}[c]{@{}c@{}}\textit{Global}\\\textit{Agnostic}\end{tabular} & \begin{tabular}[c]{@{}c@{}}\textbf{\methodname}\\ (Ours)\end{tabular} \\ \midrule
SNAP ($\downarrow$) & 25.24             & 18.23                                                    & 19.29                                                     & \textbf{13.32}                                                        \\ \bottomrule
\end{tabular}
\vspace*{-0.3\baselineskip}
\caption{\textbf{Ablation on our Factorization.} Snapping metric on three connections for chairs: back to leg or seat; seat to legs; and arms to seat or back. (CD $\times 10^{-2}$).}
\label{tab:ablation}
\vspace{-0.3cm}
\end{table}

\parahead{Factorized Joint Prior} 
We ablate our joint factorized prior $P(\mathbf{Z}, \mathbf{T})$ by replacing it with separate independent part distributions (\emph{Separate}). Each part distribution $P(S_j)$ is modeled with a separate CNF prior. \cref{tab:ablation} shows that our approach achieves better inter-part score.


\parahead{Cross Diffusion} 
We also evaluate several variants of our cross diffusion network, termed as follows: 
\emph{Post Transform} -- we remove the cross diffusion network and instead model the conditional likelihood of points only on part styles, then separately applying the part transformations as a post-processing step. 
\emph{Global Agnostic} -- we remove the cross attention component that provides global shape information through the $m$ tokens, and model only the point distribution $P(X|z_j, \tau_j, j)$ for each part. 
\cref{tab:ablation} shows that our approach also achieves better inter-part level score.
\vspace{-0.3cm}

\begin{table}[t]
\centering
\small
\setlength{\tabcolsep}{6pt}
\begin{tabular}{c|c|c|c}
    \toprule
Direct reg. & cVAE~\cite{sohn2015learning} & cGAN~\cite{cgan} & cIMLE~\cite{li2020multimodal} (Ours) \\
\midrule
13.38 & 7.33 & 11.48 & \textbf{4.97} \\
\bottomrule
\end{tabular}
\vspace*{-0.5\baselineskip}
\caption{\textbf{Multi-modality of Transformation Sampler.} Shape inversion on the chair category (CD $\times 10^{-4}$).}
\label{tab:ablation-cimle}
\vspace{-1em}
\end{table}
\begin{figure}[h]
\centering
\begin{center}
   \includegraphics[width=0.9\linewidth]{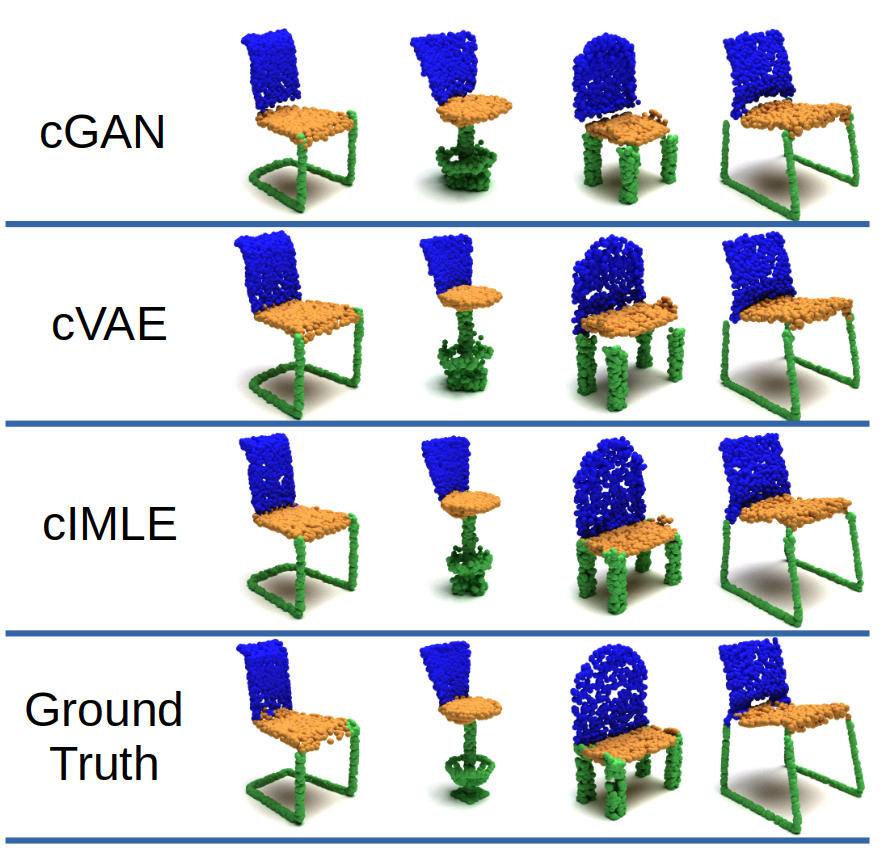}
\end{center}
\vspace{-0.4cm}
   \caption{\textbf{Shape Inversion Comparison.} We show shape inversion examples using different implicit probabilistic methods to model the transformation distribution given part style latents.}
   \vspace{-0.4cm}
\label{fig:inversion-comp}
\end{figure}

\parahead{Transformation Sampler} 
We ablate our transformation sampler that uses cIMLE~\cite{li2020multimodal} and compare it with direct regression of $\bm{\tau}$, as well as unimodal cVAE~\cite{sohn2015learning} and cGAN~\cite{cgan} on shape inversion. 
\cref{tab:ablation-cimle} quantitatively shows that our approach achieves the best results with a large margin and \cref{fig:inversion-comp} shows examples of inverted shapes using different implicit methods. Notice that across all examples, cIMLE is able to recover most accurately the correct transformation through inversion because it models a multimodal distribution where different modes can be recovered during the sampling process. 
\vspace{-1em}


\begin{figure}[h]
\centering
\begin{center}
   \includegraphics[width=\linewidth]{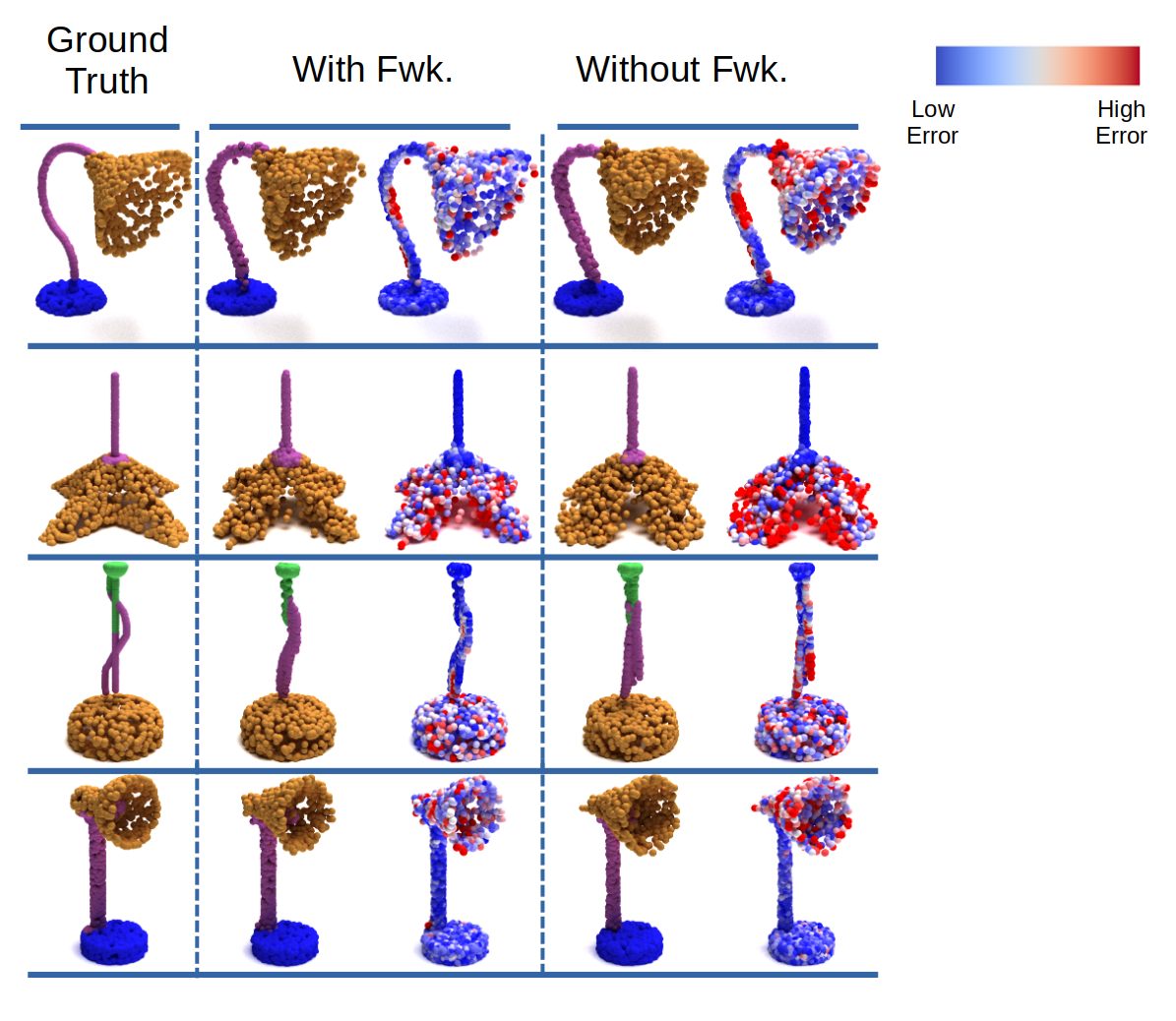}
\end{center}
    \vspace{-0.6cm}
   \caption{\textbf{Qualitative Comparison w. Generalized Forward Kernel.} Examples of reconstructed lamps with our generalized kernel \textbf{(With Fwk.)} versus without the generalized kernel \textbf{(Without Fwk.)}  For each example, we show the reconstructed shape \textbf{(Ground Truth)} where each point is colored by the minimum distance to points in the ground truth shape.}
   \vspace{-0.3cm}
   \label{fig:recon}
\end{figure}
 
\parahead{Generalized Forward Kernel} 
Our generalized forward kernel works by modeling the diffusion prior as a transformed Gaussian distribution that captures informative positional and scale information of the part.
\cref{fig:editing} shows that such a kernel allows better transformation extrapolation, where geometry is better preserved on extreme user edits. Moreover, \cref{fig:recon} shows qualitative examples of the advantages of using our generalized forward kernel as illustrated by the heat map on the per-point reconstruction error. We note that our generalized forward kernel is able to model complex part geometry better than the standard forward kernel (see the cap on rows 1, 2, and 4 in the figure) because of the additional size and location prior information incorporated into the diffusion process. 

\begin{figure}[t]
\centering
\begin{center}
\includegraphics[width=\linewidth]{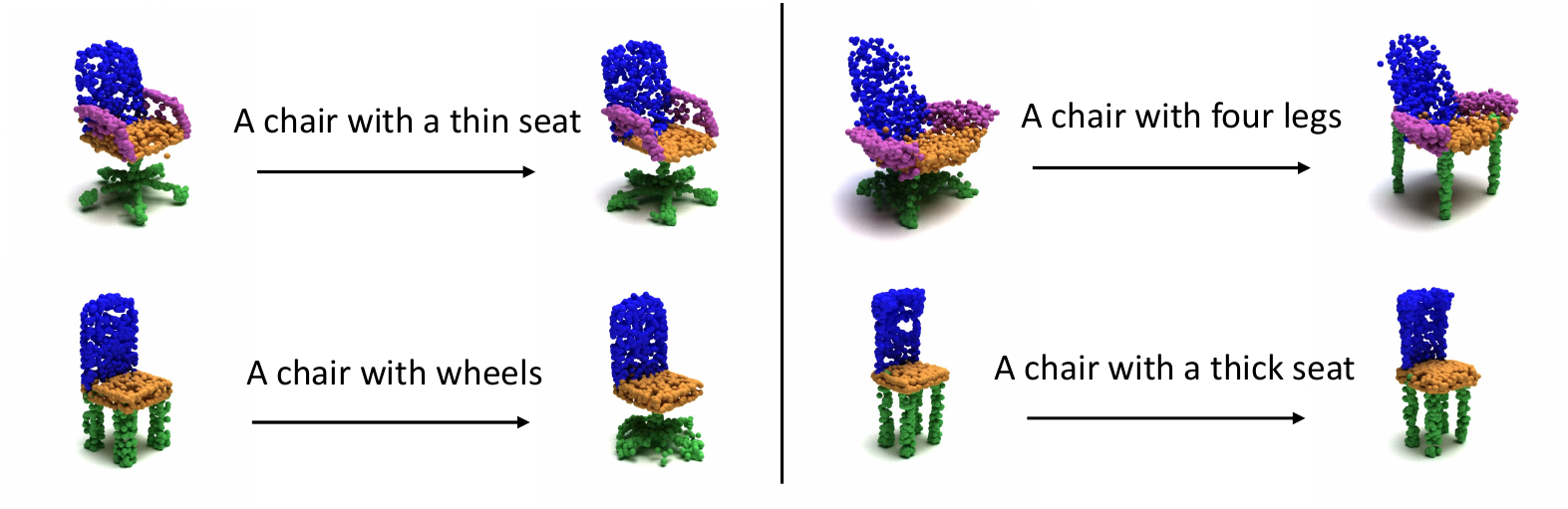}
\end{center}
\vspace{-0.7cm}
   \caption{\textbf{Potential Language Guided Edits.} We show prototypes of potential future work where we can edit existing shapes via our part style latent distributions with language guidance. The left sides are input shapes and we update the part latent vectors based on the language inputs. The resulting shapes are shown on the right side of the arrows.}
   \vspace{-0.4cm}
\label{fig:language-fig}
\end{figure}

\section{Future Works and Limitations}
\label{sec:future_work}
Our method by design requires segmented shapes for training, as our cross diffusion network and generalized forward kernel require a hard assignment of points to the semantic part of the shape, \ie it conditions on label $j$. Although this allows us to learn smooth latent spaces for each part, as well as global shape distribution $P(\mathbf{S})$, it constrains us in training with datasets that are semantically labeled. A future direction would be a formulation that enables soft assignments, which would allow us to train on unsegmented data .

Another promising future direction is to enable language-driven editing through our proposed factorization. Our factorization allows for localized edits, where language can be used to identify which semantic part, $j$, to edit as well as the direction to modify its corresponding latent code, $Z_j$, to adhere with the language description. As a preliminary demonstration, we use the dataset from ShapeGlot~\cite{shapeglot} that consists of a triplet of shapes with a corresponding language description.  We then select a negative shape and edit it using \methodname in order to match with the corresponding text description. A match is enforced by optimizing the edit to match the positive example in the triplet. \cref{fig:language-fig} shows some visual examples.

\section{Conclusion}

In this paper we propose \methodname{}, a deep generative probabilistic framework for generating 3D point clouds in a controllable manner.
By factorizing out the shape distribution into individual semantically-meaningful parts plus their transformations, we allow for intuitive control over the generated shapes.
The framework is also flexible and readily available for various applications ranging from style mixing to configuration editing.


\vspace{-0.4cm}
\paragraph{Acknowledgements.} This work is supported by ARL grant W911NF-21-2-0104, a Vannevar Bush Faculty Fellowship, the Natural Science Foundation of China (Project No. 62220106003), Tsinghua-Tencent Joint Laboratory for Internet Innovation Technology, and the Natural Sciences and Engineering Research Council of Canada. We very much appreciate Congyue Deng for her helpful discussion at the early stage of this project. We are also grateful for the advice and help from Colton Stearns and Davis Rempe.
\onecolumn
\SupplementaryMaterials
\centering
{\LARGE\bfseries Supplementary Materials}

\vspace{2em}

This document supplements our submission \methodname: Controllable Part-Based 3D Point Cloud Generation with Cross Diffusion. In particular, we provide detailed derivations and proofs of our training objective (Sec~\ref{sec:obj-derivation}) and generalized forward kernel (Sec~\ref{sec:gen_kernel}), additional experiment comparisons and results (Sec~\ref{sec:additional_experiments}), and additional details on our network (Sec~\ref{sec:network}), implementation (Sec~\ref{sec:implementation}) and experiment set-ups (Sec~\ref{sec:experiment_setup}).

\tableofcontents



\section{Objective Derivation}
\label{sec:obj-derivation}
We derive the evidence lower bound presented in Sec. 4.1 of our main paper. Let $Q_{\bm{\varphi}}(\mathbf{Z}, \TAll|S)$ be a variational family with the following factorization
\begin{equation}
    Q_{\bm{\varphi}}\paren{\mathbf{Z}, \TAll|S} = Q_{\bm{\varphi}}\paren{\mathbf{Z}|S}Q\paren{\TAll|S} = Q\paren{\TAll|S}\prod_{j=1}^mQ_{\varphi_j}\paren{Z_j|S_j} =Q\paren{\TAll|S}\prod_{j=1}^mQ_{\varphi_j}\paren{Z_j|\hat{S}_j},
\end{equation}
where we learn independent $Q_{\varphi_j}\paren{Z_j|\hat{S}_j} \forall j$, \ie our part stylizers, as a variational encoder that is parameterized as a Gaussian distribution with learnable mean and diagonal variance. 
Since given an input segmented shape $S$, its corresponding part transformations $\mathbf{T}$ is known, then the distribution $Q\paren{\TAll|S}$ is deterministic.
Furthermore, we highlight that each of the encoders $Q_{\varphi_j}$ only takes in the canonicalized part $\hat{S}_j$ as input, which enables us to model the prior distribution $P(Z_j)$ to only encode information about the part style and not of its size and scale. 
With this variational family, we can write the evidence lower bound as:

\begin{equation}
\begin{split}
&\E_{S}\bracket{\log P_{\phi, \bm{\psi}, \theta}\paren{S}}\\
& = \E_{S}\bracket{\log \iint P_{\phi}\paren{S| \bm{z}, \tAll}P_{\bm{\psi}, \theta}\paren{\bm{z}, \tAll}\,\dif\bm{z}\dif\tAll}\\
& = \E_{S}\Bigg [ \log \iint \frac{P_{\phi}\paren{S| \bm{z}, \tAll}P_{\bm{\psi}}(\bm{z})P_{\theta}(\tAll|\bm{z})}{Q_{\bm\varphi}(\bm{z}, \tAll|S)}  Q_{\bm\varphi}(\bm{z}, \tAll|S)\dif\bm{z}\dif\tAll\Bigg ]\\
& = \E_{S}\Bigg [ \log \iint \frac{P_{\phi}\paren{S| \bm{z}, \tAll}P_{\theta}(\tAll|\bm{z})\prod_{j=1}^mP_{\psi_j}(z_j)}{Q\paren{\bm\tau|S}\prod_{j=1}^mQ_{\varphi_j}\paren{z_j|\hat S_j}}  Q\paren{\bm{\tau}|S}\prod_{j=1}^mQ_{\varphi_j}\paren{z_j|\hat S_j}\dif\bm{z}\dif\tAll\Bigg ]\\
& = \E_{S}\bracket{ \log\E_{z_j\sim Q_{\varphi_j}, \bm{\tau}\sim Q(\bm{T}|S)}\bracket{\frac{P_{\phi}\paren{S| \bm{z}, \tAll}P_{\theta}(\tAll|\bm{z})\prod_{j=1}^mP_{\psi_j}(z_j)}{Q\paren{\bm\tau|S}\prod_{j=1}^mQ_{\varphi_j}\paren{z_j|\hat S_j}}}}
\shortintertext{By Jensen's Inequality,}
&\geq \E_{S\sim P(S),z_j\sim Q_{\varphi_j}, \bm\tau\sim Q(\bm{T}|S)}\bracket{\log \frac{P_{\phi}\paren{S| \bm{z}, \tAll}P_{\theta}(\tAll|\bm{z})\prod_{j=1}^mP_{\psi_j}(z_j)}{Q\paren{\bm\tau|S}\prod_{j=1}^mQ_{\varphi_j}\paren{z_j|\hat S_j}}}\\
& =  \E_{S,z_j, \bm\tau}\bracket{\log P_{\phi}\paren{S| \bm{z}, \tAll} + \sum_{j=1}^m\log \frac{P_{\psi_j}(z_j)}{Q_{\varphi_j}\paren{z_j|\hat S_j}} + \log \frac{P_\theta(\bm\tau|\bm{z})}{Q(\bm\tau|S)}}\\
\intertext{Since $Q(\bm{T}|S)$ is a deterministic distribution, $Q(\bm\tau|S) = 1$ for all $\bm\tau \sim Q(\bm\tau|S)$. Thus,}
& =  \E_{S,z_j, \bm\tau}\bracket{\log P_{\phi}\paren{S| \bm{z}, \tAll} + \sum_{j=1}^m\log \frac{P_{\psi_j}(z_j)}{Q_{\varphi_j}\paren{z_j|\hat S_j}} + \log P_\theta(\bm\tau|\bm{z})}\\
& = \E_{S,\bm{z}, \bm\tau}\bracket{\log P_{\phi}\paren{S| \bm{z}, \tAll}} - \sum_{j=1}^m\E_S \bracket{\KL{P_{\psi_j}(z_j)}{Q_{\varphi_j}\paren{z_j|\hat S_j}}}  + \E_{S,\bm{z}, \bm\tau}\bracket{\log P_\theta(\bm\tau|\bm{z})}.
    \end{split}
\end{equation}

As detailed in the main paper, each term in the final objective corresponds to the training loss for each of our components (part stylizers, transformation sampler, and cross diffusion network). The first term is the reconstruction error given prior information $\mathbf{z}, \bm{\tau}$, and its maximization is done by the cross diffusion network, which is a Denoising Diffusion Probabilistic Model with a generalized forward diffusion kernel. We elaborate on the derivation of the generalized forward kernel in Sec.~\ref{sec:gen_kernel}, and the cross diffusion network in Sec.~\ref{sec:cdn}. The second term is the KL divergence regularization loss for each of the part style latent distribution $P_{\psi_j}(Z_j)$. We implement each of the part style latent distributions with a prior flow model. For specifics, please refer to Sec.~\ref{sec:priorflow}. Lastly, the third/final term corresponds to the part transformations, where the cIMLE training strategy maximizes this the objective, as explained in  Sec.~\ref{sec:sampler}.


\section{Generalized Forward Diffusion Kernel}
\label{sec:gen_kernel}
We derive the forward diffusion process with the proposed generalized forward kernel in our Cross Diffusion Network (Sec. 4.3 of the main paper). Given a probability distribution $Q(X^{(0)})$ with $X^{(0)} \in \R^d$ that we want to model, we can define a forward diffusion process with a mean $\mu \in \R^d$ and a variance $\Sigma \in \R^{d\times d}_{> 0}$ such that 

$$
Q(X^{(0:T)}|\mu, \Sigma) = Q(X^{(0)})\prod_{t=1}^T Q(X^{(t)}|X^{(t-1)}, \mu, \Sigma)
$$ where 

$$
Q(X^{(t)}|X^{(t-1)}, \mu, \Sigma) = \mathcal{N}\paren{\sqrt{\alpha_t}X^{(t-1)} + \paren{1 - \sqrt{\alpha_t}}\mu, \paren{1 - \alpha_t}\Sigma}
$$

\noindent for each $t=1, \dots, T$. $\alpha_t = 1 - \beta_t$ are variance scheduling parameters. Notice that by setting $\Sigma = \bm{I}$ and $\mu = \bm{0}$ we recover the classical diffusion kernel. Similar to the classical kernel, we can efficiently sample from $Q\paren{X^{(t)}| x^{(0)}, \mu, \Sigma}$ using the reparameterization trick: 

$$
Q\paren{X^{(t)}| X^{(0)}, \mu, \Sigma}  = \mathcal{N}\paren{\sqrt{\overline{\alpha}_t}X^{(0)} + \paren{1 - \sqrt{\overline{\alpha}_t}}\mu,
    (1 - \overline{\alpha}_t)\Sigma}.
$$

\noindent Here, $\overline{\alpha_t} = \prod_{t^\prime=1}^t \alpha_{t^\prime}$. As $t\to \infty$ increases,  $\overline{\alpha_t}$ decreases to zero. Thus, for large $T$, the distribution $Q\paren{X^{(t)}| X^{(0)}, \mu, \Sigma}$ approaches the Gaussian distribution $\mathcal{N}\paren{\mu, \Sigma}$ with mean $\mu$ and variance $\Sigma$. By doing so, we can incorporate prior information in the forward diffusion process in terms of $\mu$ and $\Sigma$. We can also show that for each $t > 1$, the posterior distribution $Q(X^{(t-1)}|X^{(t)}, X^{(0)}, \mu, \Sigma)$ is also a Gaussian distribution, given by 

\begin{equation}\label{equ:gen-posterior}
    Q\paren{X^{(t-1)}|X^{(t)}, X^{(0)}, \mu, \Sigma} = \mathcal{N}\paren{\Xi_t, \eta_t^2\Sigma},
\end{equation}

\noindent where 

\begin{align*}
    \Xi_t &= \frac{\beta_t\sqrt{\overline{\alpha}_{t - 1}}}{1 - \overline{\alpha}_t}X^{(0)} + \frac{(1 - \overline{\alpha}_{t-1})\sqrt{\alpha_{t}}}{1 - \overline{\alpha}_t}X^{(t)} + \paren{1 + \frac{\paren{\sqrt{\overline{\alpha}_t} - 1}\paren{\sqrt{\alpha_t}+ \sqrt{\overline{\alpha}_{t - 1}}}}{1 - \overline{\alpha}_t}}\mu, \\
    \eta_t^2 &= \frac{\beta_t\paren{1 - \overline{\alpha}_{t - 1}}}{1 - \overline{\alpha}_t}. 
\end{align*}

\noindent We can further simplify the expression for $\Xi_t$ by substituting 

\begin{equation}\label{equ:efficient-sampling}
    X^{(t)} = \sqrt{\overline{\alpha}_t}X^{(0)} + \paren{1 - \sqrt{\overline{\alpha}_t}}\mu + (1 - \overline{\alpha}_t)\bm{\varepsilon}\sqrt{\Sigma},
\end{equation} 

\noindent where $\bm{\varepsilon}\sim \mathcal{N}(\bm{0}, \bm{I})$ and $\sqrt{\Sigma}$ is the Cholesky decomposition of $\Sigma$. Thus, similar to the classical diffusion parameterizations, we can define a reverse diffusion process $P_{\phi}(X^{(0:T)}| \mu, \Sigma)$ that is also a Markov chain so that 

$$
P_{\phi}(X^{(0:T)}| \mu, \Sigma) = P(X^{(T)}|\mu, \Sigma)\prod_{t=1}^T P_\phi(X^{(t-1)}|x^{(t)}, \mu, \Sigma),
$$ 
where each $P_\phi(X^{(t-1)}|x^{(t)}, \mu, \Sigma)$ is a Gaussian distribution parameterized by weights $\phi$ and tries to approximate the ground truth posterior distribution $Q\paren{X^{(t-1)}|X^{(t)}, X^{(0)}, \mu, \Sigma}$. Specifically, as in classical diffusion models~\cite{ho2020denoising}, we can parameterize each reverse kernel as a Gaussian distribution with learnable mean and fixed variance:
$$
P_\phi(X^{(t-1)}|x^{(t)}, \mu, \Sigma) = \mathcal{N}\paren{\Xi_\phi\paren{t, x^{(t)}, \mu, \Sigma}, \eta_t\sqrt{\Sigma}}
$$ for each $t > 1$. In practice, instead of directly learning $\Xi_\phi$, we can parameterize $\Xi_\phi$ using \cref{equ:gen-posterior} and learn a noise approximator $\bm{\varepsilon}_\phi$ that approximate random noises. At sampling time, we can recover $P_\phi(X^{(t-1)}|x^{(t)}, \mu, \Sigma)$ using $\bm{\varepsilon}_\phi$ by writing
\begin{equation}\label{equ:reparam}
\begin{split}
        \Xi_\phi\paren{t, x^{(t)}, \mu, \Sigma} &= \frac{\beta_t\sqrt{\overline{\alpha}_{t - 1}}}{1 - \overline{\alpha}_t}X^{(0)} + \frac{(1 - \overline{\alpha}_{t-1})\sqrt{\alpha_{t}}}{1 - \overline{\alpha}_t}X^{(t)} + \paren{1 + \frac{\paren{\sqrt{\overline{\alpha}_t} - 1}\paren{\sqrt{\alpha_t}+ \sqrt{\overline{\alpha}_{t - 1}}}}{1 - \overline{\alpha}_t}}\mu\\
        &= \frac{1}{\sqrt{\alpha_t}}\bm{x}^{(0)} - \frac{1 - \sqrt{\alpha_t}}{\sqrt{\alpha_t}}\mu - \frac{\beta_t\bm{\varepsilon}_{\phi}(t, x^{(t)}, \mu, \Sigma)}{\sqrt{\alpha_t \paren{1 - \overline{\alpha}_t}}}\sqrt{\Sigma}.
\end{split}
\end{equation}
Doing so, the training objective of the generalized diffusion process simplifies to an L2 losss with noise, as we discussed in our main paper. 

\section{Additional Experimental Results}
\label{sec:additional_experiments}
We provide additional experiment results in this section. We show additional quantitative comparisons in Sec.~\ref{sec:gen-score-add}. In Sec.~\ref{sec:autoencoding}, we evaluate \methodname's autoencoding quality compared with state-of-the-art point cloud reconstruction networks, as well as demonstrate the advantage of the generalized forward kernel in our Cross Diffusion Network in modeling complex part geometry and topology. In Sec.~\ref{sec:ablate-stylizer}, we ablate the prior flows in our part style sampler. In Sec.~\ref{sec:mixing} and Sec.~\ref{sec:interpolation}, we showcase shape mixing and part-level interpolation results compared with control-enabled baselines. Lastly, in Sec.~\ref{sec:part-sampling} and Sec.~\ref{sec:tau-sampling}, we showcase additional qualitative results on part-level and transformation sampling.

\subsection{Additional Quantitative Comparison}
\label{sec:gen-score-add}

\begin{table}[t]
\centering
\begin{tabular}{l|ccc|ccc}
    \toprule
   \multirow{2}{*}{} & \multicolumn{3}{c}{Chair} & \multicolumn{3}{c}{Airplane} \\
   \cline{2-4}\cline{5-7} 
   &  MMD-P ($\downarrow$) &  COV-P ($\uparrow$) &  1NNA-P  &  MMD-P ($\downarrow$) &  COV-P ($\uparrow$) &  1NNA-P \\ 
   \midrule
   SP-GAN~\cite{li2021sp} & 4.43 & 31.7 & 87.77 & 5.27 & 26.2 & 91.03\\
     SPAG~\cite{spaghetti}  & 4.53 & 36.3 & 78.94& 5.73 & 26.3 &90.17\\
    NW~\cite{wavelet}  & 4.77 & 34.0 & 83.36&  4.77 & 34.0 & 83.36 \\
    \midrule
   \textbf{\methodname} (Ours)  & \textbf{3.27} & \textbf{42.5} & \textbf{65.23} & \textbf{3.20} & \textbf{46.2} & \textbf{68.72}\\ \bottomrule
\end{tabular}
\caption{\textbf{Intra-part evaluation with additional baselines.} MMD-P score is multiplied by $10^{-2}$. COV-P and 1NNA-P are reported in $\%$}.
\label{tab:global-code-comp-full}
\end{table}


We provide additional quantitative comparisons for \methodname on part-level generation as shown in \cref{tab:global-code-comp-full}. Specifically, we include two state-of-the-art mesh-based generative networks  SPAGHETTI~\cite{spaghetti} and Neural Wavelets~\cite{wavelet}. As shown, although these methods require 3D mesh supervision, our approach is able to achieve better part-level generative scores compared to these works. Additionally, we also compare with SP-GAN~\cite{li2021sp}, which is a GAN-based point cloud generative network. We also note that unlike the other point cloud baselines, SP-GAN does not allow for encoding of an existing shape, making shape editing non-trivial if not infeasible.  We used pre-trained weights for the three networks and follow the same evaluation procedure as for the other baselines.


\subsection{Autoencoding}\label{sec:autoencoding}
\begin{table*}[h]
\begin{center}
\begin{tabular}{ll|cccccc}
\hline
Dataset                   & Metrics & DPM  & $\text{DPM}^\dagger$& ShapeGF & $\text{ShapeGF}^{\dagger}$& $\text{Ours}^*$ & Ours (\methodname) \\ \hline
\multirow{2}{*}{Airplane} & CD ($\downarrow$)     & 2.266 & 1.70& 2.082               & 1.599   & 1.460 &  \textbf{1.413}                                            \\
                          & EMD ($\downarrow$)    & 1.96 & 1.72  & 2.258 & 1.722           & 1.337                                   & \textbf{1.333}                                  \\ \hline
\multirow{2}{*}{Car}      & CD ($\downarrow$)     & 7.686 & 7.013 & 7.016 & 6.735        & 6.559                                   & \textbf{6.538}                                  \\
                          & EMD ($\downarrow$)    & 3.229 & 3.179  & 3.283 & 3.323           &\textbf{ 2.782}                                   & 2.788                                  \\ \hline
\multirow{2}{*}{Chair}    & CD  ($\downarrow$)    & 5.297 & 4.39  & 4.521 & 3.928   & 3.746                                   & \textbf{3.701}                                  \\
                          & EMD ($\downarrow$)   & 2.83 & 2.64   & 3.181 & 2.646              & 2.222                                   & \textbf{2.214}                              \\ \hline
\multirow{2}{*}{Lamp}     & CD ($\downarrow$)     & 8.334 & 5.55 & 8.243 &5.100       & 3.787                                   & \textbf{3.481}                                  \\
                          & EMD ($\downarrow$)    & 3.395 & 2.899     & 4.504 & 2.983     & 1.959                                   & \textbf{1.951}                                  \\ \hline
\end{tabular}
\caption{Comparison of point cloud auto-encoding performance. $\dagger$ represents a modification of baselines to incorporate part latents. $*$ represents our framework without modification of the forward kernel. Reported CD and EMD are multiplied by $10^4$ and $10^2$ respectively.}
\label{tab:scoe-recon-full}
\end{center}
\end{table*}

We report autoencoding performance of \methodname in \cref{tab:scoe-recon-full} compared to DPM~\cite{dpm} and ShapeGF~\cite{ShapeGF}. We further modify the two baselines to model part latents by also encoding each segmented part. We also report their reconstruction score denoted by $\dagger$ superscript. We see that \methodname outperforms all baselines by a margin due to our novel cross diffusion network and generalized forward kernel. We also note that there is an increase in performance for both baselines with the additional part latent modification $\dagger$. We also report the reconstruction score for our model without the generalized forward kernel, marked as \textit{Ours}$^*$. Notice that there is a larger gap between our approach with and without the generalized forward kernel in the lamp category in which parts exhibit more complex geometrical and topological structures.



\subsection{Ablation on Part Stylizer}\label{sec:ablate-stylizer}
\begin{table}[h]
\begin{center}
\begin{tabular}{c|ccc}
\toprule
 & MMD-P & COV-P & 1NNA \\
\midrule
VAE & 3.47 & \textbf{44.3} & 72.59 \\
CNF (Ours) & \textbf{3.27} & 42.5 & \textbf{65.23}\\
\bottomrule
\end{tabular}
\end{center}
\vspace{-0.4cm}
\caption{\textbf{Ablation on Part Stylizer.} Individual-part level metrics ablating our part stylizer (MMD-P $\times 10^{-2}$).}
\label{tab:ablation_part_stylizer}
\end{table}
We replace the CNF in our part stylizer by directly using a standard Gaussian as the prior, \ie VAE. \cref{tab:ablation_part_stylizer} shows intra-part evaluation metrics comparing CNF with standard VAE. With the lower 1NNA and MMD-P, we note that CNF gives an overall better performance in capturing the part-style distribution than with vanilla VAE. We also observe that shapes generated with CNF are overall much clearer and with less noise than with VAE. This is because the prior flow relaxes the latent space regularization constraint by allowing the prior distribution to be transformed into a more complex distribution. For details of the prior flow, see Sec.~\ref{sec:priorflow}.

\subsection{Additional Part Style Mixing Results}
\label{sec:mixing}
\begin{figure}[t]
\centering
\includegraphics[width=\textwidth]{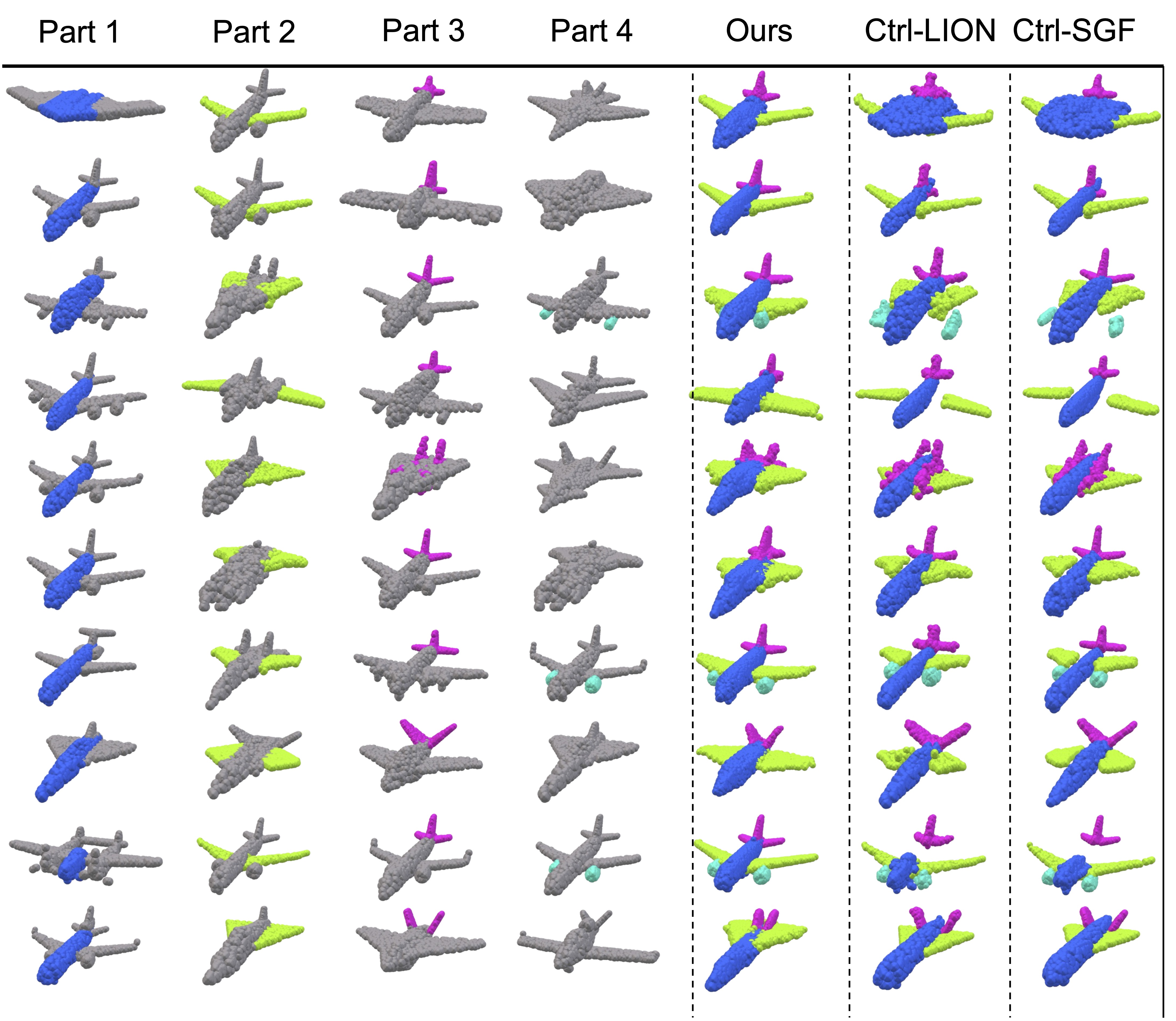}
\caption{\textbf{Shape Mixing: Airplanes.} For each row, parts from different shapes are encoded and mixed for both our method (\textbf{Ours}) and the control-enabled baselines (\textbf{Ctrl-LION}, \textbf{Ctrl-SGF}). Notice that the control-enabled baselines fail to produce coherent shapes because it does not model the distribution of valid transformations while our method transforms each part globally into a coherent configuration so that the generated shapes are plausible.}
   \vspace{-0.4cm}
\label{fig:mix-airplane}
\end{figure}

\begin{figure}[t]
\centering
\includegraphics[width=\textwidth]{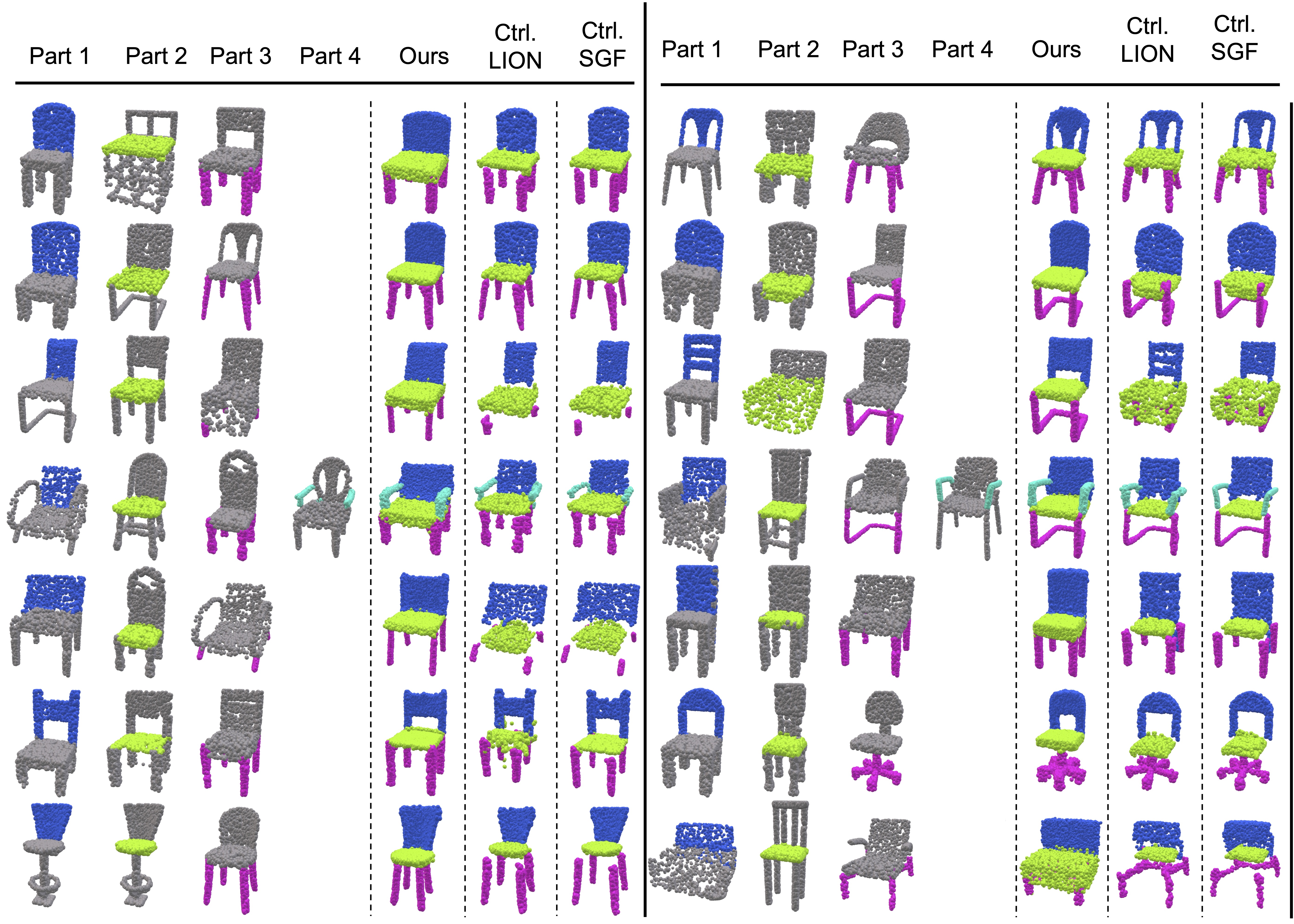}
\caption{\textbf{Shape Mixing: Chairs.} For each example (two examples per row, separated by the solid line), parts from different shapes are encoded and mixed for both our method (\textbf{Ours}) and the control-enabled baselines (\textbf{Ctrl-LION}, \textbf{Ctrl-SGF}). Notice that the control-enabled baselines fail to produce coherent shapes because it does not model the distribution of valid transformations while our method transforms each part globally into a coherent configuration so that the generated shapes are plausible.}
   \vspace{-0.4cm}
\label{fig:mix-chair}
\end{figure}

\begin{figure}[t]
\centering
\includegraphics[width=\textwidth]{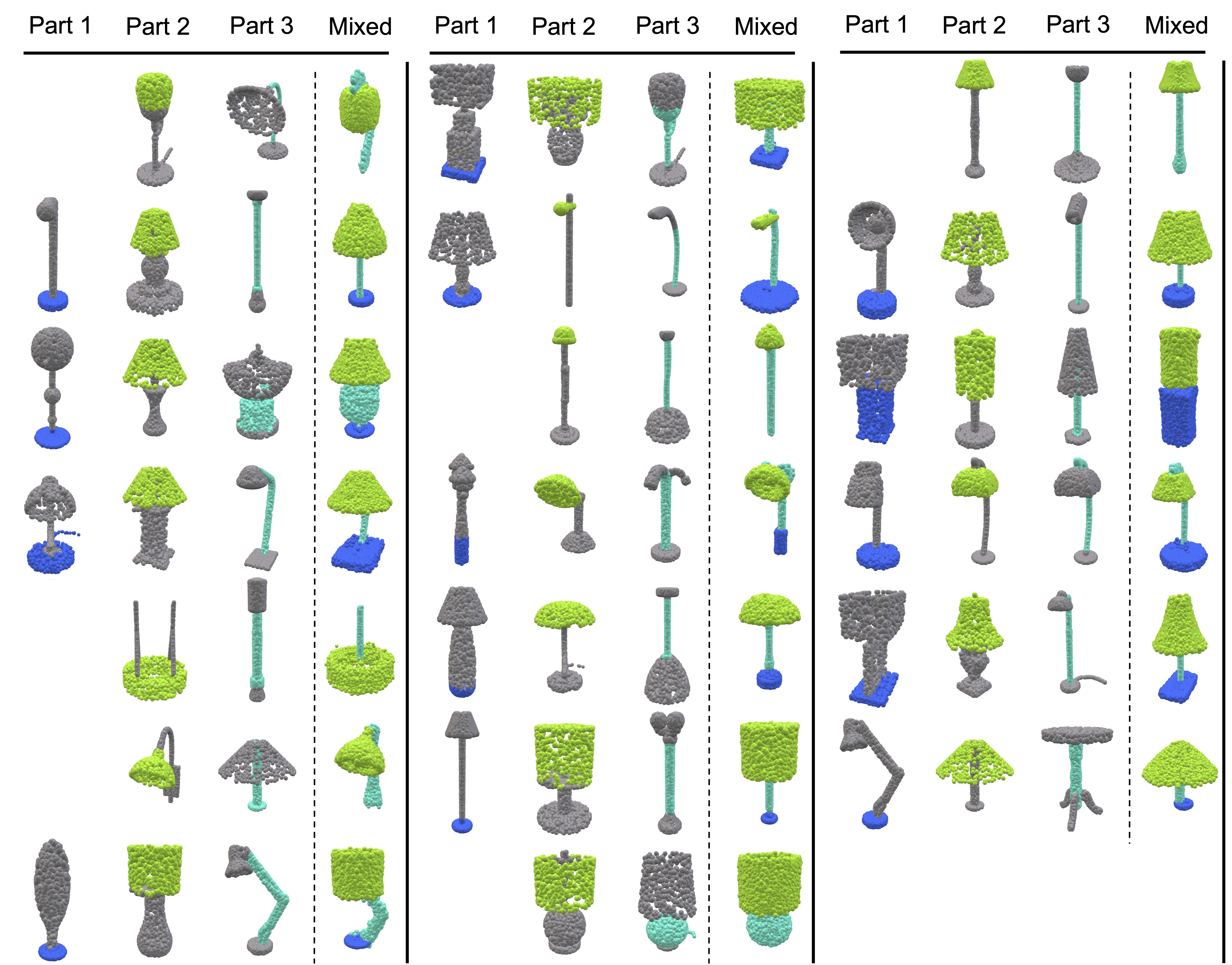}
\caption{\textbf{Shape Mixing: Lamps.} Each row shows three examples of shape mixing for lamps. Notice that our model can adjust the parts' size and location so that the mixed shape is still coherent.}
   \vspace{-0.4cm}
\label{fig:mix-lamp}
\end{figure}

Since we sample from independent part-style distributions to generate coherent shapes, shape mixing is a direct application where modeling the factorized latent distribution has a direct advantage. We showcase additional part style mixing results in chairs, airplanes, and lamps categories in \cref{fig:mix-chair}, \cref{fig:mix-airplane}, and \cref{fig:mix-lamp}. We also further evaluate our mixing quality compared to our control-enabled baselines, Ctrl-LION~\cite{lion} and Ctrl-ShapeGF~\cite{ShapeGF}, for the chair and airplane categories. Because we model a distribution of parts' size and location given a set of part style latent codes, across all the examples, we are able to produce coherent shapes despite the input parts having different sizes and locations in their original source shapes. In particular, when compared to the control-enabled baselines that do not explicitly model parts sizes, the baselines' mixing results often produce shapes with detached or interpenetrating parts as a result of incorrect part configuration. 

\subsection{Additional Part Level Interpolation Results}\label{sec:interpolation}
\begin{figure}[t]
\centering
\begin{center}
   \includegraphics[width=\textwidth]{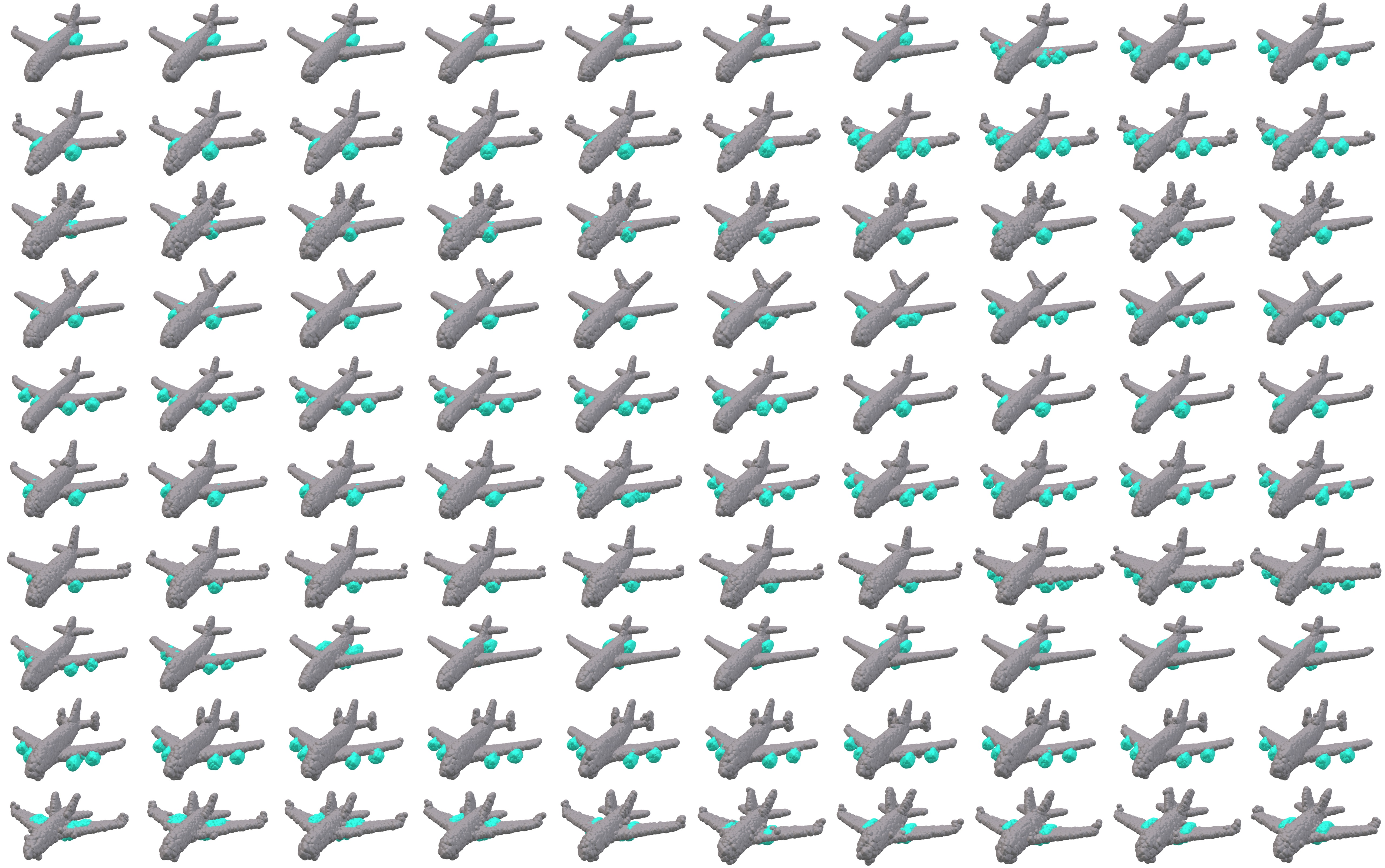}
\end{center}
   \caption{\textbf{Airplane Part Level Interpolation: Engines.} Each row is one interpolation path of the engine (colored part). Notice that because of the discrete nature of engine's number and location, the interpolation path between a different number of engines, and different locations of engines shows distinctive jumps from one configuration to another. This implies that the part latent space learned by our model only generates plausible shapes.}
   \vspace{-0.4cm}
\label{fig:engine-interp}
\end{figure}

\begin{figure}[t]
\centering
\begin{center}
   \includegraphics[width=\textwidth]{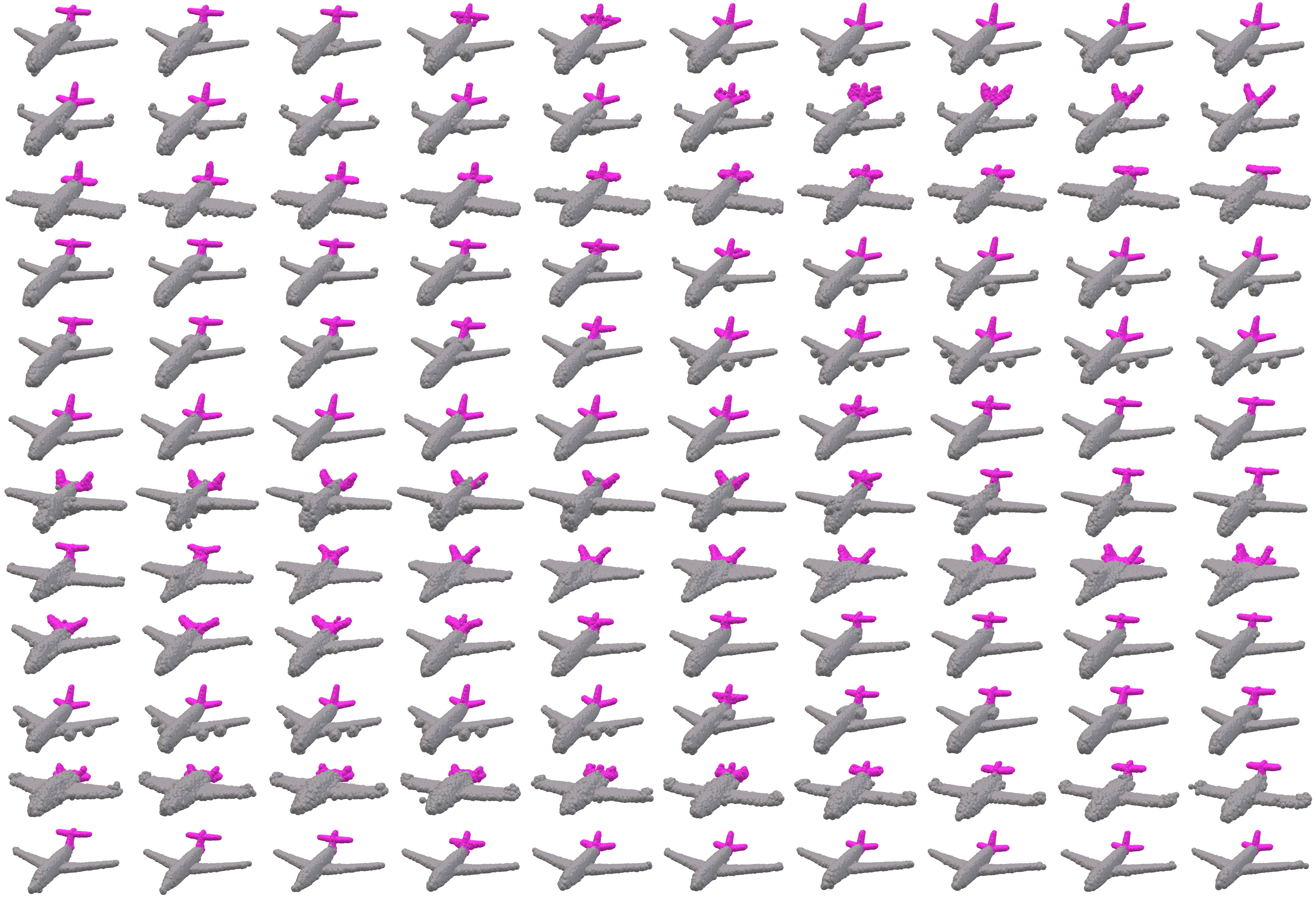}
\end{center}
   \caption{\textbf{Airplane Part Level Interpolation: Tail Wings.} Each row is one interpolation path of the tail wing (colored part). Similar to the engine's interpolation, the tail exhibits distinctive jumps from one mode to another so that the intermediate shapes are all plausible and coherent.}
   \vspace{-0.4cm}
\label{fig:tail-interp}
\end{figure}

\begin{figure}[t]
\centering
\begin{center}
   \includegraphics[width=\textwidth]{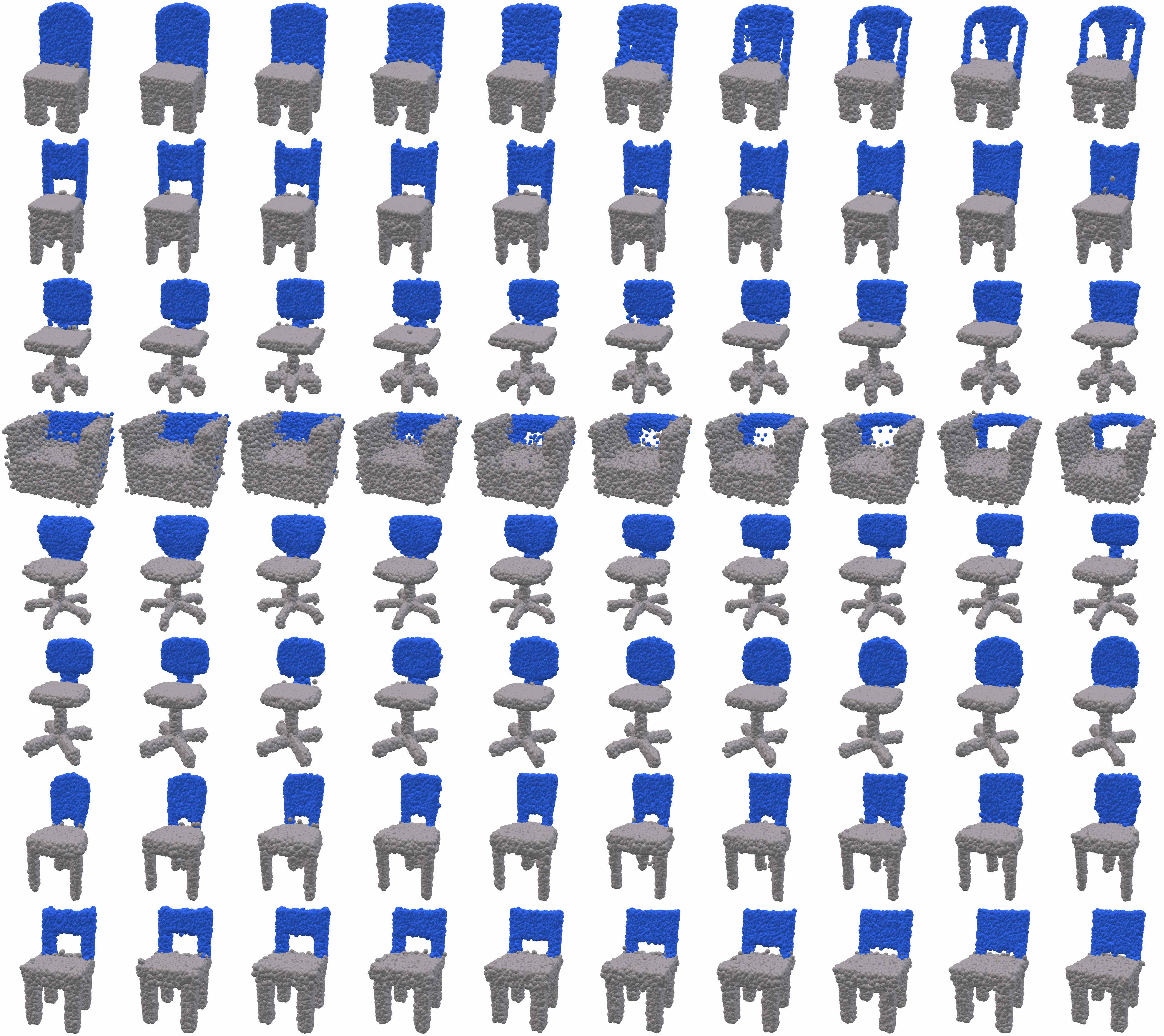}
\end{center}
   \caption{\textbf{Chair Part Level Interpolation: Back.} Each row is one interpolation path of the back (colored part). Notice that to adjust for the change in the back's style during interpolation, the unreferenced parts (in grey) adjust their size and location so that the output shapes are still coherent (see rows 3 and 6 for such covariance).}
   \vspace{-0.4cm}
\label{fig:back-interp}
\end{figure}

\begin{figure}[t]
\centering
\begin{center}
   \includegraphics[width=\textwidth]{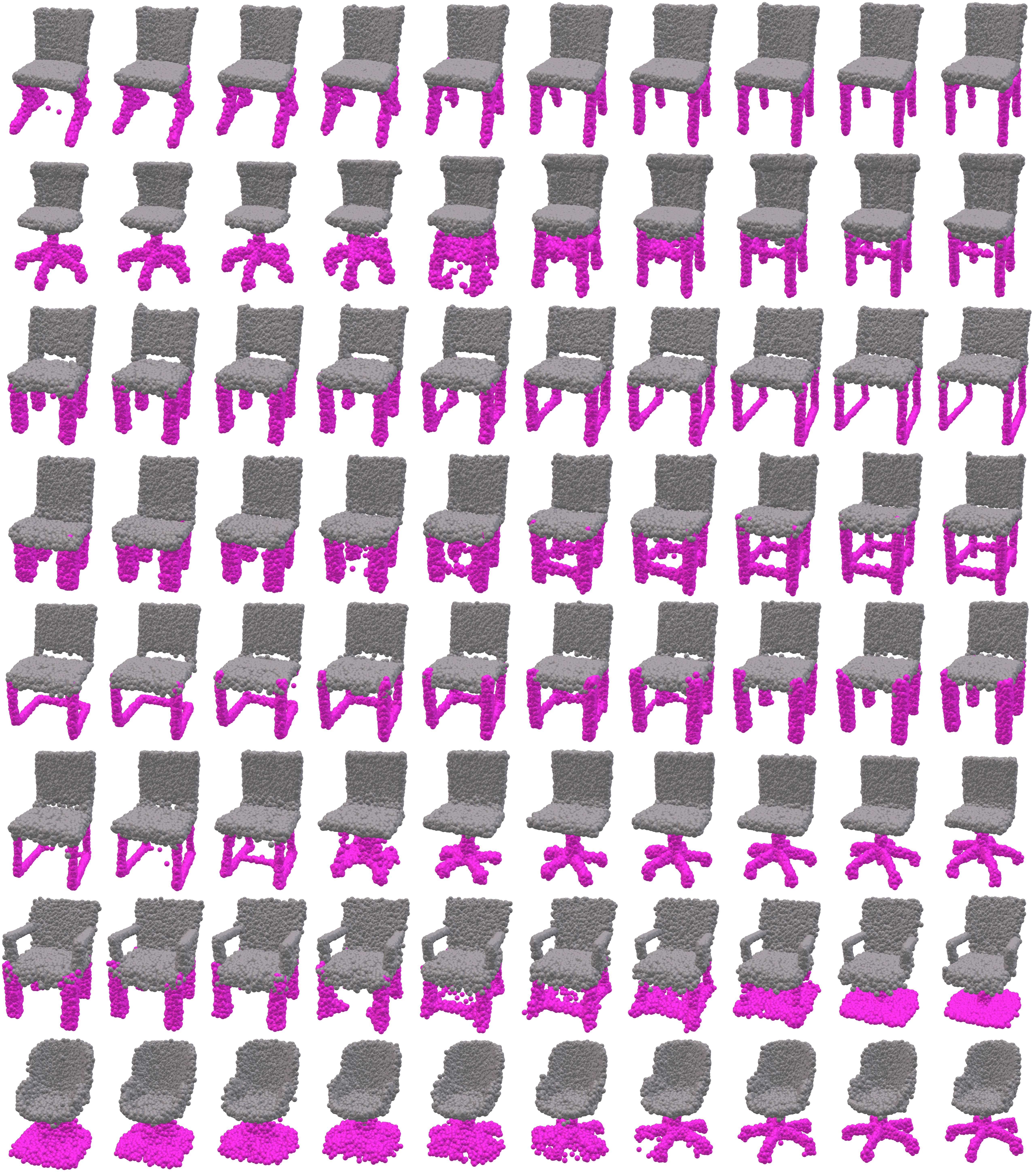}
\end{center}
   \caption{\textbf{Chair Part Level Interpolation: Legs.} Each row is one interpolation path of the legs (colored part). Notice that the intermediate shapes exhibit little artifacts and the interpolation path smoothly transitions between two different styles of legs. See, for example, rows 2 and 6.}
   \vspace{-0.4cm}
\label{fig:leg-interp}
\end{figure}

Moreover, since we explicitly model a part-style latent space, we can  also interpolate shapes on the part level. Given two samples of part style codes $z_j, z^\prime_j \sim P_{\psi_j}(Z_j),$ we use linear interpolation from $z_j$ to $z_j^\prime$ obtain intermediate part latents uniformly spaced in between: $z_j^{k} = z_j + \frac{k}{10}(z^\prime_j - z_j)$. We showcase qualitative results in all four categories in \cref{fig:engine-interp}, \ref{fig:tail-interp}, \ref{fig:back-interp}, \ref{fig:leg-interp} and \ref{fig:body-interp}. Notice that in all four categories, our interpolation results show smooth transitions from the source to the target of the interpolation, as most highlighted in the chair leg interpolation in \cref{fig:leg-interp}. 
Moreover, the intermediate shapes are also plausible, where discrete jumps can be observed when no intermediate configuration is available. An example of this can be seen in the interpolation of engines in \cref{fig:engine-interp}, where the number of engines jumps from four to two to allow a discrete transition. 
Moreover, we see that the unreferenced parts (colored in grey) remain unchanged during the interpolation process because we model independent distributions of part styles. Overall, the part-level interpolation results showcase our models' ability to generate plausible shapes with fine-grained control.

We also qualitatively compare \methodname with the control-enabled baselines, Ctrl-LION~\cite{lion} and Ctrl-ShapeGF~\cite{ShapeGF}, on part-level interpolation quality. \cref{fig:lion-interp-airplane,fig:lion-interp-chair} show part level interpolation result for Ctrl-LION and \cref{fig:sgf-interp-airplane,fig:sgf-interp-chair} show part level interpolation result for Ctrl-ShapeGF, for the chair and airplane categories. 
We follow the same part-level interpolation procedure on the control-enabled baselines as described earlier for our method.
Since the control enabled baselines do not explicitly model the distribution of part transformations given their styles, the interpolation of each part may result in detached parts. See the interpolation on chairs in \cref{fig:lion-interp-chair}. Moreover, since Ctrl-ShapeGF generates shapes by training a cGAN directly on its autoencoding latent space, its interpolation is less smooth as the autoencoding latent space is trained on discrete training examples. This can also be seen from the chair interpolation in \cref{fig:sgf-interp-chair} (see the last row where the intermediate geometry exhibits extensive artifacts). 

\subsection{Additional Part Level Sampling Results}
\label{sec:part-sampling}
We showcase that \methodname enables part-based controllable generation through part-level sampling. We include additional results on part-style sampling to supplement the left and middle column of Figure 3 in the main paper. We showcase both \emph{sampling} a part style and \emph{fixing} a part style for each shape. \cref{fig:partsample-airplane,fig:partsample-chair,fig:partsample-lamp} show additional qualitative results for \emph{sampling} individual part styles (Figure 3 main paper -left). \cref{fig:fix_part-airplane,fig:fixpart-chair,fig:fixpart-lamp} show additional qualitative results for \emph{fixing} individual part styles (Figure 3 main paper -middle).

\subsection{Additional Part-Configuration Sampling Results}\label{sec:tau-sampling}
We further show additional qualitative examples on our ability to enable part-configuration sampling, \ie generate various plausible configurations of the shape with fixed part styles. \cref{fig:sizesample-airplane,fig:sizesample-chair,fig:sizesample-lamp,fig:sizesample-car} show additional qualitative results. This is supplement the right column in Figure 3 of the main paper.

\begin{figure}[t]
\centering
\begin{center}
   \includegraphics[width=\textwidth]{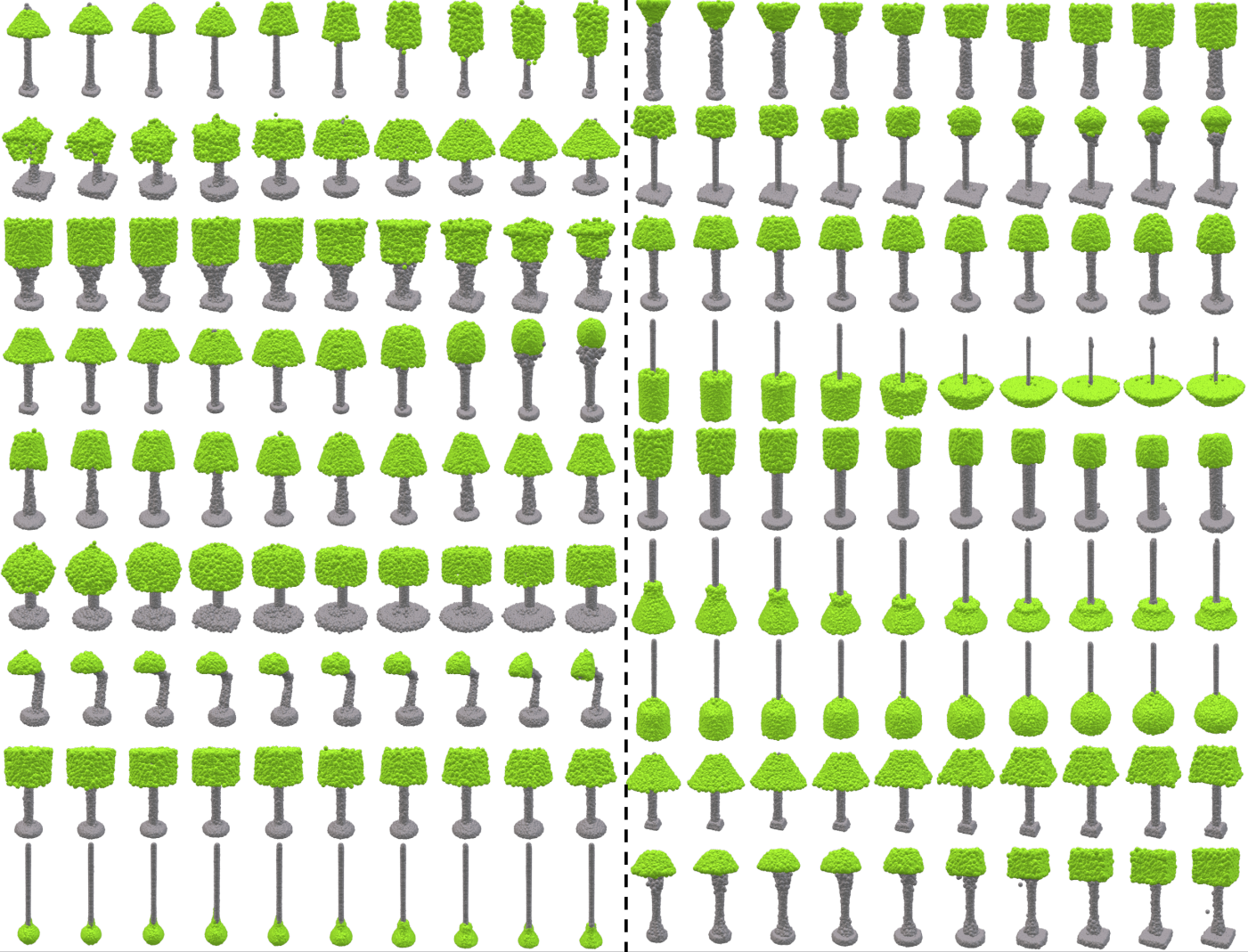}
\end{center}
   \caption{\textbf{Lamp Part Level Interpolation: Lamp Cap.} Each row shows two interpolation paths of the lamp cap (colored part) separated by the dashed line. Notice that for some of the interpolations where the starting part geometry is different from the ending geometry, the pole (in grey) needs to adjust its size so that the generated shape is coherent.}
   \vspace{-0.4cm}
\label{fig:cap-interp}
\end{figure}

\begin{figure}[t]
\centering
\begin{center}
   \includegraphics[width=\textwidth]{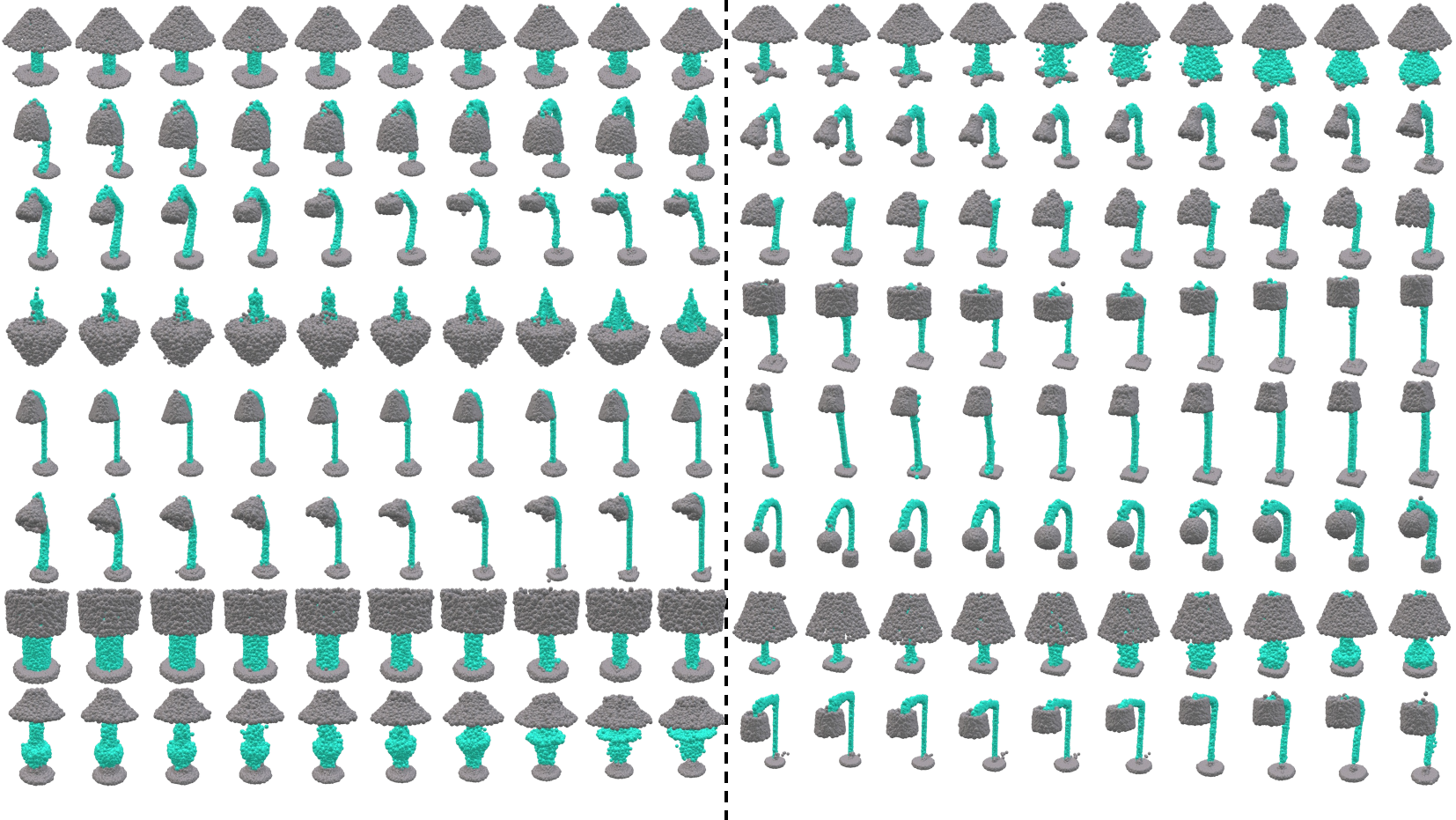}
\end{center}
   \caption{\textbf{Lamp Part Level Interpolation: Lamp Pole.} Each row shows two interpolation paths of the lamp pole (colored part) separated by the dashed line. Notice that sometimes the size of the lamp cap needs to vary accordingly to fit the change in the pole's style. See, for example, interpolations on rows 2 and 4.}
   \vspace{-0.4cm}
\label{fig:pole-interp}
\end{figure}

\begin{figure}[t]
\centering
\begin{center}
   \includegraphics[width=\textwidth]{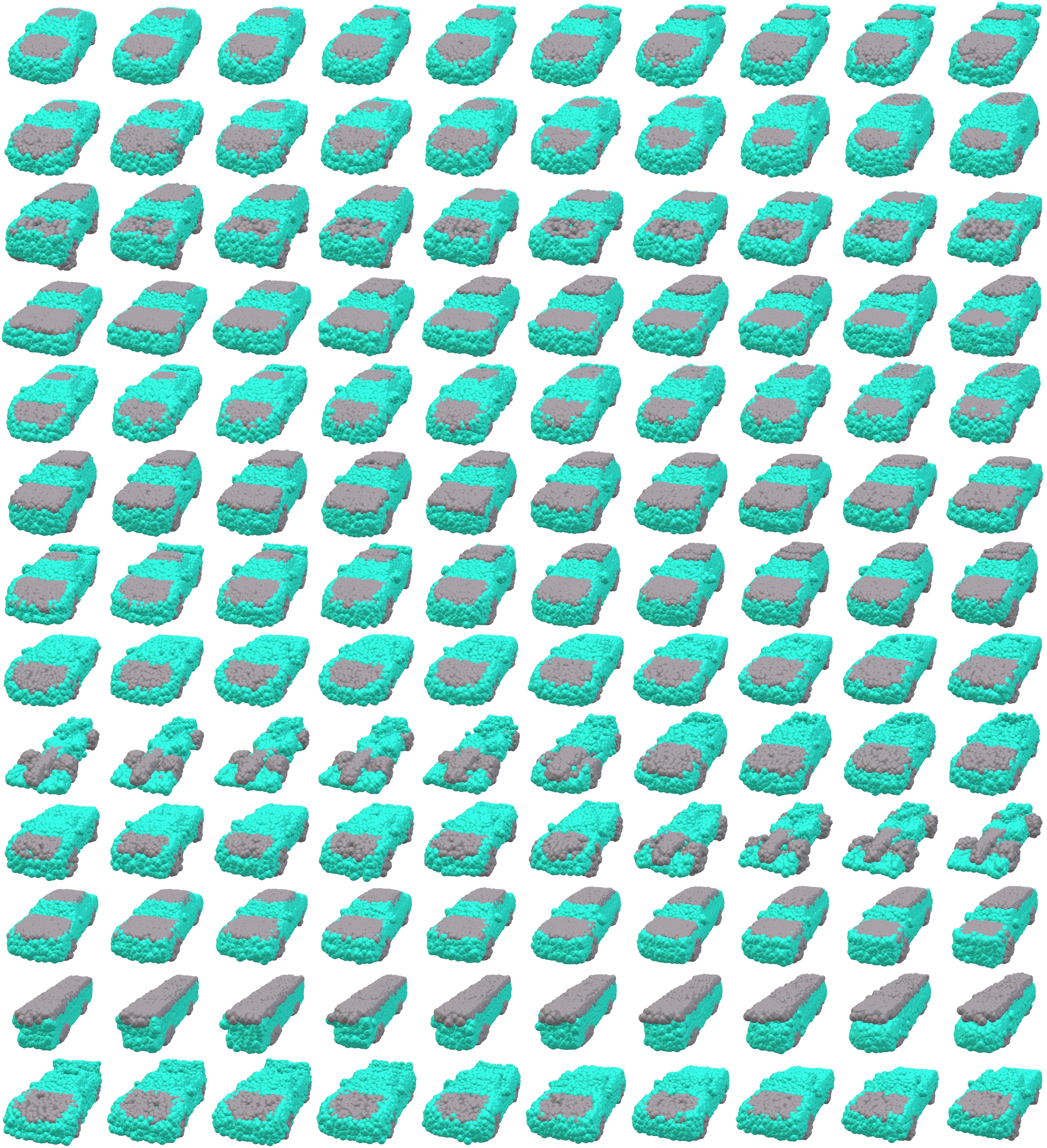}
\end{center}
   \caption{\textbf{Car Part Level Interpolation: Body} Each row shows one interpolation path of the car's body (colored part). Notice that the unreferenced parts vary accordingly to adjust to the change in the body's style (See rows 9 and 12).}
   \vspace{-0.4cm}
\label{fig:body-interp}
\end{figure}

\begin{figure}[t]
\centering
\includegraphics[width=\textwidth]{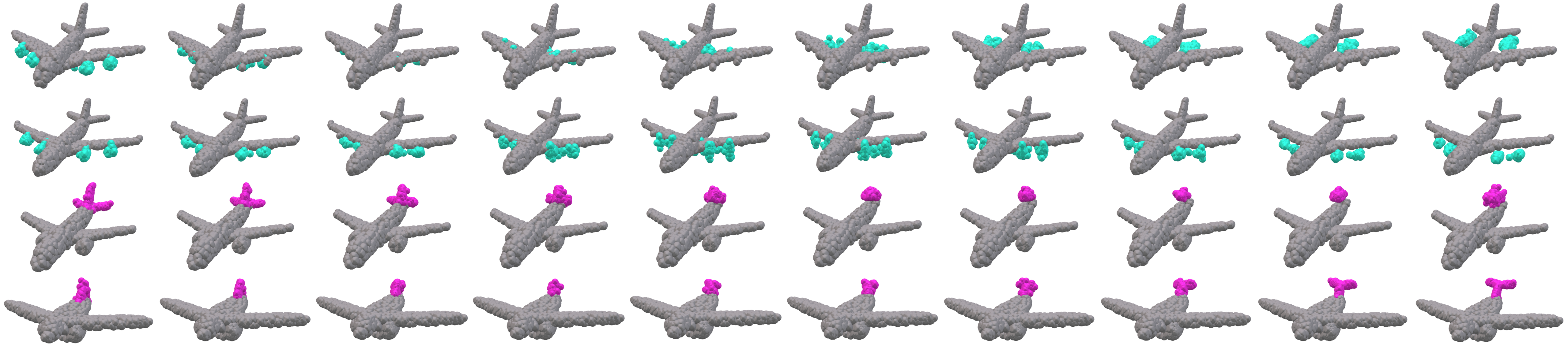}
\caption{\textbf{Control LION Part Level Interpolation: Airplanes.} Each row shows one interpolation path of the colored part. Notice that since the control enabled LION does not model the distribution of part configurations, the intermediate shapes during interpolation often become invalid and not coherent. See, for example, row 3.}
   \vspace{-0.4cm}
\label{fig:lion-interp-airplane}
\end{figure}

\begin{figure}[t]
\centering
\includegraphics[width=\textwidth]{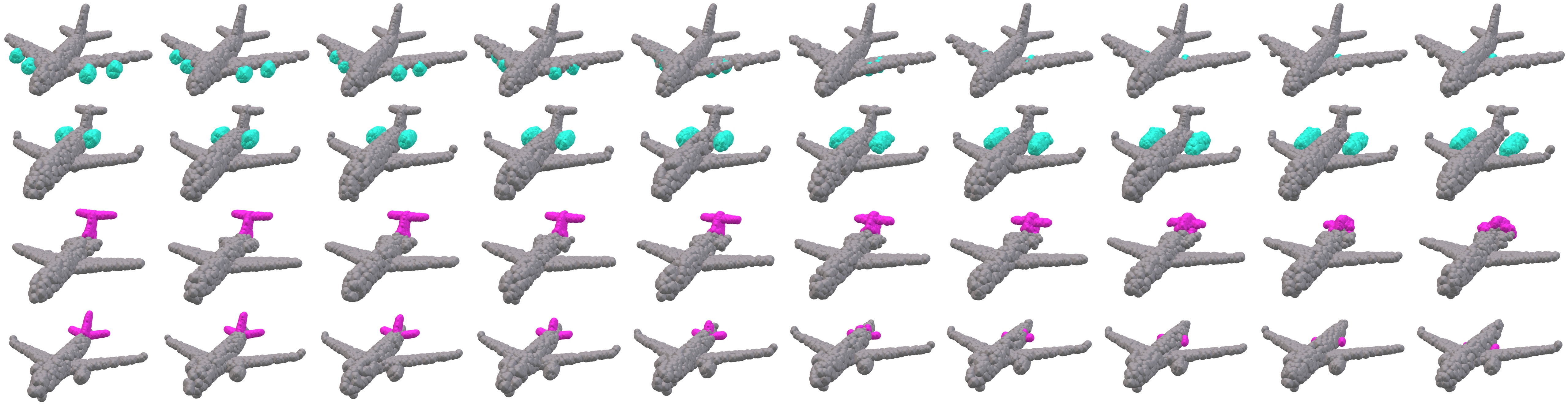}
\caption{\textbf{Control ShapeGF Part Level Interpolation: Airplanes.} Each row shows one interpolation path of the colored part. Since the control enabled ShapeGF samples part latents on its autoencoding latent space, the interpolation path often shows non-smoothness in its intermediate steps. See, for example, the last row.}
   \vspace{-0.4cm}
\label{fig:sgf-interp-airplane}
\end{figure}

\begin{figure}[t]
\centering
\begin{center}
   \includegraphics[width=\textwidth]{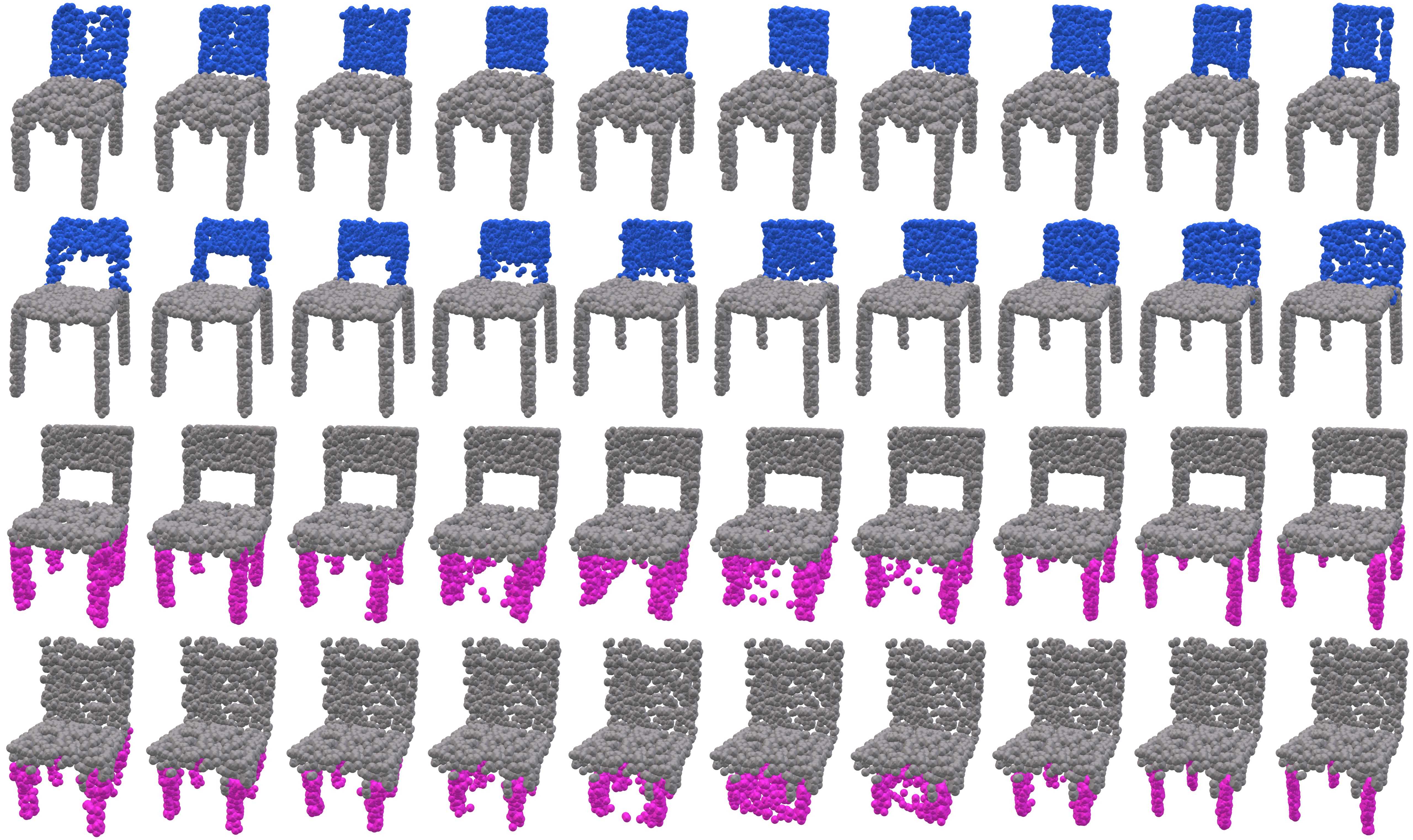}
\end{center}
   \caption{\textbf{Control LION Part Level Interpolation: Chairs.} Each row shows one interpolation path of the colored part. Since the control-enabled LION does not model the distribution of part configuration, the intermediate shapes often become detached (see the first row for such an example).}
   \vspace{-0.4cm}
\label{fig:lion-interp-chair}
\end{figure}

\begin{figure}[t]
\centering
\begin{center}
   \includegraphics[width=\textwidth]{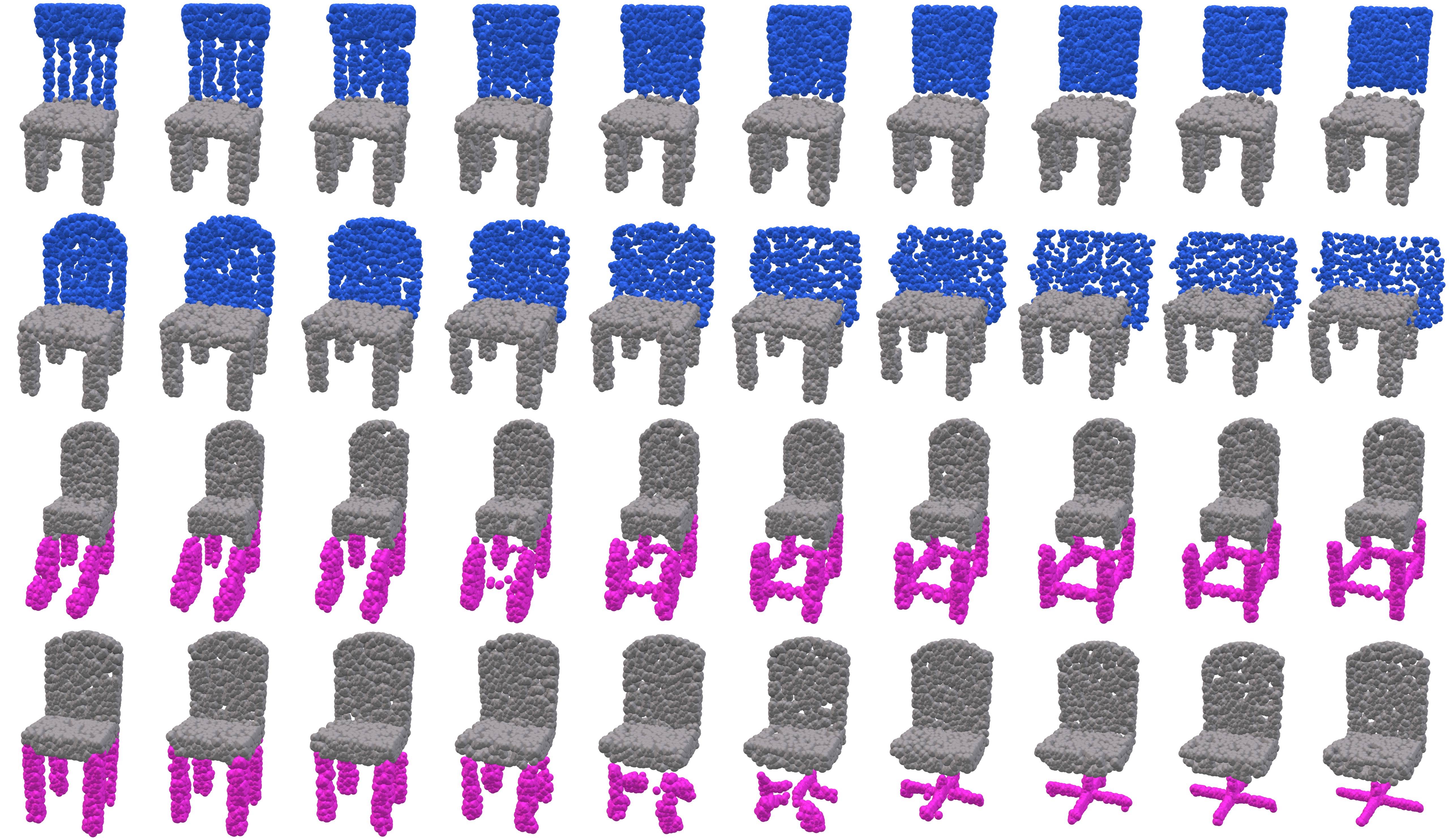}
\end{center}
   \caption{\textbf{Control ShapeGF Part Level Interpolation: Chairs.} Each row shows one interpolation path of the colored part. The last row shows the control-enabled ShapeGF exhibits artifacts in its interpolated shapes. Furthermore, parts become detached because it does not model the distribution of part transformation.}
   \vspace{-0.4cm}
\label{fig:sgf-interp-chair}
\end{figure}
\begin{figure}[t]
\centering
\begin{center}
   \includegraphics[width=\textwidth]{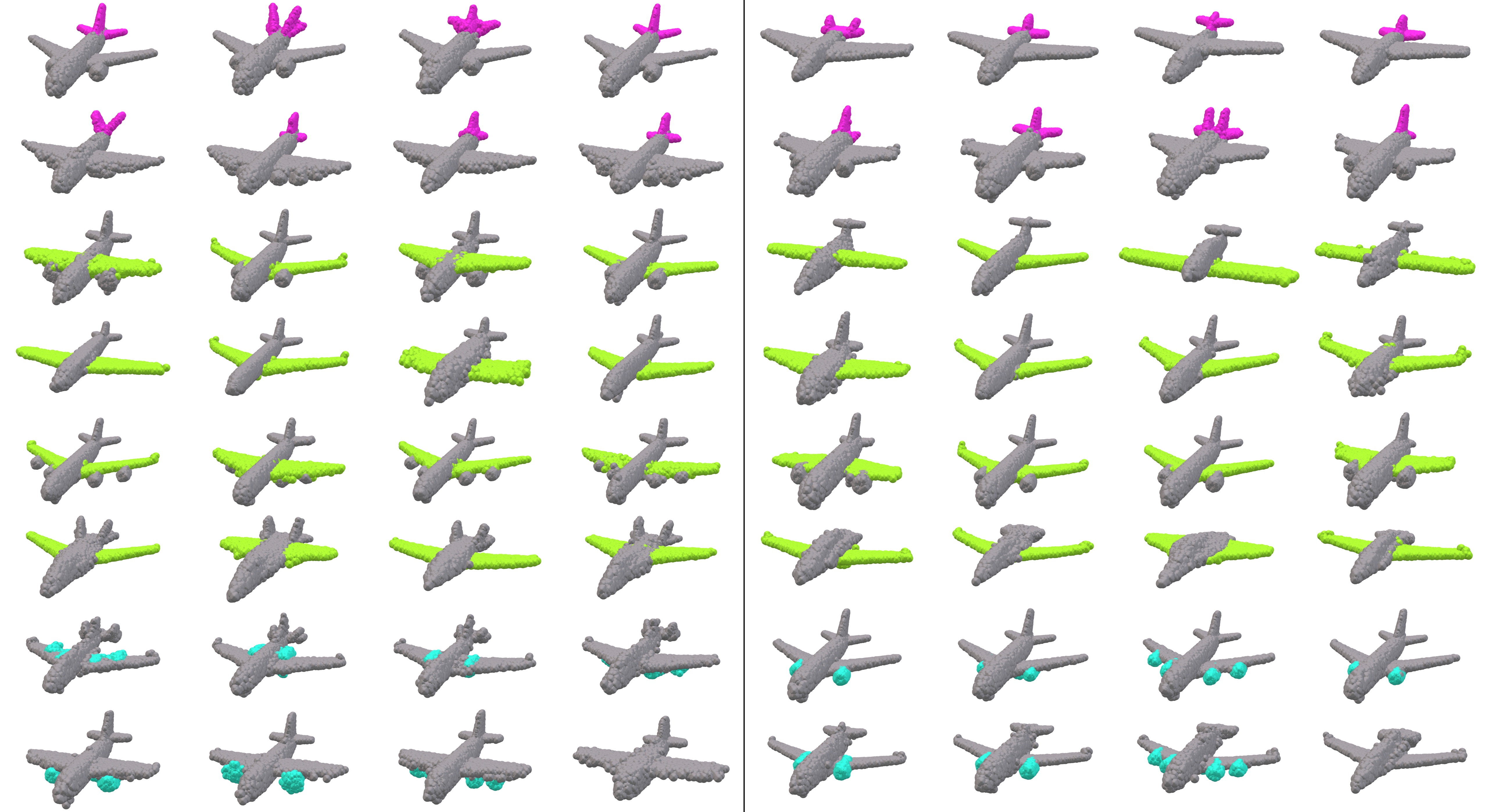}
\end{center}
   \caption{\textbf{Part-level Sampling: Airplanes, Sampling one part.} Each row shows two sets of airplanes (separated by the solid line) generated by sampling the part style latent associated with the colored part while keeping the rest of the part style latents for the grey parts fixed. Notice that the grey parts' styles stay fixed while the colored part varies across different samples. This ability to sample on the part level allows for intuitive user control in shapes generated.}
   \vspace{-0.4cm}
\label{fig:partsample-airplane}
\end{figure}

\begin{figure}[t]
\centering
\begin{center}
   \includegraphics[width=\textwidth]{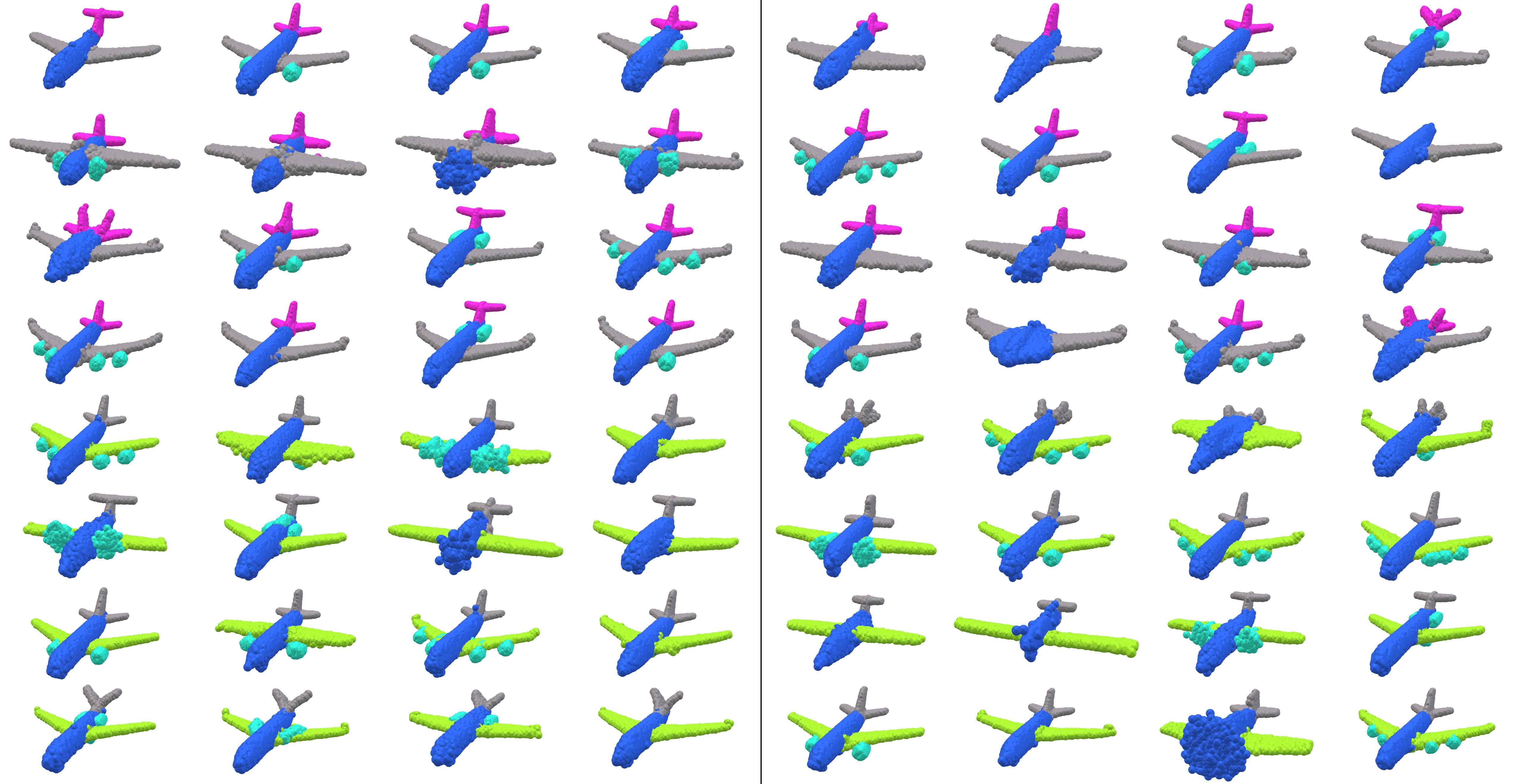}
\end{center}
   \caption{\textbf{Part-level Sampling: Airplanes, Fixing one part.} Each row shows two sets of airplanes (separated by the solid line) generated by sampling the set of part style latents associated with the colored parts while keeping the part style latent for the grey part fixed. Notice that the grey part's style stays fixed while the colored parts vary across different samples. The size and location of the grey part adjust to the sampled other part styles to generate coherent shapes.}
   \vspace{-0.4cm}
\label{fig:fix_part-airplane}
\end{figure}

\begin{figure}[t]
\centering
\begin{center}
   \includegraphics[width=0.85\textwidth]{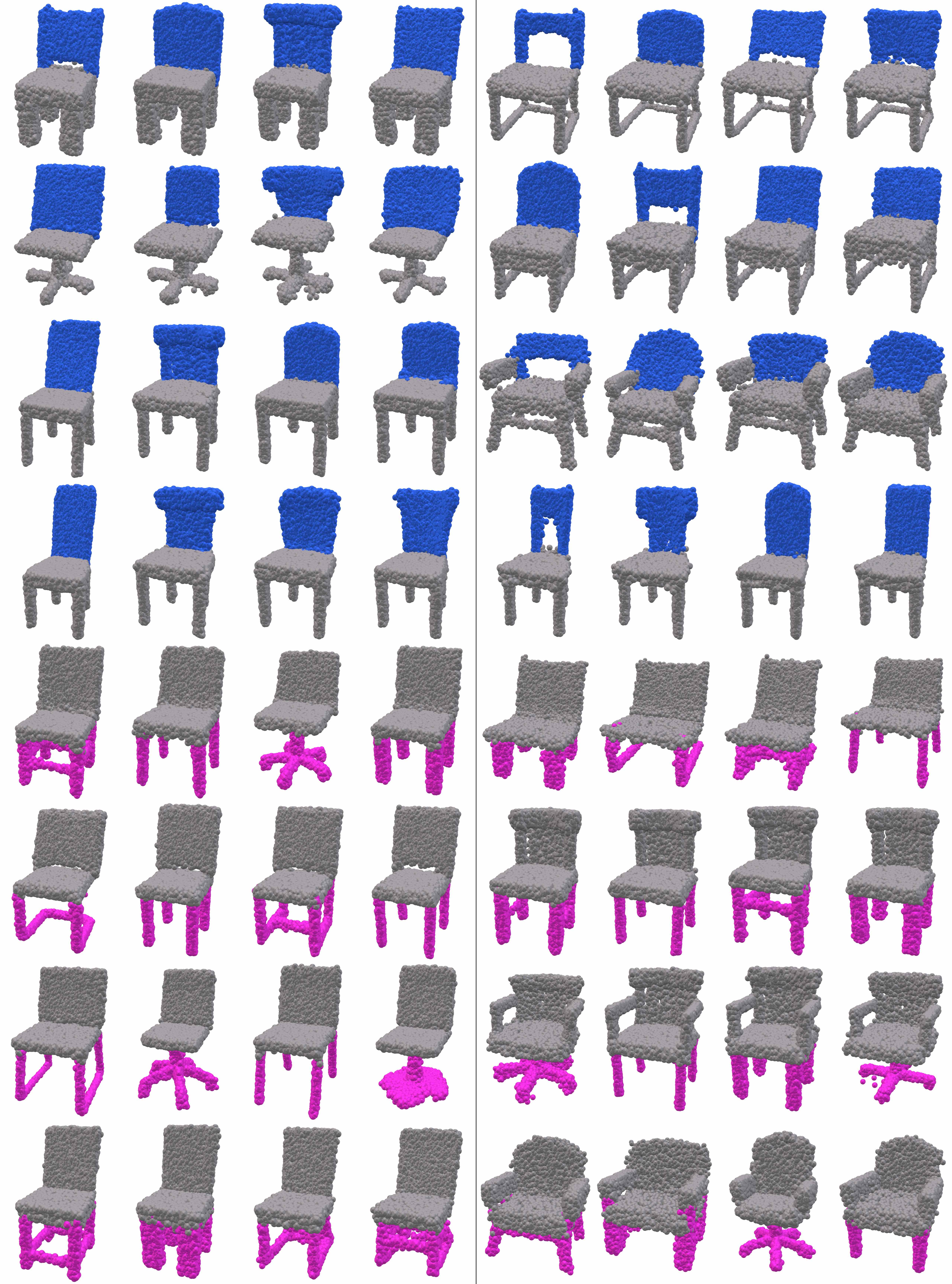}
\end{center}
   \caption{\textbf{Part-level Sampling: Chairs, Sampling one part.} Each row shows two examples (separated by the solid line) of sampling the colored part style latent while keeping the other part style latents unchanged. Notice that, based on the sampled part style, the unreferenced part changes their sizes accordingly so that the output shape stays plausible and coherent. See, for example, the change of seat's size on rows 2 and 3.}
   \vspace{-0.4cm}
\label{fig:partsample-chair}
\end{figure}

\begin{figure}[t]
\centering
\begin{center}
   \includegraphics[width=0.8\textwidth]{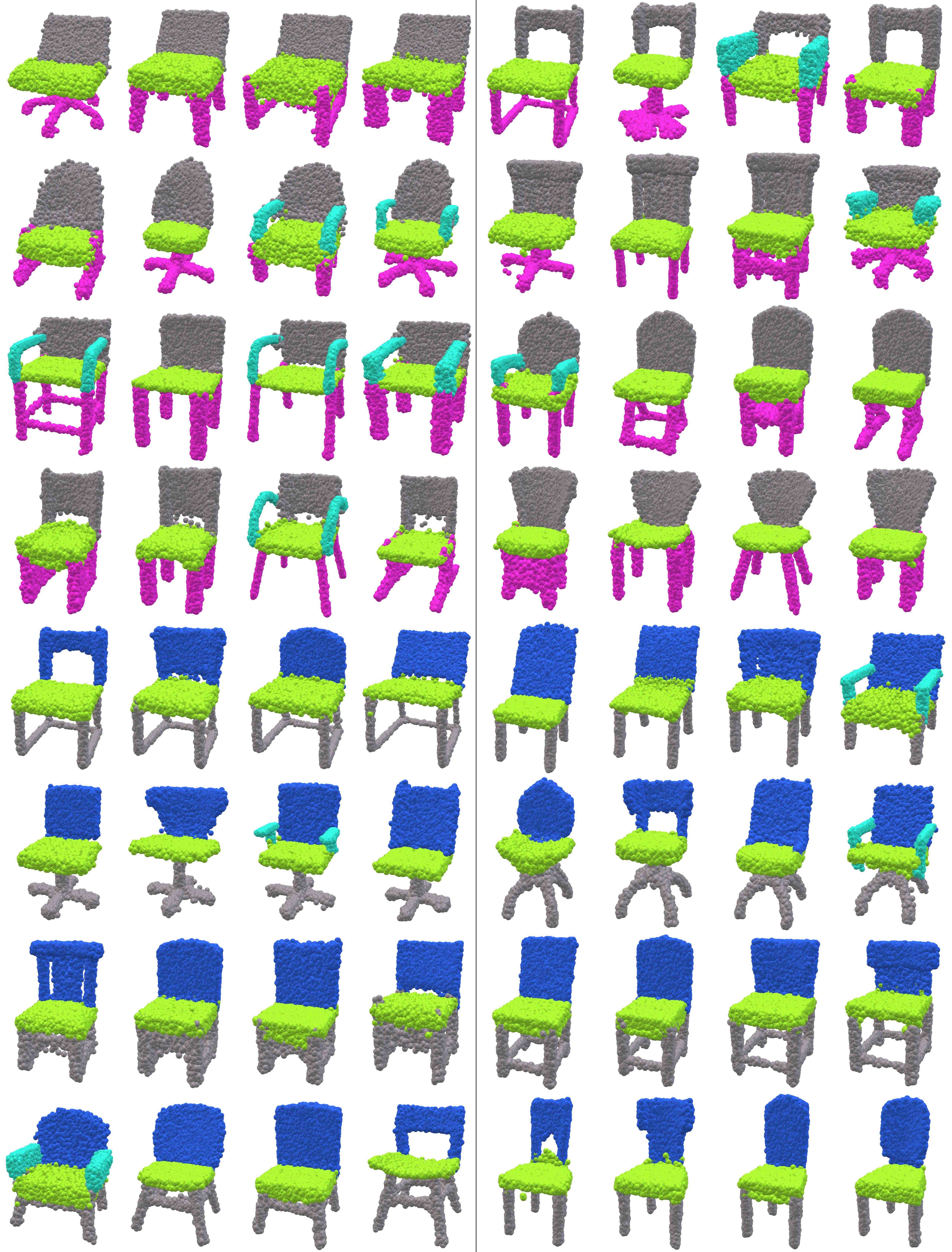}
\end{center}
   \caption{\textbf{Part-level Sampling: Chairs, Fixing one part.} Each row shows two sets of chairs (separated by the solid line) generated by sampling the set of part style latents associated with the colored parts while keeping the part style latent for the grey part fixed. Because we model the part configurations independently from their styles, the generated shapes are all plausible and coherent while showing desired controls.}
   \vspace{-0.4cm}
\label{fig:fixpart-chair}
\end{figure}

\begin{figure}[t]
\centering
\begin{center}
   \includegraphics[width=0.8\textwidth]{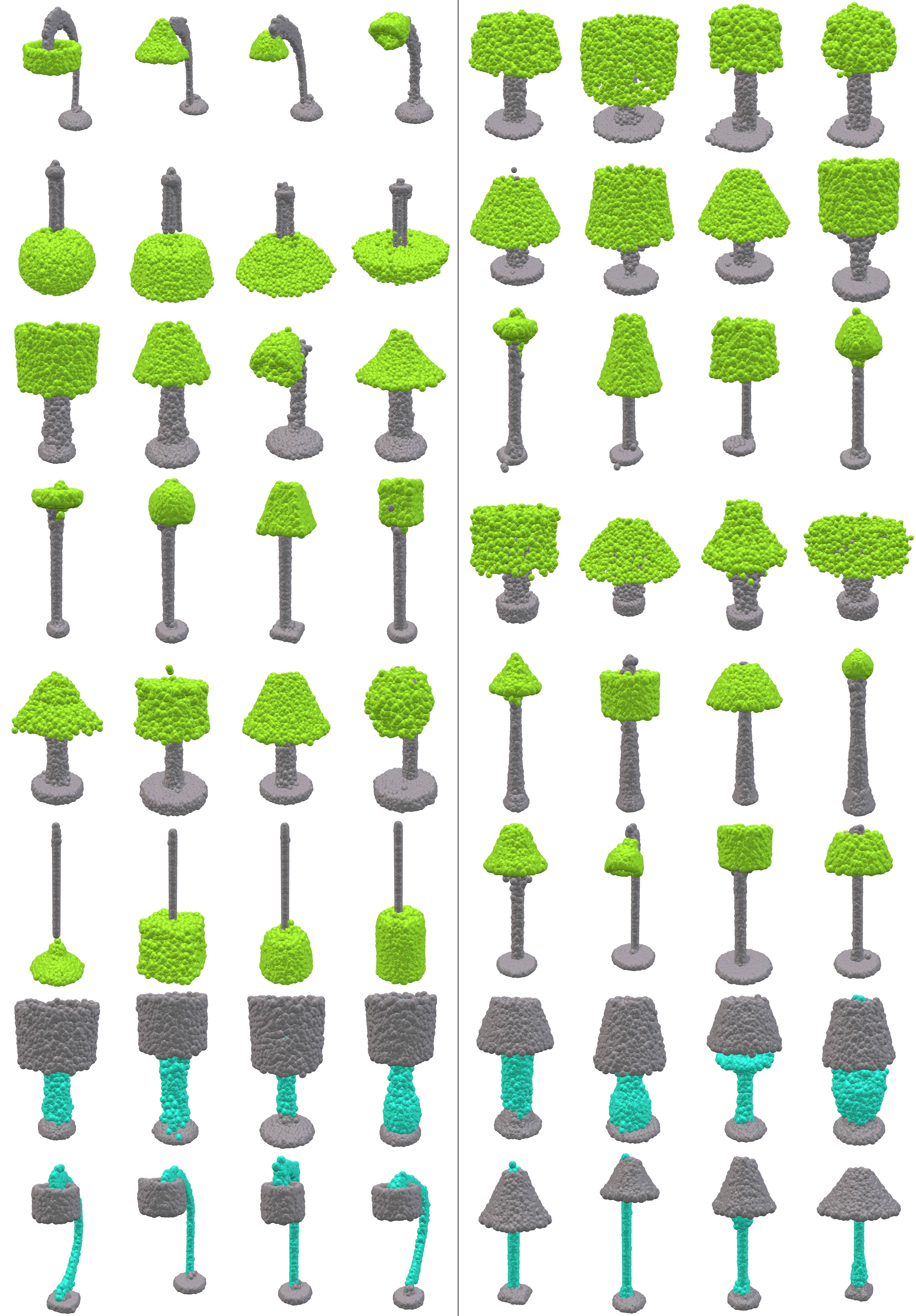}
\end{center}
   \caption{\textbf{Part-level Sampling: Lamps, Sampling one part.} Each row shows two sets of lamps (separated by the solid line) generated by sampling the colored part style latent while keeping the other part style latents (colored in grey) unchanged. Notice that the generated samples show diversity in style for colored parts, while the grey part's style stays fixed while changing its size and location to produce coherent shapes.}
   \vspace{-0.4cm}
\label{fig:partsample-lamp}
\end{figure}

\begin{figure}[t]
\centering
\begin{center}
   \includegraphics[width=0.68\textwidth]{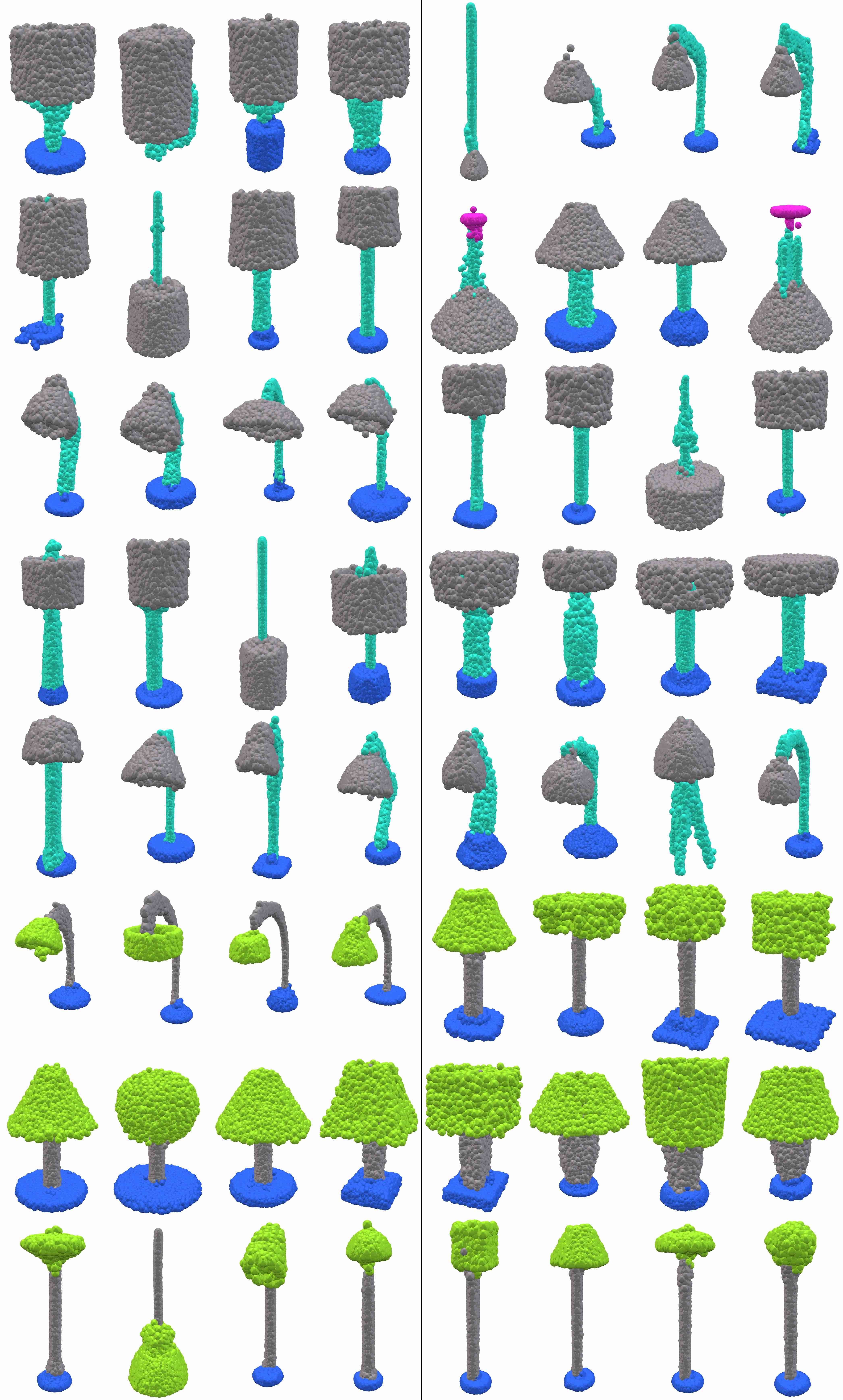}
\end{center}
   \caption{\textbf{Part-level Sampling: Lamps, Fixing one part.} Each row shows two sets of lamps (separated by the solid line) generated by sampling the set of part style latents associated with the colored parts while keeping the part style latent for the grey part fixed. Notice that the size and location of the grey part change accordingly to fit the different samples of part styles of the other parts}
   \vspace{-0.4cm}
\label{fig:fixpart-lamp}
\end{figure}

\begin{figure}[t]
\centering
\begin{center}
   \includegraphics[width=0.8\textwidth]{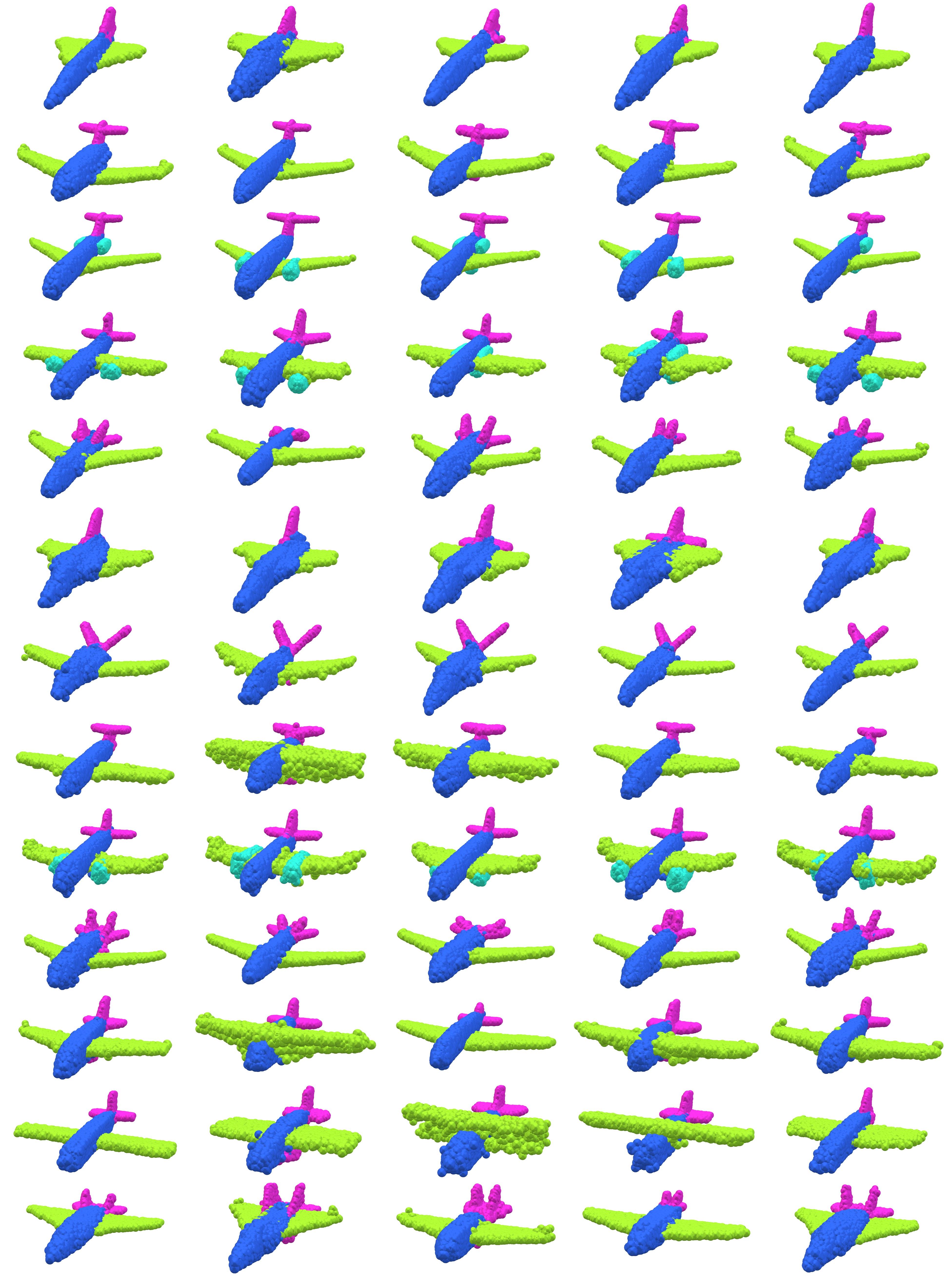}
\end{center}
   \caption{\textbf{Part Transformation Sampling: Airplanes.} Each row shows generated airplanes conditioned with different part transformation samples from the transformation sampler. All the part style latent for each row is kept fixed. Notice that the multimodality modeled by cIMLE training method allows for a diverse set of part transformation samples from the transformation sampler.}
   \vspace{-0.4cm}
\label{fig:sizesample-airplane}
\end{figure}

\begin{figure}[t]
\centering
\begin{center}
   \includegraphics[width=\textwidth]{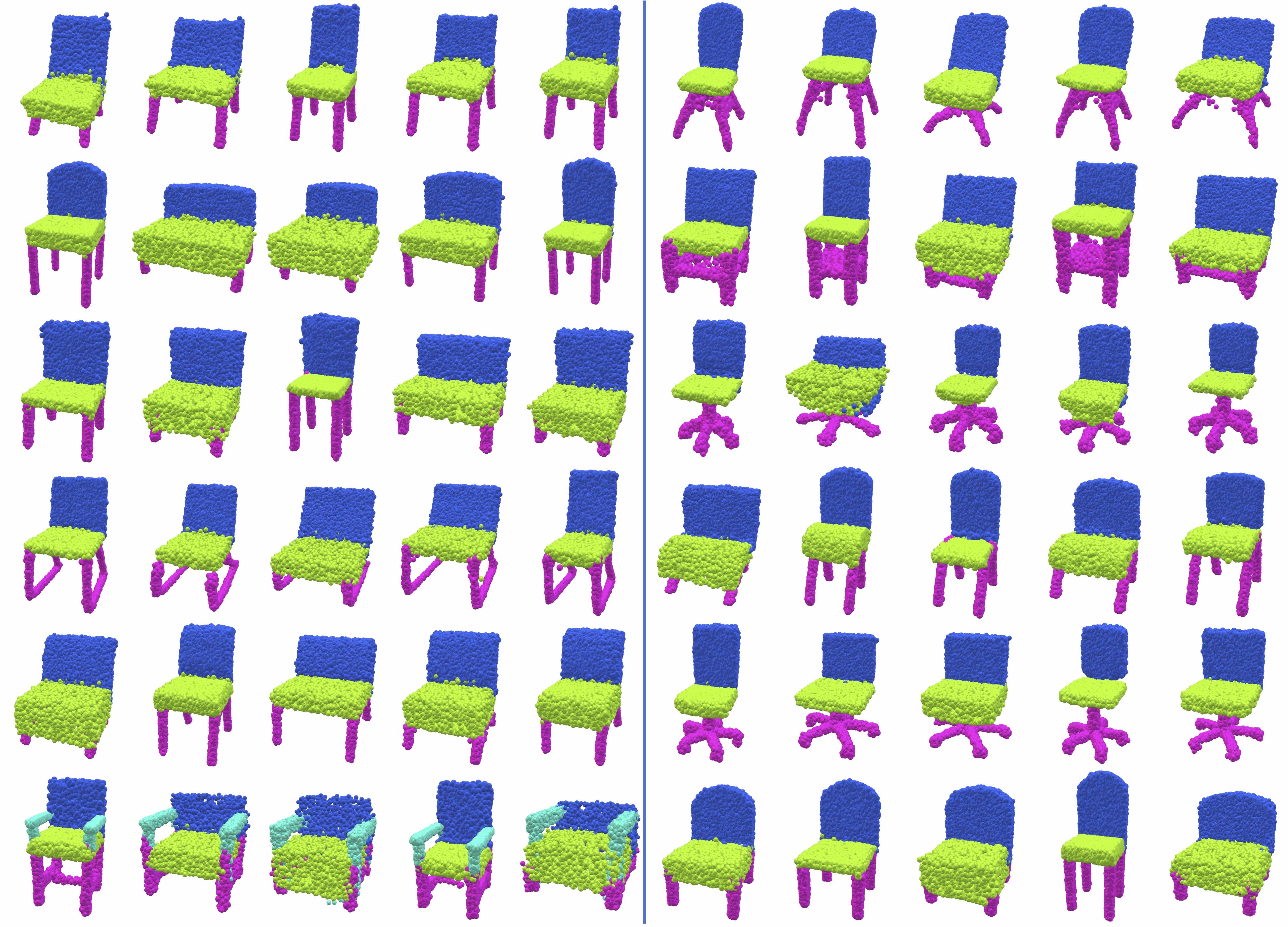}
\end{center}
   \caption{\textbf{Part Transformation Sampling: Chairs.} Each row shows generated airplanes conditioned with different part transformation samples from the transformation sampler. All the part style latent for each row is kept fixed.}
   \vspace{-0.4cm}
\label{fig:sizesample-chair}
\end{figure}

\begin{figure}[t]
\centering
\begin{center}
   \includegraphics[width=\textwidth]{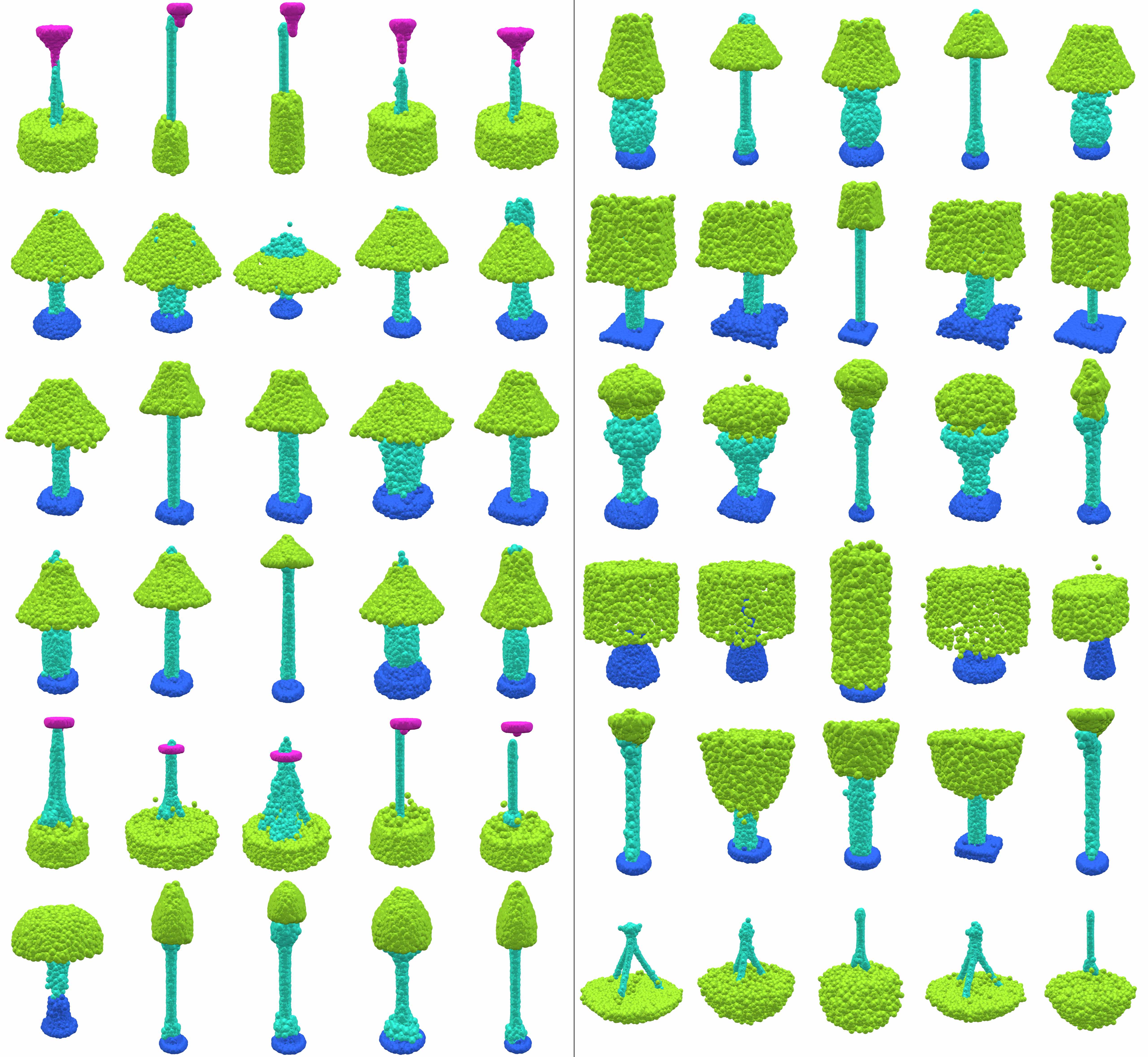}
\end{center}
   \caption{\textbf{Part Transformation Sampling: Lamps.} Each row shows generated airplanes conditioned with different part transformation samples from the transformation sampler. All the part style latent for each row is kept fixed.}
   \vspace{-0.4cm}
\label{fig:sizesample-lamp}
\end{figure}

\begin{figure}[t]
\centering
\begin{center}
   \includegraphics[width=0.8\textwidth]{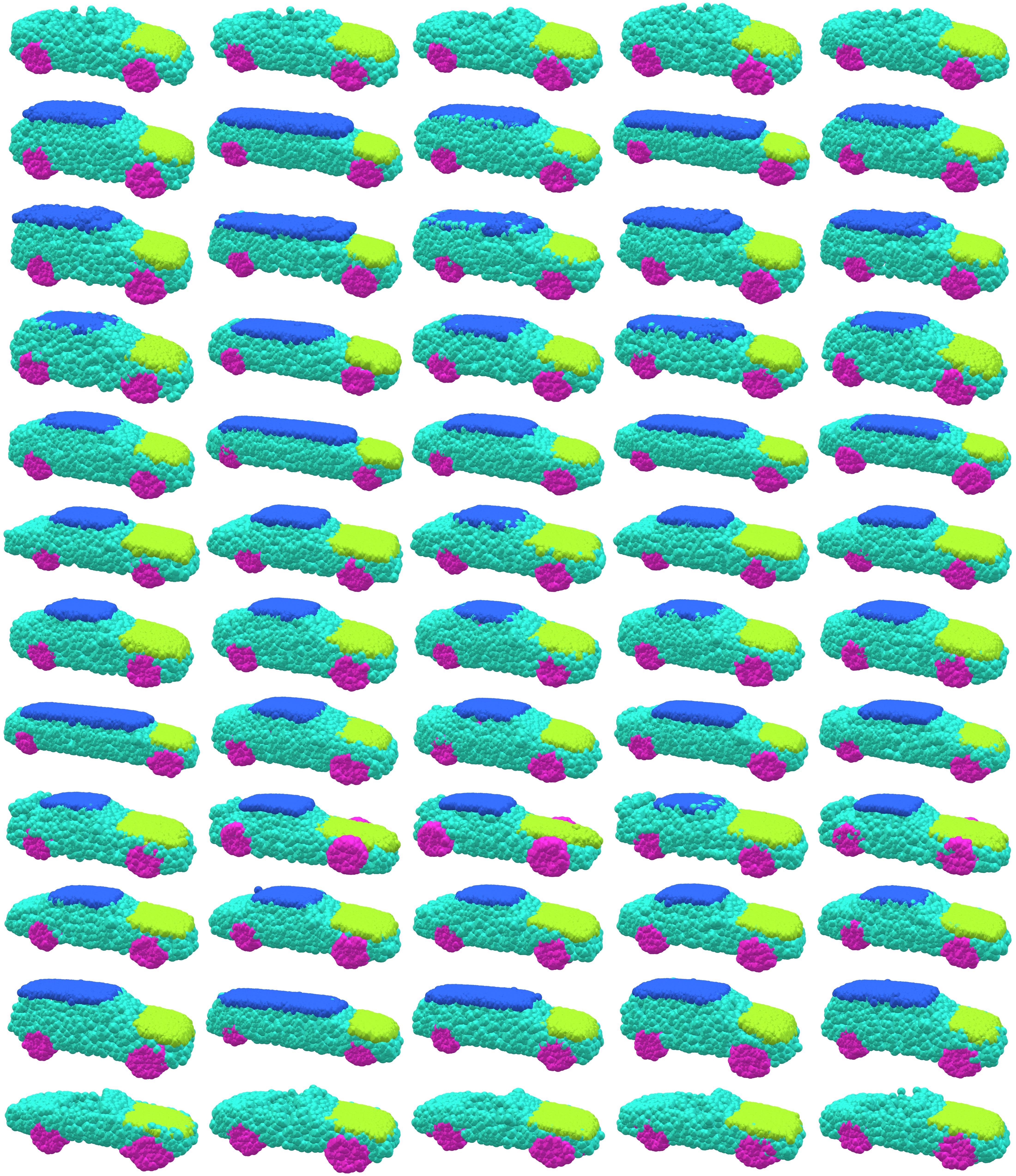}
\end{center}
   \caption{\textbf{Part Transformation Sampling: Cars.} Each row shows generated airplanes conditioned with different part transformation samples from the transformation sampler. All the part style latent for each row is kept fixed.}
   \vspace{-0.4cm}
\label{fig:sizesample-car}
\end{figure}


\clearpage
\newpage
\onecolumn

\section{Network Details}
\label{sec:network}
We elaborate on each of our components in this section. Sec.~\ref{sec:priorflow} presents our part stylizers that learn an independent part style latent space for each part with a continuous normalizing flow model (CNF). Sec.~\ref{sec:sampler} elaborates on our transformation sampler that is trained with conditional Maximum Likelihood Estimation (cIMLE) to model multimodality natural to the distribution of valid part configurations. Lastly, Sec.~\ref{sec:cdn} details the Cross Diffusion Network using Denoising Diffusion Probabilistic Model (DDPM) with the generalized forward kernel. 

\subsection{Part Style Sampler}
\label{sec:priorflow}

To model the part style distribution $P_{\psi_j}(Z_j)$ for each part, we learn a variational encoder $Q_{\varphi_j}(Z_j|\hat{S}_j)$ that models a Gaussian distribution with a learnable mean and diagonal variance given a canonicalized part $\hat{S}_j$. In practice, each part is canonicalized such that $\hat{S}_j$ shifted by the mean and scaled by one standard deviation on each of the axes.

To efficiently learn the variation encoder, we use the reparametrization trick~\cite{vae} giving us $z_j = \mu_{\varphi_j} + \sigma_{\varphi_j}\varepsilon$ where $\varepsilon \sim \mathcal{N}(\bm{0}, \bm{I})$.



Although it is possible to use a standard Gaussian for the prior distribution $P_{\psi_j}(Z_j),$ it has been shown~\cite{pointflow, dpm, lossvae} that a simple prior distribution used in VAE models can be less than ideal. To enrich the expressivity of the prior distribution, continuous normalizing flow (CNF)~\cite{pointflow} can be used to parameterize the prior distribution as a learnable, invertible network. To this end, we parameterize each per part style distribution $P_{\psi_j}(Z_j)$ as an invertible network and write the KL divergence as \cite{pointflow}
\begin{equation}
    \begin{split}
        \KL{P_{\psi_j}(z_j)}{Q_{\varphi_j}\paren{z_j|\hat S_j}} & = -\E_{z_j \sim Q_{\varphi_j}(Z_j|\hat{S}_j)}\bracket{\log P_{\psi_j}\paren{z_j}}-H\paren{Q_{\varphi_j}\paren{Z_j|\hat{S}_j}}.
    \end{split}
\end{equation}
for each part $j=1, \dots, m$ and for each shape $S$.
Here $H$ is the entropy and $P_{\psi_j}\paren{Z_j}$ is the learnable part style prior distribution obtained by transforming a standard Gaussian with an invertible network:
$$
z_j := F_{\psi_j}\paren{\xi(t_0)} = \xi(t_0) + \int_{t_0}^{t_1}f_{\psi_j}(\xi(t), t)\,dt
$$
where $\xi(t_0)\sim \mathcal{N}(\bm{0}, \bm{I})$ and $f_{\bm\psi}$ is the continuous-time dynamics of the flow $F_{\psi_j}$. Then, the log probability can be calculated by 
$$
\log P_{\psi_j}\paren{z_i} = \log P\paren{F_{\psi_j}^{-1}\paren{z_j}} + \int_{t_0}^{t_1} \Tr\paren{\frac{\partial f_{\psi_j}}{\partial \xi(t)}}\,dt.
$$
We finally note that for shapes with missing parts, we replace its part style latent with a dummy latent variance, and mask it out at training time. 

\subsection{Transformation Sampler}\label{sec:sampler}
As detailed in Sec. 4.2 our the main paper, we train an implicit probabilistic model with conditional Impicit Maximum Likelihood (cIMLE) method. Specifically, we model a multimodal distribution of part transformations $P_\theta(\bm{T}|\bm{z})$ given the part style latents by a deep neural network $T_\theta : (\bm{z}, y) \mapsto \bm{\tau}$ where $\bm{z} \sim P_{\bm{\psi}}(\bm{Z})$ and $y\sim \mathcal{N}(\bm{0}, \bm{I})$, which is a $32-$dimensional noise latent code. By sampling different noise latent code from the standard Gaussian, the neural network $T_\theta$ is able to produce different valid modes of part transformations for the given set of part style latents. To train the implicit model with cIMLE, we follow \cite{li2018implicit} and cache the best fitting noise latent code for each set of part style latents $\mathbf{z}$ during training, and optimize the transformation sampler using the cached noise codes for a number of iterations. The best fitting is defined by the cIMLE objective 
\begin{equation}\label{equ:cimle}
\begin{split}
       & \ell_{\tAll} = \sum_{S\in\mathbf{S}} \min_{k=1, \dots, K}\ell_{\text{fit}}\paren{T_\theta\paren{\bm{z}_S, y_k}, \tAll_S},  
\end{split}
\end{equation}
where 
$\tAll_S$ is the observed part transformations for shape $S$. $\ell_{\text{fit}}$ is defined to be the distance between the generated transformation $\{\tAll_k\}$ and observation $\tAll_S$ summed across all parts:
$$
\ell_{\text{fit}}\paren{\tAll, \tAll_S} = \sum_{j=1}^m \norm{c_j - c_{S, j}}^2_2 + \norm{\log s_j - \log s_{S, j}}^2_2.
$$ Same as in the main paper, $c_j, c_{S, j}\in \R^3, s_j, s_{S, j}\in \R^3$ are the shifts and scales associated with $j$-th part of the generated transformation and the observed transformation respectively. 

\subsection{Cross Diffusion Network}\label{sec:cdn}
To learn the data distribution $P_\phi(S|\bm{z}, \bm{\tau}),$ we learn a point distribution $P_\phi(X^{(0)}|\bm{z}, \bm{\tau}, j)$ for each part conditioned on the part style latents $\bm{z}$ and the part transformations $\bm{\tau}$. For each observed shape $S$, represented as a point cloud, we assume that points from each part are sampled independently from $P_\phi(X|\bm{z}, \bm{\tau}, j)$. We define a forward diffusion process $Q(X^{(0:T)}|\tau_j)$ for each part using the generalized forward kernel with $\mu = c_j$ and $\Sigma = \Diag(s_j)$ for each part. To learn the denoising process, we define a reverse process as another Markov chain $P_\phi(X^{(0:T)}|\bm{z}, \bm{\tau}, j)$ for each part such that
$$
P_\phi(X^{(0:T)}|\bm{z}, \bm{\tau}, j) = P(X^{(T)}|\tau_j)\prod_{t=1}^TP{\phi}(X^{(t-1)}|X^{(t)}, \bm{z}, \bm{\tau}, j)
$$ where 
$$
P(X^{(T)}|\tau_j) = \mathcal{N}\paren{c_j, \Diag{s_j}}
$$ and each reversion kernel is parameterized as a Gaussian Distribution with learnable means to approximate the posterior distribution $q\paren{X^{(t-1)}|X^{(t)}, X^{(0)}, c_j, \Diag(s_j)}$:
$$
P_{\phi}(X^{(t-1)}|X^{(t)}, \bm{z}, \bm{\tau}, j) = \mathcal{N}\paren{\Xi_{\phi}\paren{t, X^{(t)}, \bm{z}, \bm{\tau}, j}, \eta^2_t\Diag\paren{s_j}}.
$$ 
Following \cite{dpm, Zhou_2021_pvd, nichol2022point, lion}, instead of directly learning the mean $\Xi_\phi,$ we reparameterize the forward posterior distribution according to Eq. \ref{equ:reparam} to learn $\bm{\varepsilon}_\phi$ that approximate noises from standard Gaussian. Thus, the final objective becomes 
\begin{equation}
\ell_{\text{cross}} = \sum_{j=1}^m\sum_{x\in S_j}\E_{\varepsilon \sim \mathcal{N}\paren{\bm{0}, \bm{I}}, \bm{z}\sim Q_{\bm{\varphi}}, t \in\set{1, \dots, T}}\bracket{\norm{\bm{\varepsilon} - \bm{\varepsilon}_\phi\paren{x^{(t)}, \bm{z}, \tAll, j, t}}^2_2}.
\end{equation}

\section{Implementation Details}
\label{sec:implementation}
In this section, we provide the implementation details: our training pipeline and hyperparameters, network architecture, and training and sampling algorithms.

\subsection{Training Details}\label{sec:training-details}
As detailed in the main paper, our training loss is transformed to 
$\ell_{\text{total}} = \ell_{\text{recon}} + \lambda_1\ell_{\mathbf{Z}} + \lambda_2\ell_{\tAll}$. In practice, we adopt a two-stage training strategy. During the first stage, we train the prior flow $P_{\bm{\psi}}(Z_j),$ variational encoders $Q_{\bm{\varphi}}(\bm{Z}|\hat{S}),$ and the cross diffusion network $P_\phi(X|\bm{Z}, \bm{\tau}_S, j)$. For the Cross Diffusion Network, We use timestep $T=100$ for all of our models and the variance scheduling parameters $\alpha_t$ are set to linearly decrease from $0.9999$ to $0.08$ following the setting of \cite{dpm}. Ground truth transformations are used to condition the Cross Diffusion network during the first stage. The loss optimized for the first stage of training is 
\begin{equation}
\begin{split}
        \ell_{\text{first stage}} &= \lambda_1\ell_{\mathbf{Z}} + \ell_{\text{recon}}\\
    &=  \E_{S\sim P(S)}\bigg[-\lambda_1\E_{z_j \sim Q_{\varphi_j}(Z_j|\hat{S}_j)}\bracket{\log P_{\psi_j}\paren{z_j}}-H\paren{Q_{\varphi_j}\paren{Z_j|\hat{S}_j}}\\
    & \phantom{=}+\frac 1m\sum_{j=1}^m\frac{1}{\abs{S_j}}\sum_{x\in S_j}\E_{\varepsilon \sim \mathcal{N}\paren{\bm{0}, \bm{I}}, \bm{z}\sim Q_{\bm{\varphi}}, t \in\set{1, \dots, T}}\bracket{\norm{\bm{\varepsilon} - \bm{\varepsilon}_\phi\paren{x^{(t)}, \bm{z}, \tAll_S, j, t}}^2_2}\bigg].
\end{split}
\end{equation}
\noindent where $\lambda_1=$5e-4 for chairs and $\lambda_1=$1e-3 for lamps, airplanes, and cars. For all models, we use a batch size of $128$, and the first stage is trained for $8000$ epochs using Adam \cite{adamoptim} optimizer without weight decay. Momentum parameters $\beta_1$ and $\beta_2$ for Adam optimizer are set to $0.9$ and $0.999$ respectively. The learning rate is set to 2e-3 at the start of training as is linearly decreased starting from $4000$ epochs to 1e-4 at the end of training. The first stage takes around 36 hours for chairs, 30 hours for airplanes, 24 hours for lamps, and 16 hours for cars. 

In the second stage, we freeze the weights of the part stylizers and the cross diffusion network, and only train the transformation sampler using the cIMLE training strategy. The second stage loss is the cIMLE loss plus the reconstruction loss: 
\begin{equation}
\begin{split}
        \ell_{\text{second stage}} &=  \ell_{\text{recon}} + \lambda_2\ell_{\tAll},
\end{split}
\end{equation}
where $\lambda_2=1$ for all classes. The training of cIMLE requires recaching of noise latent codes. Specifically, we recache the noise latent codes for each shape in our training data every $50$ epoch, during which we sample $K=20$ noise latent code and associate each set of part style latents $\bm{z}$ with a best fitting code according to Eq. \ref{equ:cimle}. These pairs are then trained for $50$ epochs before the noise codes are recached. The second stage is trained for a maximum of $4000$ epochs and the models used for evaluation are selected from the best checkpoint. Similar to the first stage, we use the Adam optimizer, and keep the learning rate fixed at 2e-4 for the entire training process. 

\subsection{Network Parameters}

\subsubsection{Part Style Sampler}
Each variation encoder $Q_{\varphi_j}(Z_j|\hat{S}_j)$ is implemented as a PointNet \cite{pointnet} following the architecture of \cite{dpm, ShapeGF} with a shared per point feature regression layer. Specifically, we feed the canonicalized parts through a 3-128-256-512 MLP layer with ReLU nonlinearity and batch normalization. We share the per-point MLP across all parts to reduce parameter count. The per-point features are then max-pooled for each part to obtain a $512$ dimensional feature for each part. Then, the feature is fed into a 512-256-128-256 MLP with the ReLU nonlinearity to output a $256$ dimensional mean and a diagonal variance which parameterize the Gaussian distribution of $Q_{\varphi_j}(Z_j|\hat{S}_j) = \mathcal{N}\paren{\mu_{\varphi_j}\paren{\hat S_j}, \sigma_{\varphi_j}\paren{\hat S_j}}$. We note that the MLP after max pooling is not shared across different parts as the objective decomposition assumes independence of part style latent distributions. The $256$ dimensional part style latent is then sampled from this Gaussian for downstream training. 

Because each part style latent distribution is independent of the other, we use one prior flow network for each part. Each prior flow is implemented with the same architecture as \cite{dpm, pointflow}. Specifically, we use 14 affine coupling layers with the dimension of hidden states being $256$, identical
to the dimension of part style latents. Following each of the layers, we apply moving batch normalization~\cite{batchnorm, neuralode}. Both the scaling and translation networks $F(\cdot)$ and $G(\cdot)$ are 128-256-256-128 MLPs with ReLU nonlinearity. 
\subsubsection{Transformation Sampler}
The network for transformation sampler $T_\theta$ is implemented using a self-attention-only transformer following~\cite{nichol2022point, rombach2021highresolution}. The input is $m$ tokens with each token being the concatenation of the part style latent $z_j\in \R^{256}$ and a scaled noise latent code $\lambda y\in \R^{32}$. The $\lambda$ parameter is class specific and controls the degree of transformation variations produced by different modes of the transformation sampler. For the reported models, we use $\lambda=100$ for chairs, $\lambda=50$ for airplanes and cars, and $\lambda=10$ for lamps. Then, each token is projected back to a $256$ dimensional vector and the part label associated with each token is added as a learnable embedding vector to the projected token. For the transformer architecture, we use 5 layers of multi-head self-attention followed by a feed-forward layer and layer normalization. Parameters can be found in \cref{tab:tau-param}.
\begin{table}[h]
\begin{center}
\begin{tabular}{l|c}
\hline
parameters  & Values  \\ \hline
layers & 5\\
head dimension & 32\\
num heads  & 8\\
drop out rate & 0 \\ \hline
\end{tabular}
\caption{Parameters for Transformation Sampler.}
\label{tab:tau-param}
\end{center}
\end{table}
The output of $T_\theta$ is a three-dimensional scale and shift vector for each part style latent. For missing parts, we mask out its associated token during the self-attention layers. 
\subsubsection{Cross Diffusion Network}
As derived in \cref{equ:reparam}, we learn a noise approximator $\bm{\varepsilon}_{\phi}$ that estimates pure noises given noisy inputs at time step $t$. We use a cross-attention-only transformer to implement $\bm{\varepsilon}_\phi$. Specifically, the input consists of $N$ tokens where $N$ is the number of points in a shape $S$. Each input token is the concatenation of $(x^{(t)}, \tau_j, j)$ if the point $x^{(t)}$ belongs to the $j$-th part. Because of computation cost and independence assumption across different point samples, the input tokens are first projected to $128$ dimensional vectors and then individually attend to $m$ context tokens each being the concatenation of $(z_j, \tPart_j, j, t)$, for $j=1, ..., m$ followed by a feedforward layer and layer normalization. The transformer architecture is similar to that of the transformation sampler, with parameters listed in Tab. \ref{tab:ddpm-param}.
\begin{table}[h]
\begin{center}
\begin{tabular}{l|c}
\hline
parameters                   & Values  \\ \hline
layers & 5\\
head dimension & 16\\
num heads  & 8\\
drop out rate & 0.2 \\ \hline
\end{tabular}
\caption{Parameters for Cross Diffusion Network.}
\label{tab:ddpm-param}
\end{center}
\end{table}
\subsection{Algorithm}
We provide the training and sampling algorithms for \methodname in Algorithm~\ref{algo:1} (first stage training), Algorithm~\ref{algo:2} (second stage training), and Algorithm~\ref{algo:sampling} (sampling). 

\begin{algorithm}
\caption{\methodname Training: Stage 1}
\label{algo:1}
\begin{algorithmic}
\Repeat
\State Let $S \in \bm{S}_{\text{data}}$
\For{$j = 1, \dots, m$}
    \If{$S_j \neq \varnothing$} 
        \State Sample $z_j \sim Q_{\varphi_j}\paren{Z_j|\hat{S}_j}$.
    \Else
        \State $z_j = z_{\text{dummy}}$.
    \EndIf 
\EndFor
\State $\bm{z} \gets \set{z_j}_{j=1}^m$
\State Let $\bm{\tau}_S$ to be the part transformations for $S$. 
\State Sample $t \sim \text{Uniform}\paren{\set{1, \dots, T}}$
\State Sample $\bm{\varepsilon}\sim \mathcal{N}\paren{\bm{0}, \bm{I}}$
\State Compute $\nabla_{\phi, \bm\varphi, \bm{\psi}} [\ell_{\text{first stage}}]$; then perform gradient descent.
\Until{\text{converged.}}
\end{algorithmic}
\end{algorithm}
\begin{algorithm}
\caption{\methodname Training: Stage 2}
\label{algo:2}

\begin{algorithmic}
\Repeat
    \For{$S \in \bm{S}_{\text{data}}$}
        \State Let $\bm{\tau}_S$ to be the part transformations for $S$. 
        \For{$j = 1, \dots, m$}
            \If{$S_j \neq \varnothing$} 
                \State Sample $z_j \sim Q_{\varphi_j}\paren{Z_j|\hat{S}_j}$.
            \Else
                \State $z_j = z_{\text{dummy}}$.
            \EndIf 
        \EndFor
        \State $\bm{z} \gets \set{z_j}_{j=1}^m$
        \State Sample $y_{1}, \dots, y_K\sim \mathcal{N}\paren{\bm{0}, \bm{I}}$
        \State Set $y^*_S = \arg\min_{y_i \in \set{y_{1}, \dots, y_K}}\ell_{\text{fit}}\paren{T_\theta(\bm{z}, y_i), \bm{\tau}_S}$.
    \EndFor
    \For{$l = 1, \dots, 50$}
        \State Sample $S \in \bm{S}_{\text{data}}$
        \For{$j = 1, \dots, m$}
            \If{$S_j \neq \varnothing$} 
                \State Sample $z_j \sim Q_{\varphi_j}\paren{Z_j|\hat{S}_j}$.
            \Else
                \State $z_j = z_{\text{dummy}}$.
            \EndIf 
        \EndFor
        \State $\bm{z} \gets \set{z_j}_{j=1}^m$
        \State Let $\bm{\tau}_S$ to be the part transformations for $S$. 
        \State Sample $t \sim \text{Uniform}\paren{\set{1, \dots, T}}$
        \State Sample $\bm{\varepsilon}\sim \mathcal{N}\paren{\bm{0}, \bm{I}}$
        \State Compute $\nabla_{\theta} \bracket{\ell_{\text{recon}} +\ell_{\text{fit}}(T_\theta(\bm{z}, y^*_S), \bm{\tau}_S)}$; then perform gradient descent.
    \EndFor
\Until{\text{converged.}}
\end{algorithmic}
\end{algorithm}

\begin{algorithm}
\caption{\methodname Sampling}
\label{algo:sampling}
\begin{algorithmic}
\For{$j = 1, \dots, m$}
    \State Sample $\xi \sim \mathcal{N}\paren{\bm{0}, \bm{I}}$.
    \State $z_j = F_{\psi_j}(\xi(t_0))$
\EndFor
\State $\bm{z} \gets \set{z_j}_{j=1}^m$
\State Sample $y\sim \mathcal{N}(\bm{0}, \bm{I})$
\State Compute $\bm{\tau} = \set{(c_j, s_j)}_{j=1}^m = T_\theta(\bm{z}, y)$
\For{$j = 1, \dots, m$}
    \State $S_j^{(T)} = \set{x^{(T)}} \sim \mathcal{N}\paren{c_j, \Diag(s_i)}$
\EndFor
\State $S^{(T)} \gets \set{S_j^{(T)}}_{j=1}^m$
\For{$t = T, \dots, 1$}
    \For{$i = 1, \dots, m$}
        \For{$x^{(t)} \in S_j^{(t)}$}
            \State $x^{(t - 1)} \sim P_{\bm{\theta}}\paren{X^{(t - 1)}|x^{(t)}, \bm{z}, \bm{\tau}, j}$
        \EndFor
        \State $S_i^{(t-1)}\gets \set{x^{(t - 1)}}$
    \EndFor
    \State $S^{(t-1)}\gets \set{S_j^{(t - 1)}}_{j=1}^m$
\EndFor
\State \Return $S^{(0)}$
\end{algorithmic}
\end{algorithm}

\section{Experiment Set-up Additional Details}
\label{sec:experiment_setup}

\subsection{Control-Enabled Baselines Additional Details}
Because none of the existing networks enable part-level generation, we modify two state-of-the-art baselines (LION~\cite{lion} and ShapeGF~\cite{ShapeGF}) to enable part-level sampling for a fair comparison with \methodname, we refer to these control-enabled modification as Ctrl-LION and Ctrl-ShapeGF. To enable part-level generation, we train their method per part, \ie one model for each part of an object class. We use the same dataset as our method, namely ShapeNet with semantic segmentation labels. As a result, we obtain a generative model for each part for both ShapeGF and LION networks. Then, to obtain the global shape, we simply sample from each of the perpart Ctrl-ShapeGF and Ctrl-LION and concatenate the points together. To sample on the part level, we simply sample from a specific part latent space while fixing the rest of the generated shapes. 

\subsection{Evaluation Protocol Additional Details}
We elaborate on the evaluation protocol we use to compare with baselines here. Sec.~\ref{sec:pgen} details the intro-part evaluation procedure, which we use to evaluate the quality and diversity of our per-part distributions. Sec.~\ref{sec:psnap} elaborate on the snapping metric, which we use to measure against the control-enabled baselines. Sec.~\ref{sec:human_study_supp} details the human study that was conducted.

\subsubsection{Intra-part Evaluation Additional Details}\label{sec:pgen}
As presented in the main paper, we use the standard generation metrics to measure the similarity between the distributions of canonicalized parts of the generated shapes compared to the test set from~\cite{Yi16} of segmented shapes. For a set of canonicalized part $\bm{S}_{t, j}$ from the test set and a set of canonicalized part $\bm{S}_{g, j}$ from the generated set, the generation metrics are computed as

\noindent \textbf{Minimum matching distance (MMD-P)}
$$
\text{MMD-P}(\bm{S}_{g, j}, \bm{S}_{t, j}) = \frac{1}{\abs{\bm{S}_{t, j}}}\sum_{S_{t, j}\in \bm{S}_{t, j}}\min_{S_{g, j}\in \bm{S}_{g, j}}\text{Chamfer}\paren{S_{g, j}, S_{t, j}}.
$$
The idea behind MMD-P is to calculate the average distance between the point clouds in the reference set and their closest neighbors in the generated set. A smaller MMD-P implies that the parts from the test set are well represented in the parts from the generated set.

\noindent \textbf{Coverage (COV-P)}
$$
\text{COV-P}(\bm{S}_{g, j}, \bm{S}_{t, j}) = \frac{\abs{\set{\arg\min_{S_{t, j}\in \bm{S}_{t, j}}\text{Chamfer}\paren{S_{t, j}, S_{g, j}}}}}{\abs{S_{t, j}}}.
$$
A higher COV-P percentage means that parts in the test set are well covered by the parts in the generated set, so it implies that the generated parts are diverse. 

\noindent\textbf{1-NN classifier accuracy (1NNA-P)}
$$
\text{1NNA-P}(\bm{S}_{g, j}, \bm{S}_{t, j}) = \frac{\sum_{S_{g, j}\in \bm{S}_{g, j}} \mathbbm{1}[N_{S_{g, j}}\in \bm{S}_{g, j}] + \sum_{S_{t, j}\in \bm{S}_{t, j}} \mathbbm{1}[N_{S_{t, j}}\in \bm{S}_{t, j}]}{\abs{S_{t, j}} + \abs{S_{g, j}}}.
$$
where $\mathbbm{1}[\cdot]$ is the indicator function and $N_{S_j}$ is the nearest neighbor of $S_j$ in the set $S_{t, j} \cup S_{g, j} \setminus \set{S_j}$ (i.e., the union of the sets $S_{g, j}$ and $S_{t, j}$ excluding, $S_j$). With this definition, 1NNA-P represents the leave-one-out accuracy of the 1-NN classifier (measured in Chamfer distance) defined above. If the performance of the 1NN classifier is close to $50$ percent, the parts from the test set and the generated set are well mingled. Thus, 1NNA-P directly measures the similarity between the test parts and the generated parts, both in diversity and quality. 

To prepare for the sets of canonicalized parts, we use $512$ points for each of the part samples for the test set. For our model and all the baselines, we sample $2048$ points on the entire shape and obtain for each shape its canonicalized parts from these points. We sample the same number of shapes as in the test set. Since the baselines do not generate shapes with part segmentation, we enable their evaluation using a segmentor with PointNet++ backbone~\cite{qi2017pointnetplusplus} trained on the ShapeNetCore PartSeg dataset. Then, MMD-P. COV-P. and 1NNA-P is computed for each part and in the end, the score is reported as a weighted average based on the number of part samples in each of the part sets of the test shapes. For implicit methods, we sample $2048$ points from their generated mesh. The canonicalization of each part is done by fitting all three axes into the unit cube. 

\subsubsection{Inter-Part Evaluation Additional Details}\label{sec:psnap}
To measure the connectivity of the generated shapes, we select a set of connected parts and compute the Chamfer distance between the closest $N_{\text{SNAP}} = 30$ points between the two parts. Specifically, if the $j$-th part is connected to a set of parts $\set{S_k}$ (note that a part can be connected to multiple parts due to different structures of shapes), we define the snapping distance of connection $j$ to be
$$
\text{SNAP}(S_j) = \min_{S_k\in\set{S_k}}\text{Chamfer}\paren{N^{(N_{\text{SNAP}})}_{S_k}(S_j), N^{(N_{\text{SNAP}})}_{S_j}(S_k)},
$$
where $N^{(N_{\text{SNAP}})}_X(Y)$ are the $N_{\text{SNAP}}$ closest points in shape $Y$ to shape $X$. 

For chairs, we define connections of the back, seat, and arms respectively. The back connection is defined to \textit{connect to either the legs or the seat}; the seat connection is defined to be \textit{connected to the legs}, and lastly, the arm connection is defined to \textit{connect to either the back or the seat}. 

\subsubsection{Human Study Additional Details}
\label{sec:human_study_supp}
\begin{figure}[h]
\centering
\begin{center}
   \includegraphics[width=0.4\textwidth]{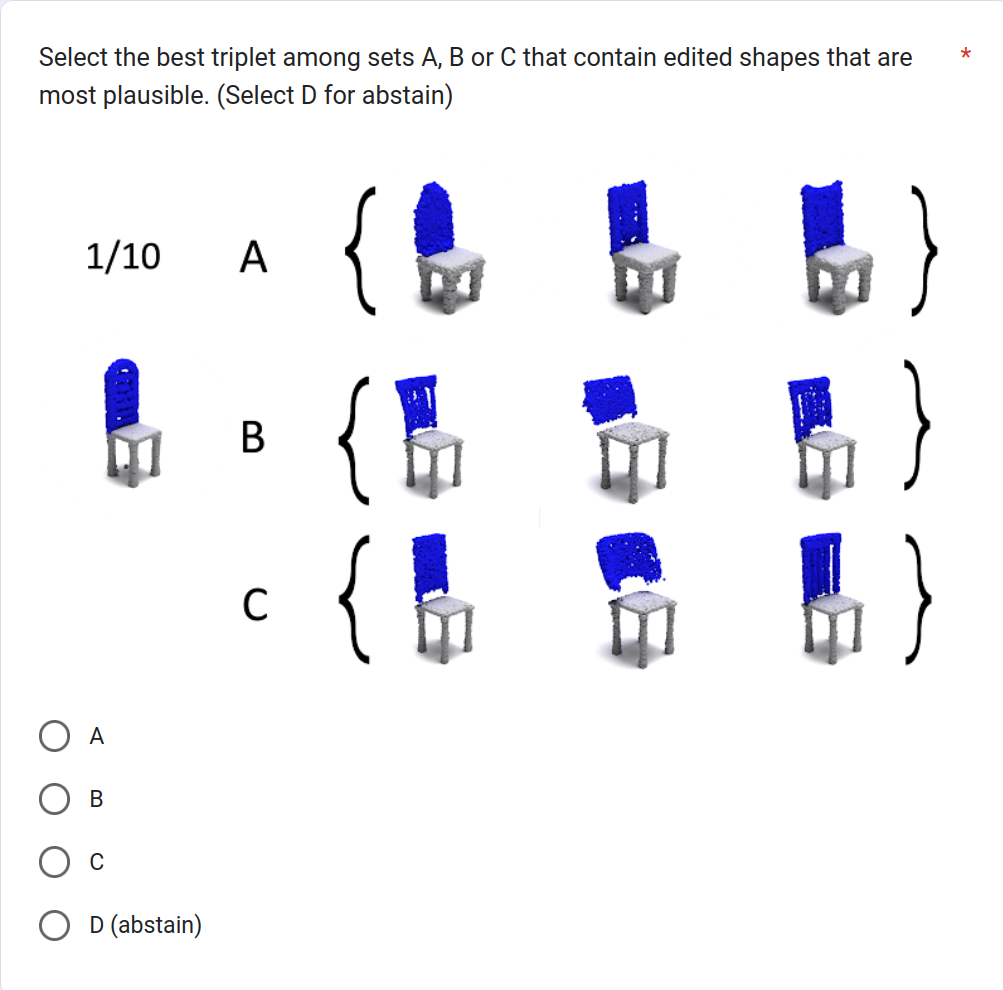}
\end{center}
   \caption{\textbf{User Study Example.} The subject is asked ten questions similar to the example above in which they are asked to pick the most coherent group out of the three.}
\label{fig:user}
\end{figure}

We provide further details on the user study we conducted to evaluate controllability of our method compared to the control-enabled baselines (Ctrl-ShapeGF and Ctrl-LION). A user is asked to select the triplet of edited shapes that are most plausible, where examples of plausible shapes to be: 1) does not have detachment of parts, 2) does not have interpenetrating parts, 3) the size of the edited parts agrees with the rest of the shape, which was included in the prompt. As described in the main paper, a (controlled) edit is defined as re-sampling a  pre-selected part of the given shape. Our user study contained 10 questions, which are randomly selected shapes from the dataset, each with random parts to edit. Fig.~\ref{fig:user} shows one example question from our user study.


{\small
\bibliographystyle{ieee_fullname}
\bibliography{egbib}
}

\end{document}